\documentclass{article}

\newif\iffastbuild
\ifdefined\FASTBUILD
\fastbuildtrue
\else
\fastbuildfalse
\fi

\usepackage{microtype}
\usepackage{graphicx}
\usepackage{subcaption}
\usepackage{booktabs} 

\usepackage{hyperref}

\usepackage[preprint]{neurips_2026}

\makeatletter
\renewcommand{\@noticestring}{Accepted to the 40th Conference on Neural Information Processing Systems (NeurIPS 2026), Evaluations and Datasets Track.}
\makeatother

\usepackage{amsmath}
\usepackage{amssymb}
\usepackage{mathtools}
\usepackage{amsthm}

\usepackage[table]{xcolor}
\usepackage{etoolbox}
\usepackage{array}

\newcolumntype{T}[1]{>{\ttfamily\raggedright\arraybackslash}p{#1}}

\newcommand{\llamabase}{\texttt{Llama-3.1-8B}}

\usepackage{hyperref}

\usepackage[utf8]{inputenc} 
\usepackage[T1]{fontenc}    
\usepackage{hyperref}       
\usepackage{url}            
\usepackage{booktabs}       
\usepackage{amsfonts}       
\usepackage{nicefrac}       
\usepackage{microtype}      
\usepackage[most]{tcolorbox}
\usepackage{amssymb} 

\usepackage{amsmath}
\usepackage{makecell}
\usepackage{graphicx}
\usepackage{subcaption}
\usepackage{enumitem}
\usepackage{fvextra}
\usepackage{multirow}
\usepackage{multicol}
\usepackage{longtable}
\usepackage{booktabs}
\usepackage{color,colortbl} 
\usepackage[normalem]{ulem}
\usepackage{amssymb}
\usepackage{pifont}
\DeclareRobustCommand{\cmark}{\textcolor{lightgreen}{\ding{51}}} 
\DeclareRobustCommand{\xmark}{\textcolor{lightred}{\ding{55}}}   

\definecolor{lightergreen}{RGB}{115,181,74}
\definecolor{lightgreen}{RGB}{0,153,0}
\definecolor{darkgreen}{RGB}{0,102,0}

\definecolor{lightblue}{RGB}{51,204,255}
\definecolor{darkblue}{RGB}{47,85,151}
\definecolor{lightred}{RGB}{251,51,51}
\definecolor{darkred}{RGB}{190,0,0}

\definecolor{orange}{RGB}{230,115,0}
\definecolor{grey}{RGB}{128,128,128}

\definecolor{cellblue}{RGB}{220,230,255}
\definecolor{cellred}{RGB}{255,220,220}

\iffastbuild
\AtBeginDocument{\setkeys{Gin}{draft}}
\fi

\title{DEBATE: A Large-Scale Benchmark for Evaluating Opinion Dynamics in Role-Playing LLM Agents}

\author{%
  \normalfont\small
  \begin{tabular}{c}
  \textbf{Yun-Shiuan Chuang}\textsuperscript{1}\footnotemark[2] \quad
  \textbf{Ruixuan Tu}\textsuperscript{1,2,3}\footnotemark[1] \quad
  \textbf{Chengtao Dai}\textsuperscript{1}\footnotemark[1] \quad
  \textbf{You Li}\textsuperscript{1}\footnotemark[1] \quad
  \textbf{Smit Vasani}\textsuperscript{1}\footnotemark[1] \\
  \textbf{Binwei Yao}\textsuperscript{1} \quad
  \textbf{Michael Henry Tessler}\textsuperscript{5} \quad
  \textbf{Sijia Yang}\textsuperscript{1} \quad
  \textbf{Dhavan Shah}\textsuperscript{1} \quad
  \textbf{Robert Hawkins}\textsuperscript{4} \\
  \textbf{Junjie Hu}\textsuperscript{1} \quad
  \textbf{Timothy T.~Rogers}\textsuperscript{1,6}\footnotemark[3] \\[2pt]
  \textsuperscript{1}University of Wisconsin--Madison \quad
  \textsuperscript{2}University of Tokyo \\
  \textsuperscript{3}RIKEN Center for Advanced Intelligence Project (AIP) \\
  \textsuperscript{4}Stanford University \quad
  \textsuperscript{5}Google DeepMind \quad
  \textsuperscript{6}UCL Institute of Cognitive Neuroscience
  \end{tabular}
}

\begin{document}

\maketitle
\begingroup
\renewcommand{\thefootnote}{\fnsymbol{footnote}}
\footnotetext[1]{Co-authors with core contributions.}
\footnotetext[2]{First author contact: \texttt{yunshiuan.chuang@gmail.com}.}
\footnotetext[3]{Correspondence to Timothy T.~Rogers: \texttt{timothy.rogers@ucl.ac.uk}.}
\endgroup
\setcounter{footnote}{0}

\begin{abstract}
Modeling opinion change through social interactions is important for understanding polarization, misinformation, and societal conflict. Recent work simulates \emph{opinion dynamics} with role-playing LLM agents (RPLAs), but multi-agent simulations often display unnatural group behavior (e.g., premature convergence) and lack empirical benchmarks for assessing alignment with real human group interactions. We introduce \textbf{DEBATE}, a benchmark for evaluating the authenticity of opinion dynamics in multi-agent RPLA simulations. Its three-round design makes each participant's public interaction history complete and traceable. DEBATE contains multi-round public messages and private Likert-scale beliefs from U.S.-based participants across 107 topics. The dataset cover 2,788 participants in 697 groups, enabling evaluation at the utterance and group levels. We instantiate “digital twin” RPLAs with seven LLMs and evaluate across two settings: \textit{Next Message Prediction} and \textit{Full Dynamics Simulation}, using stance-based opinion-dynamics metrics. In zero-shot settings, RPLA groups exhibit strong opinion convergence relative to human groups. Across group-generalization tests with unseen participant groups and topic-generalization tests with unseen discussion topics, supervised fine-tuning (SFT) of \texttt{Llama-3.1-8B-Instruct}, \texttt{Qwen3.5-4B}, and \texttt{Mistral-7B-Instruct-v0.3} improves stance-based alignment and reduces excessive convergence, bringing RPLA group convergence closer to that observed in the corresponding human groups, though discrepancies in opinion change and belief updating remain. The dataset, benchmark website, and codebase are publicly available.\footnote{Dataset: \url{https://huggingface.co/datasets/seantw/DEBATE_LLM}; Benchmark website: \url{https://debate-llm-benchmark.vercel.app/}; Codebase: \url{https://github.com/yunshiuan/DEBATE-neurips2026}.}
\end{abstract}

\section{Introduction}
\label{sec:introduction}

Understanding how individual opinions change through social interactions is crucial across numerous domains, e.g., public health campaigns, conflict resolution, and misinformation mitigation \citep{lu2015competition,pennycook2021shifting,budak2011limiting,loomba2021measuring,ginossar2022cross}. Accurate modeling of these \textit{opinion dynamics} not only helps predict critical societal phenomena like opinion polarization but also informs effective interventions to mitigate adverse outcomes.

\begin{figure*}[tb!] 
\centering
\includegraphics[width=\linewidth]{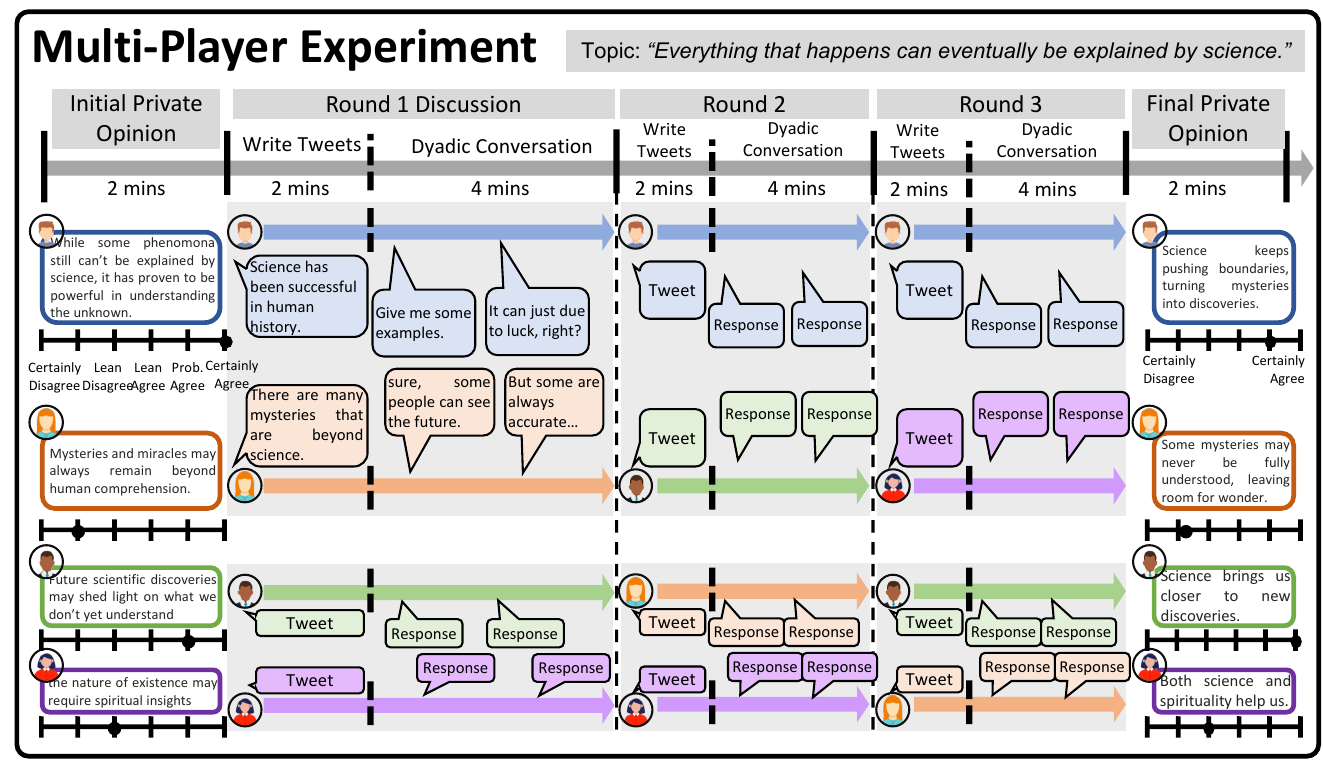}
\caption{Procedure for the multi-player deliberation experiment. Participants report an initial private opinion, complete three rounds of public tweet-like posts and dyadic chats with different partners, and then report a final private opinion.}
\label{fig:exp_diagram}
\end{figure*}

Recent advances in large language models (LLMs) have unlocked new possibilities for simulating human social interactions, particularly through the use of role-playing LLM agents (RPLAs) that embody diverse personas and engage in multi-turn dialogue \citep{park2023generative,chuang2024simulating,chuang2024wisdom}. Although individual RPLAs can often convincingly emulate human-like behaviors, prior research indicates that this single-agent authenticity does not guarantee realistic emergent dynamics in multi-agent settings. Specifically, when multiple RPLAs interact, they frequently exhibit premature consensus convergence, overly moderate stances, or unnatural patterns of opinion alignment, regardless of their initial diverse personas \citep{chuang2024simulating,taubenfeld2024systematic}. Existing evaluations of RPLAs predominantly focus on single-agent scenarios or employ artificial, structured tasks, lacking robust empirical benchmarks capturing authentic human group dynamics in naturalistic contexts \citep{santurkar2023whose,chuang2024beyond,chuang2024wisdom}.

To address this critical gap, we introduce \textbf{D}eliberative Opinion \textbf{E}xchanges for \textbf{B}enchmarking \textbf{A}gent-based \textbf{T}rajectory \textbf{E}volution (\textbf{DEBATE}), the first large-scale empirical benchmark specifically designed for evaluating the authenticity of simulated opinion dynamics from multi-agent RPLAs. While the acronym suggests debate, DEBATE emphasizes naturalistic deliberation rather than competitive or adversarial argumentation. DEBATE is built from short-horizon, four-person discussions collected from 2,832 recruited U.S.-based participants across 107 controversial topics. The dataset covers 2,788 participants in 697 groups. Here, ``large-scale'' refers to the benchmark's number and diversity of participants, groups, and topics, rather than to the size of each interaction network. The three-round, round-robin protocol makes each participant's public exposures complete and traceable, which enables matched human--RPLA evaluation. It captures both publicly expressed \textit{messages} (including both tweet-like posts and chat utterances) and privately reported beliefs (Likert-scale ratings). The benchmark splits used in the experiments are summarized in Table~\ref{tab:dataset-split-summary}, and Appendix~\ref{appendix:dataset-cleaning} details the data cleaning procedure, validation criteria, train/test split construction, and evaluation coverage. The benchmark enables quantitative assessment of alignment between simulated and actual human interactions at the \textit{utterance}, \textit{individual}, and \textit{group} levels.

\textbf{Contributions.} (1) We introduce \textbf{DEBATE}, the first large-scale empirical benchmark for evaluating the human-likeness of opinion dynamics in multi-agent role-playing LLM agents (RPLAs). (2) DEBATE supports two simulation setups, \textit{Next Message Prediction} and \textit{Full Dynamics Simulation}, covering key scenarios in opinion dynamics simulation. (3) We design quantitative evaluation metrics at three different levels (utterance, individual, and group) to assess different aspects of alignment between simulated and human opinion trajectories. (4) We identify behavioral gaps between RPLAs and humans such as over-convergence, over-sensitivity to peer opinion, highlighting the challenge of achieving fully human-like opinion dynamics. (5) We study supervised fine-tuning (SFT) across \texttt{Llama-3.1-8B-Instruct}, \texttt{Qwen3.5-4B}, and \texttt{Mistral-7B-Instruct-v0.3}, using group-generalization tests with unseen participant groups and topic-generalization tests with unseen discussion topics. Across these models, SFT improves the stance-based metrics and reduces excessive convergence, bringing RPLA group convergence closer to that observed in the corresponding human groups. Our evaluation, fine-tuning, and analyses are intended as examples of how the benchmark can be used to evaluate human–RPLA alignment. We expect DEBATE to enable future studies on RPLA opinion dynamics simulations. The dataset and benchmark website are publicly available at \url{https://huggingface.co/datasets/seantw/DEBATE_LLM} and \url{https://debate-llm-benchmark.vercel.app/}, respectively; Appendix~\ref{app:supp} summarizes the available artifacts.\footnote{ChatGPT was used for language polishing (Appendix~\ref{app:llm-disclosure}).}

\section{Related Work}
\label{sec:related}

\textbf{Simulating and Evaluating Opinion Dynamics with RPLAs.} Opinion dynamics concerns how beliefs form and change through social interaction \citep{flache2017models,lorenz2021individual,chuang2023computational}. Some recent multi-agent LLM work studies opinion exchange mainly as a way to improve downstream task performance \citep{zhang2023exploring, chan2023chateval, chen2023agentverse, du2023improving, liang2023encouraging, hu2024debate}. A separate line of work instead aims to simulate human-like opinion change with persona-grounded RPLAs \citep{chuang2024simulating, taubenfeld2024systematic, liu2024skepticism}, but the evidence is still limited by toy settings, qualitative analyses, or evaluations that do not compare against real multi-agent human interactions \citep{santurkar2023whose, chuang2024beyond, chuang2024wisdom}. DEBATE addresses this gap by providing a large-scale benchmark for evaluating natural-language multi-agent opinion dynamics against human data.

\textbf{Micro- and Macro-Level Opinion Dynamics.} A complementary literature studies population-level dynamics in large social networks, including ideological exposure \citep{bakshy2015exposure}, longitudinal polarization \citep{garimella2017long}, and echo chambers across platforms \citep{cinelli2021echo}. Network models further examine how biased assimilation, homophily, and heterogeneous activity can produce polarization and segregation \citep{dandekar2013biased,baumann2020modeling}. These studies address evolving interaction structures and long-run population patterns. DEBATE complements this work by providing controlled, matched human trajectories for evaluating whether RPLAs reproduce the interpersonal opinion updates which larger simulations are built on.

\textbf{Existing Opinion Dynamics Corpora.} Prior corpora with opinionated interaction can be loosely grouped into competitive debate, asymmetric persuasion, and naturalistic deliberation \citealp{walton1995commitment, walton2010argumentation, bozdag2025must}. For benchmarking human-like opinion dynamics, naturalistic deliberation is the most relevant genre, but it is relatively underrepresented. More importantly, existing datasets usually miss at least one requirement that matters for this benchmarking setting: multi-party multi-turn interaction, reader-context traceability, enforced turn-taking, demographic coverage, or both public and private opinion measurements \citep{tan2016winning,walker2012iac,abbott2016internet,wang2019persuasion,flache2017models,tourangeau2007sensitive}. Observational social-media corpora provide far greater scale but offer complementary evidence: reply trees need not reveal each reader's exposure history or opinion trajectory, and attention to the available content often must be inferred \citep{ojer2026modeling}. Appendix~\ref{app:dataset_comparison} details this quality--scale tradeoff.

\textbf{Positioning DEBATE as a Benchmark.} DEBATE is designed to combine the properties most useful for benchmarking multi-agent opinion dynamics: multi-party interaction, full reader context, enforced turn-taking, demographic attributes, and both public and private opinion measurements. It is therefore not just a dataset but an evaluation benchmark for testing whether simulated trajectories recover real human opinion dynamics. \iffastbuild The appendix provides the full corpus-by-corpus comparison.\else Appendix Table~\ref{tab:dataset_comparison} provides the full corpus-by-corpus comparison.\fi

\section{DEBATE Benchmark: Empirical Opinion Dynamics from Humans}
\label{sec:dataset}

\begin{table}[tb!]
\caption{\textbf{DEBATE train/test partitions.} Counts summarize the size of human data; \%on-topic reports the fraction satisfying the message-validity criteria (Appendix~\ref{appendix:dataset-cleaning}).}
\centering
\small
\resizebox{\linewidth}{!}{%
\begin{tabular}{ll r rr rr rr rr}
\toprule
& & & \multicolumn{2}{c}{\textbf{Tweet posts}} & \multicolumn{2}{c}{\textbf{Init./post opinions}} & \multicolumn{2}{c}{\textbf{Chat messages}} & & \\
\cmidrule(lr){4-5}\cmidrule(lr){6-7}\cmidrule(lr){8-9}
\textbf{Split} & \textbf{Partition} & \textbf{\# topics}
  & \textbf{N} & \textbf{\%on-topic}
  & \textbf{N} & \textbf{\%on-topic}
  & \textbf{N} & \textbf{\%on-topic}
  & \textbf{\# subjects} & \textbf{\# groups} \\
\midrule
Group-gen. & Train & 107 & 6,474 & 93.8\% & 4,300 & 86.0\% & 12,306 & 99.4\% & 2,260 & 565 \\
 & Test & 104 & 1,498 & 94.5\% & 1,002 & 86.1\% & 2,706 & 99.4\% & 528 & 132 \\[2pt]
Topic-gen. & Train & 75 & 5,729 & 93.4\% & 3,813 & 85.2\% & 10,814 & 99.2\% & 2,004 & 501 \\
 & Test & 32 & 2,243 & 95.3\% & 1,489 & 88.0\% & 4,198 & 99.9\% & 784 & 196 \\
\midrule
Total      &       & 107 & 7,972 &        &  5,302 &       & 15,012 &        & 2,788 & 697 \\
\bottomrule
\end{tabular}%
}
\label{tab:dataset-split-summary}
\end{table}

\subsection{Task}
\label{subsec:task}

We design a multi-player conversational experiment to elicit naturalistic opinion dynamics (Figure~\ref{fig:exp_diagram}). The dataset comprises $G$ groups, each consisting of $N = 4$ participants $\{s_1, s_2, s_3, s_4\}$ assigned to a single controversial topic $t \in \mathcal{T}$. We deliberately use four-person groups: three rounds let every participant interact with each of the other three exactly once, yielding complete and traceable public interaction histories while preserving multi-party deliberation. Thus, for every human response and corresponding agent simulation, we know the content presented beforehand---a requirement for the matched \textit{Next Message Prediction} and \textit{Full Dynamics Simulation} evaluations in Section~\ref{subsec:agent_simulation}. Each session lasts 25--30 minutes and has four phases; see Appendix~\ref{app:human_exp_ui} for the interface and Appendix~\ref{app:example_chats} for example conversations.

\textbf{(1) Initial Private Opinion} (2-minute): Each participant $s_i$ reports an initial opinion $o^{\text{init}}_{s_i} \in \{1,2,\dots,6\}$ on a 6-point Likert scale\footnote{Participants selected from the six labels displayed in the interface: \textit{(1) Certainly disagree}, \textit{(2) Probably disagree}, \textit{(3) Lean disagree}, \textit{(4) Lean agree}, \textit{(5) Probably agree}, \textit{(6) Certainly agree}.}, together with a free-text justification $j^{\text{init}}_{s_i}$.

\textbf{(2) Public Opinion Exchanges} (6-minute): Participants complete $R = 3$ rounds of dyadic interaction, meeting each of the other three group members exactly once. In each round, both participants first write a tweet-like post summarizing their opinion, then engage in a 4-minute real-time chat about the assigned topic. This structure yields fully observed per-participant exchange histories while keeping the interaction pattern close to pairwise-update settings in classic opinion-dynamics models. Speaker turns alternate during collection, and consecutive messages from the same speaker are merged during preprocessing.

\textbf{(3) Post-discussion Private Opinion} (2-minute): After the final round, each participant privately submits a final opinion $o^{\text{final}}_{s_i}$ and a free-text justification $j^{\text{final}}_{s_i}$.

\textbf{(4) Demographic Survey}: Finally, participants report demographic attributes $d_{s_i}$ (e.g., age, gender, education, political orientation).

\subsection{Topics}
\label{subsec:topic}

The benchmark covers 107 controversial discussion topics curated from the World Values Survey~\citep{haerpfer2022world}, the Pew Global Attitudes Survey~\citep{pew_2025}, and a study of human belief networks~\citep{chuang2024beyond}. We selected topics with high response entropy among U.S. participants so that individuals naturally disagree rather than converge. Topics are phrased as self-contained declarative statements spanning domains such as science, policy, and social values; examples include \textit{``The position of the planets at the time of your birth can influence your personality.''} and \textit{``Euthanasia can be justified.''} Appendix~\ref{appendix:topics} provides representative topics and examples of topic formulation.

\subsection{Human Data Collection and  Dataset Summary}
\label{subsec:recruitment}

We recruited 2,832 unique participants who reside in the U.S. via the Prolific platform~\citep{palan2018prolific}\footnote{\url{https://www.prolific.com/}}. Participants were randomly assigned to four-person groups, producing 708 raw groups before cleaning. They remained anonymous to each other, identified only by randomly-assigned avatars and pseudonyms (e.g., \texttt{ZK48UT}). All procedures were approved by the Institutional Review Board (IRB). Participants were compensated at a rate of \$10/hour. Some groups experienced participant dropouts or technical interruptions, resulting in partially completed sessions.

To construct the DEBATE benchmark, we retain only groups in which all four participants complete the experiment end-to-end and each participant provides valid/on-topic content at every required stage: the initial and final private opinions, all three public tweets, and at least one chat utterance in each discussion round. Thus, every retained participant has a complete opinion trajectory and at least eight valid/on-topic items. Here, ``on-topic'' is shorthand for a broader message-level validity assessment: content must be relevant to the discussion, coherent, and in an appropriate format for its stage; filler-only, malformed, incoherent, repetitive, or otherwise nonfunctional content is invalid.The resulting cleaned benchmark contains 2,788 participants in 697 groups. Table~\ref{tab:dataset-split-summary} summarizes the human data under the two train/test partitions and on-topic rates after data cleaning. Appendix~\ref{appendix:dataset-cleaning} reports the number of groups available for SFT evaluation.

We operationalize this validity assessment with the binary label $I_{\text{topic}}(u,t)$, validated with a 10\% stratified manual sample across all 107 topics, the seven LLMs we evaluated, and both simulation setups for the tweet and private-opinion output families (11,161 tweets and 7,423 initial/post private-opinion texts in total), and reuse this label for benchmark cleaning and SFT data construction. Further validation details appear in Appendix~\ref{app:llm_eval_classifier}: LLM--human agreement is approximately 90\%, and two independent human annotators achieved 96.8\% agreement and Cohen's $\kappa=0.89$ for validity/topic relevance on a separate 400-message sample. A complementary audit of 120 dyadic conversations from 20 randomly sampled groups found that exchanges were coherent, responsive, and substantively engaged in 97.5--100\% of cases, with 99.4\% overall annotator agreement (Appendix~\ref{appendix:dataset-cleaning}).

The participants spanned a broad range of ages (18--84, $M = 39.7$, $SD = 13.0$), genders (47.5\% male, 51.4\% female), ethnicities (e.g., 67.6\% White, 22.9\% Black, 6.1\% Asian, 5.5\% Hispanic), educational backgrounds (ranging from high school to doctoral degrees), and income levels (from under \$25k to over \$200k). Participants also reported a wide variety of occupations (e.g.,  finance, engineering, healthcare, manufacturing). This diversity provides a robust foundation for modeling opinion dynamics across varied social perspectives (see Appendix~\ref{appendix:demographics} and Figure~\ref{fig:demographics} for details).

\section{Constructing and Simulating Role-Playing LLM Agents}
\label{subsec:agent}

\subsection{RPLA Construction Grounded in Human Data}
\label{subsec:agent_construction}

Each RPLA $a_i$ is designed as a \textit{digital twin} of a human participant $s_i$, simulating $s_i$’s conversational behavior throughout the multi-round interaction. Each $a_i$ is conditioned on a memory module $\mathcal{M}_{a_i,k}$ that aims to reflect $s_i$’s first-person perspective right before producing the $k$-th utterance in round~$r$. The memory is dynamically updated as tweets and utterances are exchanged.

The memory module $\mathcal{M}_{a_i,k}$ is instantiated via prompt templates that convert structured participant information into natural-language inputs for the LLM (see Appendix~\ref{sec:prompt_templates} and Table~\ref{tab:prompt-templates-simulation} for prompt examples). At each turn, it summarizes the participant's demographic profile, initial opinion and justification, previously observed public interaction history, and the current-round local context. We use notation with a hat and subscript $a$ (e.g., $\widehat{\tau}^r_{a_i}$, $\widehat{u}^r_{k, a_i}$) to denote LLM-generated content, and notation without a hat and with subscript $s$ (e.g., $\tau^r_{s_i}$, $u^r_{k, s_i}$) to denote human-written content. The exact sources of memory vary by the simulation setup (Section~\ref{subsec:agent_simulation}; Table~\ref{tab:sim_modes}). For example, the interaction history may come from real human exchanges in \textit{Next Message Prediction} or from recursively simulated interaction in \textit{Full Dynamics Simulation}.

\subsection{Simulating Social Interactions with RPLAs}
\label{subsec:agent_simulation}

\begin{table}[htbp!]
\centering


\caption{Memory contents in the two simulation setups. All agents receive demographics $d_{s_i}$, initial opinion and justification $(o^{\text{init}}_{s_i}, j^{\text{init}}_{s_i})$, the initial tweet $\tau^1_{s_i}$, and task instructions; blue entries indicate simulated content recursively added to memory.}
\label{tab:sim_modes}
\resizebox{0.8\linewidth}{!}{%
\begin{tabular}{p{3cm}p{3.2cm}p{2.8cm}p{2.8cm}}
\toprule
\textbf{Simulation Setup} & \textbf{Tweets in Memory} & \textbf{Utterances from Prior Rounds} & \textbf{Utterances from Current Round} \\
\midrule
\textbf{Next Message Prediction} & Human $\{\tau^{r'} : 1 \leq r' \leq r\}$ & Human $\{\mathcal{C}^{r'}_{s_i, s_j} : 1 \leq r' < r\}$ & Human $\{u^r_{m} : 1 \leq m < k\}$ \\
\midrule
\textbf{Full Dynamics Simulation} & Human $\tau^1$ + \textcolor{blue}{Simulated $\{\widehat{\tau}^{r'} : 2 \leq r' \leq r\}$} & \textcolor{blue}{Simulated $\{\widehat{\mathcal{C}}^{r'}_{a_i, a_j} : 1 \leq r' < r\}$} & \textcolor{blue}{Simulated $\{\widehat{u}^r_{m} : 1 \leq m < k\}$} \\
\bottomrule
\end{tabular}
}
\end{table}

We simulated each RPLA $a_i$’s utterance $\widehat{u}^r_{k, a_i}$ in round~$r$, turn~$k$ by generating:
\begin{align}
    \widehat{u}^r_{k, a_i} \sim P\left( u^r_{k, s_i} \,\middle|\, \mathcal{M}_{a_i,k} \right),
\end{align}
where the speaker identity $s_i$ is given, and only the utterance content is predicted.\footnote{Since consecutive messages from the same speaker are merged during preprocessing, speakers alternate turns, making the speaker order known (Section~\ref{subsec:task}).} The same framework applies to generating tweets $\widehat{\tau}^r_{a_i}$ and final opinions $(\widehat{o}^{\text{final}}_{a_i}, \widehat{j}^{\text{final}}_{a_i})$.

DEBATE supports two simulation setups (Table~\ref{tab:sim_modes}). \textit{Next Message Prediction} mirrors standard next-message prediction with human-written history and is treated as a secondary best-case baseline \citep{welch2019learning}. \textit{Full Dynamics Simulation} recursively rolls out the interaction from initial conditions, matching classic agent-based opinion-dynamics simulation and serving as the primary benchmark condition because it tests whether agents sustain realistic opinion dynamics once their own generations are fed back into memory \citep{flache2017models,taubenfeld2024systematic}. Comparing the two shows how much alignment is lost when gold human context is replaced with model-generated context.

As a proof-of-concept post-training use case, we also apply supervised fine-tuning (SFT) to \llamabase{}. SFT uses the same next-message prediction format as the simulation framework: each training example pairs an agent memory state \(\mathcal{M}_{a_i,k}\) with the corresponding human-written response observed in that context, and the model is optimized to predict that response. We evaluate the resulting SFT model under both group-generalization setting with unseen participant groups and topic-generalization setting with unseen topics. Details about SFT are in Appendix~\ref{app:sft}.

We evaluate seven LLMs spanning proprietary and open-weight models. Appendix~\ref{app:compute_resources} gives compute details.\footnote{We set \texttt{reasoning\_effort=medium} for gpt-5-mini.}

\section{Evaluating Opinion-Dynamics Fidelity}
\label{sec:opinion_dynamics}

Unless otherwise noted, all results and findings in this section are reported on the group-generalization test split. We summarize the cross-model topic-generalization replication below; Appendix~\ref{app:sft} reports its full results and further topic-generalization counterparts.

\subsection{Evaluation Metrics}
\label{subsec:eval}

We evaluate RPLA simulations primarily through \emph{stance-based opinion-dynamics fidelity}. This emphasis is deliberate: classic scalar-opinion agent-based models such as DeGroot abstract away from surface form and track only how agents' stances evolve over time. To stay aligned with that literature, our main metrics operate on public tweet stance and private reported opinion rather than on lexical similarity alone. \textit{Full Dynamics Simulation} is our main evaluation setting because agents must continue the discussion using their own earlier generations, which aligns with the recursive rollouts studied in classic opinion-dynamics work \citep{flache2017models,lorenz2021individual,taubenfeld2024systematic}. \textit{Next Message Prediction} is an easier best-case baseline because it conditions on human-written history. The conversation is therefore the mechanism through which agents reveal and revise their opinions, not a separate surface-form evaluation target.

For participant--agent pair $i$, opinion type $c \in \{\mathrm{pub},\mathrm{priv}\}$, and model $m$ (with $h$ denoting the human discussion), let $x^{c,0}_i(m)$ and $x^{c,1}_i(m)$ denote the start and end stance. Public states are derived from the first and third tweets, and private states from the initial and final reported opinions. Let $\mathcal{A}^{c}_m$ and $\mathcal{G}^{c}_m$ be the participant and group sets with complete observations for opinion type $c$. Figure~\ref{fig:heldout_breadth_group_posttrain_comparison} reports three stance-based errors for public and private opinions.

All displayed metrics below are defined for $c \in \{\mathrm{pub},\mathrm{priv}\}$. We report three stance-based errors: \emph{single-opinion prediction error}, \emph{opinion-update prediction error}, and \emph{opinion-convergence prediction error}. For single-opinion prediction, let $\mathcal{O}^{c}_m$ contain the valid, aligned human--model opinion pairs, including the tweets in rounds 2 and 3 for public opinions, and final private reports for private opinions. The first tweet and initial private report are excluded because it is provided as context. Let $z^c_j(m)$ and $z^c_j(h)$ denote the model and human stances in pair $j$.
\begin{equation}
E^{c}_{\mathrm{single}}(m)
= \frac{1}{|\mathcal{O}^{c}_m|}
\sum_{j \in \mathcal{O}^{c}_m}
\big|z^{c}_j(m)-z^{c}_j(h)\big|.
\label{eq:single-opinion-error}
\end{equation}

\begin{equation}
E^{c}_{\mathrm{update}}(m)
= \frac{1}{|\mathcal{A}^{c}_m|}
\sum_{i \in \mathcal{A}^{c}_m}
\Big|\big(x^{c,1}_i(m)-x^{c,0}_i(m)\big)-\big(x^{c,1}_i(h)-x^{c,0}_i(h)\big)\Big|.
\label{eq:opinion-update-error}
\end{equation}

For convergence, let
\begin{equation}
\Delta \mathrm{SD}^{c}_g(m)
= \mathrm{SD}\big(\{x^{c,1}_i(m): i \in g\}\big)
- \mathrm{SD}\big(\{x^{c,0}_i(m): i \in g\}\big).
\label{eq:group-sd-change}
\end{equation}

If $\mathcal{G}^{\mathrm{pub}}_m$ and $\mathcal{G}^{\mathrm{priv}}_m$ denote the groups with complete public and private observations, then
\begin{equation}
E^{c}_{\mathrm{conv}}(m)
= \frac{1}{|\mathcal{G}^{c}_m|}
\sum_{g \in \mathcal{G}^{c}_m}
\big|\Delta \mathrm{SD}^{c}_g(m)-\Delta \mathrm{SD}^{c}_g(h)\big|.
\label{eq:opinion-convergence-error}
\end{equation}
Lower values are better for all three stance-based error metrics in Fig.~\ref{fig:heldout_breadth_group_posttrain_comparison}. To contextualize scale, that figure also includes simple lower- and upper-bound baselines together with a fitted DeGroot comparator. For the opinion convergence analysis in Fig.~\ref{fig:heldout_breadth_group_posttrain_group_opinion_convergence}, we additionally report raw average $\Delta \mathrm{SD}$ values with standard errors, together with the One-Step Toward First Partner and 50\% SD Reduction reference targets. Appendix~\ref{app:degroot_baseline} gives the exact baseline constructions and fitting details.

\subsection{Finding 1: SFT Improves Opinion-Dynamics Fidelity Across Model Families, but Full-Dynamics Simulation Remains the Harder Setting}

\begin{figure*}[tb!]
\centering
\includegraphics[width=\linewidth]{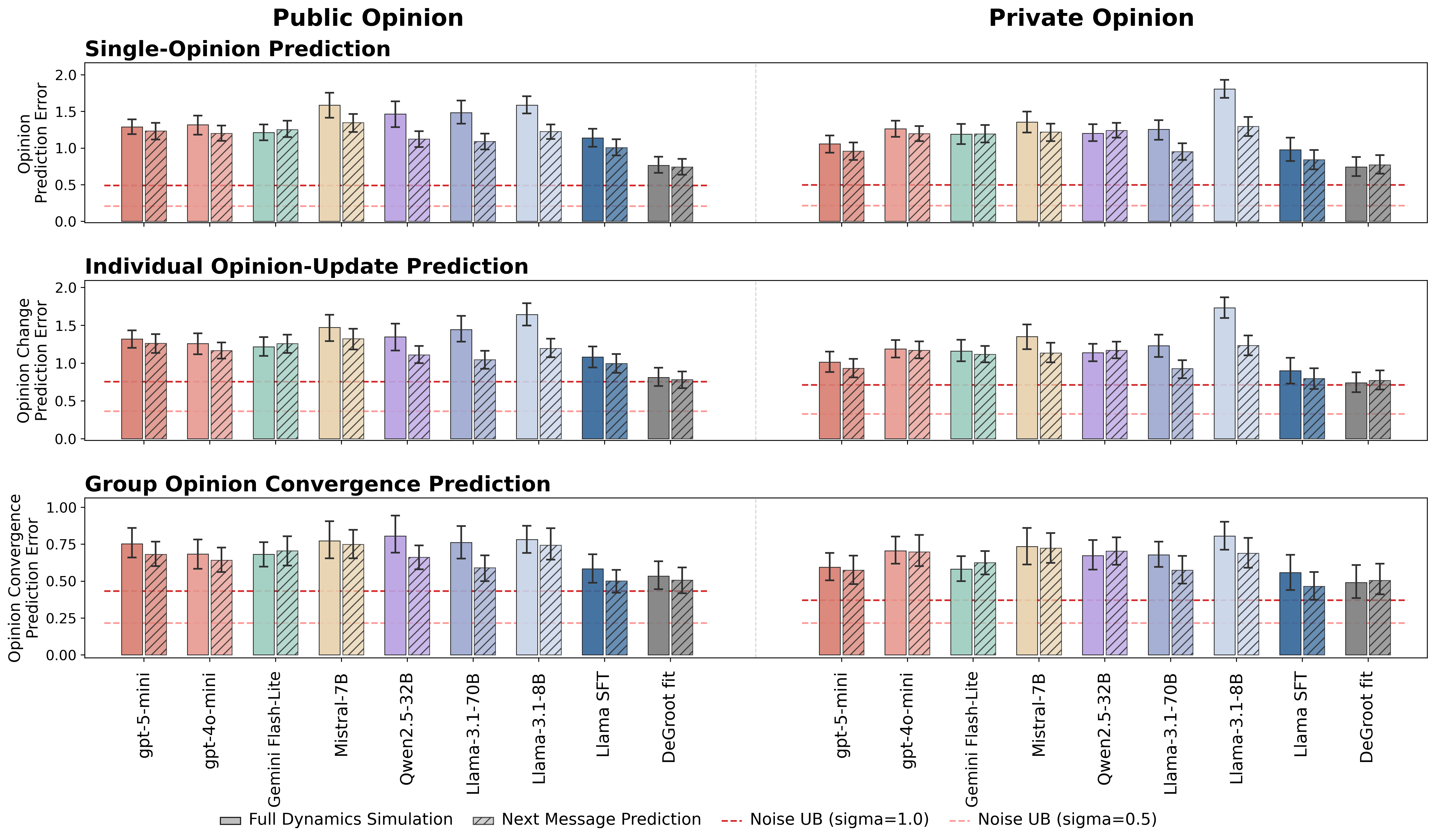}

\caption{SFT comparison across the three stance-based error metrics on the group-generalization test split. Rows show $E^{c}_{\mathrm{single}}$, $E^{c}_{\mathrm{update}}$, and $E^{c}_{\mathrm{conv}}$; the left column is public and the right column is private. Solid bars denote Full Dynamics Simulation, hatched bars denote Next Message Prediction, and lower values are better. Error bars are bootstrap 95\% confidence intervals; the reference baselines are defined in Sec.~\ref{subsec:eval}. Appendix~\ref{app:degroot_baseline} gives DeGroot-fitting details.}
\label{fig:heldout_breadth_group_posttrain_comparison}
\end{figure*}

Figure~\ref{fig:heldout_breadth_group_posttrain_comparison} summarizes SFT performance across all three stance-based metrics for \llamabase{}. The broad pattern is that SFT improves over the base model on single-opinion prediction, opinion-update prediction, and opinion-convergence prediction in both settings, while errors remain consistently larger in \textit{Full Dynamics Simulation}, our primary setting, than in the secondary \textit{Next Message Prediction} baseline. This gap shows that recursively generated interaction histories still degrade stance fidelity, even after SFT. On the other hand, the classic numerical agent-based model DeGroot mdoel performs well in terms of opinion-convergence error in \textit{Full Dynamics Simulation}, although it cannot generate language. Appendix~\ref{app:main_finding_stats} reports supporting statistical tests.
These conclusions are robust to judge choice; Appendix~\ref{app:judge_robustness} reports the same qualitative pattern under direct human labeling, and Appendix~\ref{app:llm_eval_classifier} shows that it is preserved under Gemini-3-pro-preview as a second LLM judge.

We also replicate the SFT experiments with \texttt{Qwen3.5-4B} (with thinking disabled) and \texttt{Mistral-7B-Instruct-v0.3}. For all three models, SFT aligns their opinion dynamics closer to human across public and private opinions, on both the group-generalization setting with seen topics and the topic-generalization setting with unseen topics. Appendix~\ref{app:sft-cross-model} reports the detailed results of SFT model's improvement across all three stance-based metrics.

\begin{figure*}[tb!]
\centering
\includegraphics[width=\linewidth]{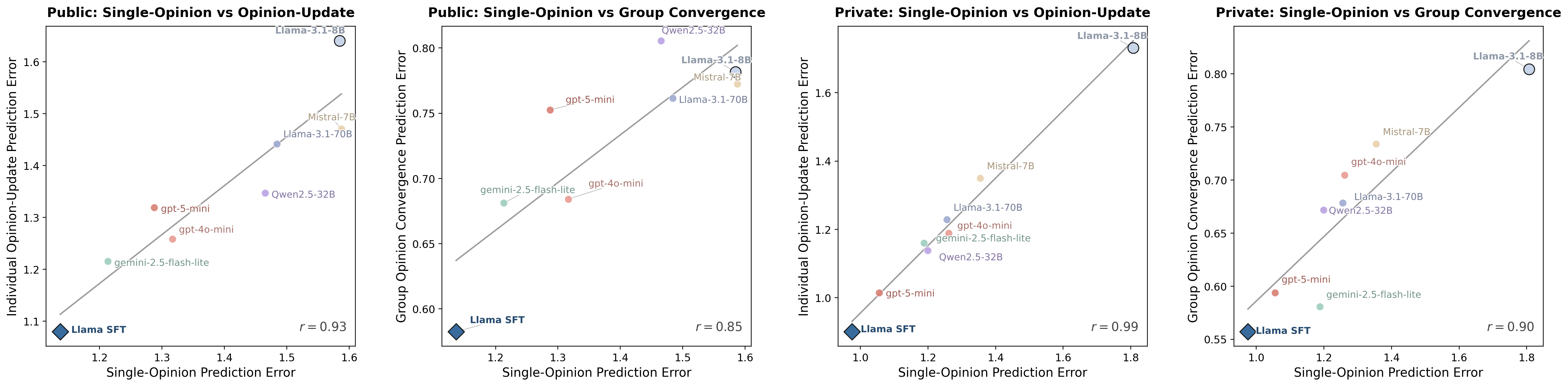}

\caption[Model-level correlations among the group-generalization test-set stance-based error metrics.]{Model-level correlations among the three stance-based error metrics on the group-generalization test split under Full Dynamics Simulation. From left to right, the scatterplots compare single-opinion prediction error against opinion-update and opinion-convergence prediction error for public and private opinions. Each point is one model, and lower-left positions indicate lower error on both compared metrics. Appendix Fig.~\ref{fig:heldout_breadth_topic_posttrain_metric_correlations} shows the same analysis on the topic-generalization split.}
\label{fig:heldout_breadth_group_posttrain_metric_correlations}
\end{figure*}

Figure~\ref{fig:heldout_breadth_group_posttrain_metric_correlations} helps interpret the results in Figure~\ref{fig:heldout_breadth_group_posttrain_comparison} by showing how the three stance-based errors co-vary across models in \textit{Full Dynamics Simulation}. The association is very strong between single-opinion prediction and opinion-update prediction in both public tweets and private opinions. The relation with opinion-convergence prediction is also positive, albeit weaker. These patterns indicate that the three metrics are aligned but not redundant: single-opinion prediction is a strong leading indicator of opinion-dynamics fidelity, but group-level convergence still contributes distinct information that motivates the detailed convergence analysis in Figure~\ref{fig:heldout_breadth_group_posttrain_group_opinion_convergence}.

\subsection{Finding 2: Base Models Over-Converge Relative to Human Groups, and SFT Partially Restores Human-Like Resistance to Convergence}

\begin{figure*}[tb!]
\centering
\includegraphics[width=\linewidth]{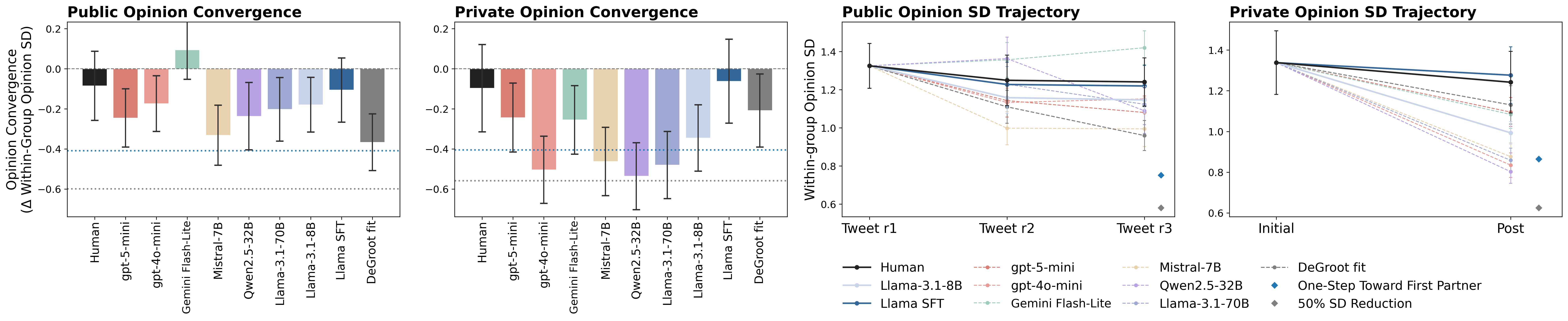}

\caption[SFT opinion convergence summaries and SD trajectories on the group-generalization test split.]{SFT results of opinion convergence and stance-SD trajectories on the group-generalization test split under \textit{Full Dynamics Simulation}. The left two panels report public and private changes in mean stance SD ($\Delta\mathrm{SD}$); the right two show the matching stance-SD trajectories. More negative bars indicate stronger convergence. Error bars denote standard errors, and the reference baselines are defined in Sec.~\ref{subsec:eval}. Appendix~\ref{app:degroot_baseline} gives DeGroot-fitting details, and Appendix~\ref{app:posttrain_histograms} shows the same convergence summary on the topic-generalization split}
\label{fig:heldout_breadth_group_posttrain_group_opinion_convergence}
\end{figure*}

\begin{figure*}[tb!]
\centering
\includegraphics[width=\linewidth]{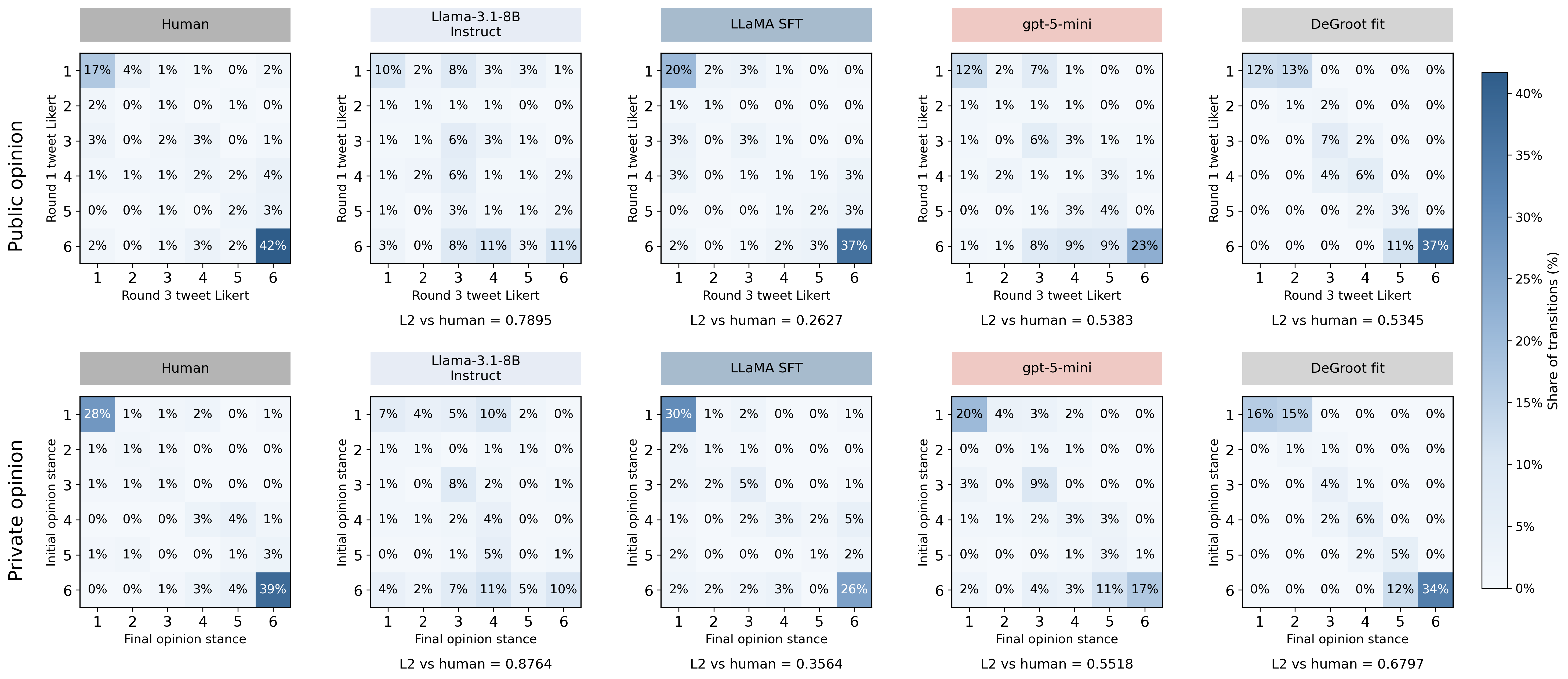}

\caption[Public/private transition matrices on the group-generalization test split.]{Public/private transition matrices on the group-generalization test split under \textit{Full Dynamics Simulation}. The top row shows public transitions from $S(\tau^1)$ to $S(\tau^3)$, and the bottom row shows private transitions from $o^{\text{init}}$ to $o^{\text{final}}$. Each cell shows the percentage of all observed transitions within that transition matrix. Diagonal mass indicates persistence, off-diagonal mass indicates movement across stance bins, and the annotated normalized $L_2$ distances summarize deviation from the human reference. Appendix~\ref{app:posttrain_histograms} shows the same transition-matrix analysis on the topic-generalization split.}
\label{fig:heldout_breadth_group_public_private_transition_matrices}
\end{figure*}

Prior work on multi-agent debate highlights excessive convergence as a central failure mode \citep{chuang2024simulating,taubenfeld2024systematic}, so we next zoom in on opinion convergence. Figure~\ref{fig:heldout_breadth_group_posttrain_group_opinion_convergence} reports SFT convergence summaries and stance-SD trajectories, showing how strongly groups contract over time; Fig.~\ref{fig:heldout_breadth_group_public_private_transition_matrices} reports public/private transition matrices, showing how that convergence is distributed across stance bins. We focus on the \llamabase{} SFT comparison.

In Figure~\ref{fig:heldout_breadth_group_posttrain_group_opinion_convergence}, the first two panels report changes in mean within-group stance SD for public and private opinions, where more negative values indicate stronger within-group opinion convergence; the last two panels show the corresponding stance-SD trajectories.

The qualitative pattern is consistent in both public tweets and private opinions: \llamabase{} over-converges relative to humans, and SFT moves the model much closer to the human opinion-convergence pattern. For public opinions, human convergence is modest ($-0.08 \pm 0.17$), whereas \llamabase{} converges more strongly at $-0.18 \pm 0.14$ and the fitted DeGroot baseline is even more convergent at $-0.37 \pm 0.14$. Critically, SFT reduces the Llama model's excessive convergence, shifting $\Delta\mathrm{SD}$ from $-0.18$ to $-0.11$. For private opinions, the same pattern holds, with the base model showing stronger opinion convergence than humans ($-0.34$ versus $-0.10$) and SFT reducing that excessive convergence to $-0.06$. The cross-model results in Appendix~\ref{app:sft-cross-model} show the same pattern for the two additional model families on both held-out groups and held-out topics.
The same over-convergence direction holds under additional spread metrics, and the convergence-robustness analyses in Appendix~\ref{app:convergence_robustness} indicate that the effect is not primarily a decoding artifact but is sensitive to removing explicit stance memory.

Figure~\ref{fig:heldout_breadth_group_public_private_transition_matrices} shows the same pattern from a different angle by tracking how probability mass moves across stance bins from the start to the end of the discussion. Humans retain more mass at the extreme bins, whereas the base model shifts too much public and private mass inward; SFT partially restores the human transition pattern. The annotated normalized $L_2$ distances summarize this improvement: for \llamabase{}, the distance from the human transition matrix is $0.79$ in public and $0.88$ in private, whereas SFT lowers it to $0.26$ and $0.36$. Appendix~\ref{app:main_finding_stats} reports supporting statistical tests. Appendix~\ref{app:posttrain_histograms} extends these histogram and transition-matrix views to the topic-generalization split and broader model comparisons. Residual mismatch remains, especially in private opinions, plausibly because those opinions are only indirectly constrained by the dialogue history.

\subsection{Finding 3: Over-Convergence Is Linked to Over-Responsiveness to Partner Stance}

\begin{figure*}[tb!]
\centering
\includegraphics[width=\linewidth]{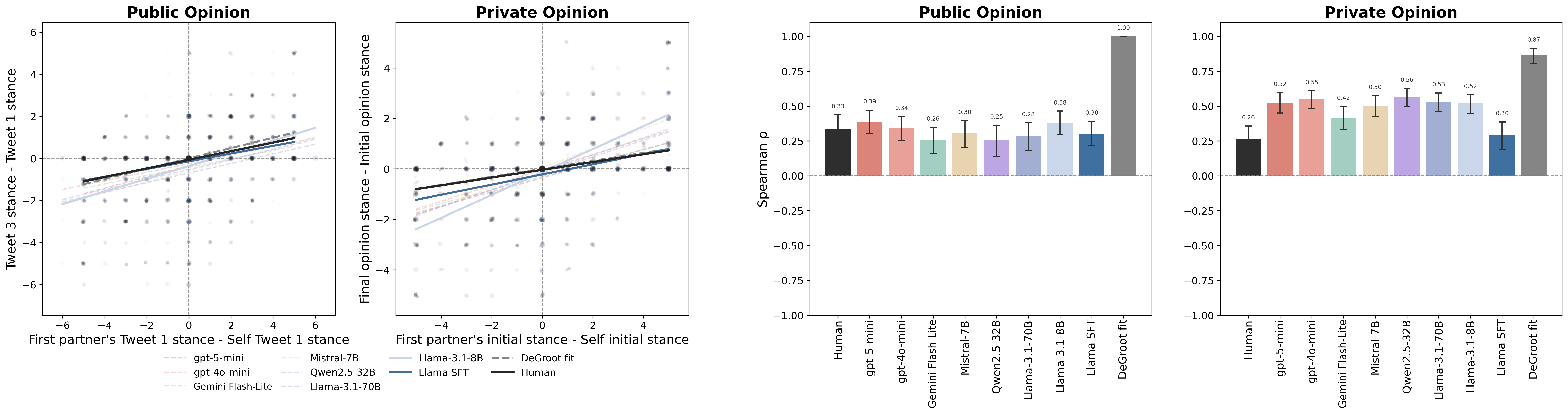}

\caption[Partner-gap responses on the group-generalization test split.]{Partner-gap responses on the group-generalization test split under \textit{Full Dynamics Simulation}. Left: changes in public and private stance versus the first partner's initial stance gap. Right: Spearman $\rho$ (larger positive values indicate stronger monotonic association); error bars show bootstrap 95\% confidence intervals. See Appendices~\ref{app:degroot_baseline} and~\ref{app:posttrain_histograms} for DeGroot fitting and topic-generalization results.}
\label{fig:heldout_breadth_group_partner_gap_scatter}
\end{figure*}

Figure~\ref{fig:heldout_breadth_group_partner_gap_scatter} tests one possible underlying mechanism by asking whether participants and agents move in the direction of the partner-stance gap, to assess whether over-convergent models are overly responsive to local interaction signals before aggregation to the group level.

Figure~\ref{fig:heldout_breadth_group_partner_gap_scatter} shows that humans tend to move toward their partner's stance in both public tweets and private opinions. The base model shows the same tendency but responds too strongly, consistent with its excessive convergence. In contrast, SFT keeps this directional response while moving it closer to the human pattern. Spearman correlations and their bootstrap confidence intervals quantify this pattern across models.

\subsection{Finding 4: Beyond Stance, DEBATE Supports Text-Level Evaluation}

Beyond the stance-only setting, RPLA simulations also support text-alignment evaluation of public tweets and private opinions, which classic numerical agent-based models such as DeGroot do not support. In \textit{Full Dynamics Simulation} (Appendix~\ref{app:beyond_stance}), the SFT model reaches public semantic similarity $0.76$ and public ROUGE-L $0.45$, compared with $0.71$ and $0.41$ for the base model. For the private post-discussion opinion target, the base model overshoots the human output by $49.00$ words on average, compared with $9.08$ words for SFT, while private ROUGE-L is $0.11$ for SFT versus $0.09$ for the base model. This matters because stance alignment alone could in principle be achieved by a naive model that emits strongly worded but low-quality text that still lands in the correct stance bin. A qualitative example from a SFT run (Appendix~\ref{app:example_chats}; condensed and paraphrased here) illustrates this distinction: the human begins at \textit{Stance 1} with \textit{``Mainstream news does not do a good job. It is biased, divisive, billionaire-owned, and not trustworthy,''} first tweets \textit{``I don't trust major news organizations; they are owned by billionaires and full of bias,''} and in Round 3 still tweets \textit{``They are very biased; only small independent outlets are reporting the truth.''} The base model instead swings to \textit{``I've had a change of heart: transparency, diverse perspectives, and independent fact-checkers could balance profit and truthful storytelling''} and ends at \textit{Stance 6}, whereas the SFT model stays much closer to the human trajectory with \textit{``Major mainstream outlets push a billionaire-controlled agenda; local papers may offer better factual reports''} and ends at \textit{Stance 1}: \textit{``I still disagree that mainstream news organizations are doing a good job; bias and misinformation make trustworthy reporting hard to find.''} Appendix~\ref{app:on_topic_rate_diagnostics} summarizes the companion topicality checks, and Appendix~\ref{app:main_finding_stats} reports supporting statistical tests.

\section{Conclusion}
\label{sec:conclusion}

We introduced \textbf{DEBATE}, an empirical benchmark for evaluating opinion dynamics in multi-agent role-playing LLM agents (RPLAs). By capturing traceable public interaction histories together with initial and final private opinions from 2,788 participants in 697 four-person groups, DEBATE supports evaluation at 3 levels:  single-opinion prediction, opinion-update prediction, and group-level opinion convergence. The results show a consistent pattern: zero-shot RPLAs over-converge and quickly agree with one another relative to humans, and supervised fine-tuning improves alignment with human opinion dynamics in terms of the stance-based metrics. We also show that recursively generated \textit{Full Dynamics Simulation} remains harder than conditioning on human history. In sum, DEBATE is a benchmark for making clearer evaluative claims about which short-horizon, small-group opinion-dynamics behaviors RPLAs match and where they still fail.

\clearpage
\begin{ack}
This work was supported by the Department of Defense Multi University Research Initiative grant W911NF2110317 (Rogers, Co-I); a University of Wisconsin--Madison Research Forward award, funded by the Wisconsin Alumni Research Foundation (MSN282355; Rogers, PI); the University of Wisconsin--Madison Office of the Vice Chancellor for Research with funding from the Wisconsin Alumni Research Foundation; the School of Journalism \& Mass Communication at the University of Wisconsin--Madison through the Journal Communications/Warren J.~Heyse Faculty Excellence Award (Sijia Yang); a grant from the John S. and James L. Knight Foundation (AWD-002111; Shah, co-PI); and the National Science Foundation under Award \#IIS-2449768. The views and conclusions expressed in this work are those of the authors and should not be interpreted as representing the official policies or endorsements of the U.S. Government, or the National Science Foundation.

The authors declare no competing interests.

We are grateful to Aditya Dharap for assistance with human annotations of on-topicness and stance, for verifying part of our evaluation pipeline, and for facilitating access to Gemini models.
\end{ack}

\iffastbuild
\else
\clearpage
\bibliography{custom}
\bibliographystyle{abbrvnat}

\appendix
\newpage
\onecolumn
\appendix
\label{sec:appendix}

\section{Code and Data Availability}
\label{app:supp}

The full DEBATE dataset is publicly available at \url{https://huggingface.co/datasets/seantw/DEBATE_LLM}, the codebase at \url{https://github.com/yunshiuan/DEBATE-neurips2026}, and the benchmark website at \url{https://debate-llm-benchmark.vercel.app/}. The codebase contains five integrated pipelines: (1) preprocessing for data standardization, (2) simulation for generating LLM-based conversations across different modes, (3) evaluation for comparing human and LLM outputs using similarity metrics and belief trajectories, (4) group-level statistical analysis, and (5) model fine-tuning.

The code supports both simulation setups (full dynamics simulation and next message prediction) and includes prompt templates, evaluation scripts, and detailed documentation for reproducing the main figures and tables based on cached LLM generations and labels.

\clearpage
\section{Comparison with Existing Opinion-Dynamics Datasets}
\label{app:dataset_comparison}

\paragraph{Existing opinion-dynamics corpora.}
To our knowledge, no prior corpus was constructed to benchmark whether multi-agent RPLA simulations reproduce matched human opinion trajectories through interaction. Existing corpora with opinionated human interactions can be grouped into three genres (Table~\ref{tab:dataset_comparison}): \emph{competitive debate}, in which participants seek to win an argument (e.g., IAC); \emph{asymmetric persuasion}, in which one or more participants seek to change a designated opinion holder's view (e.g., ChangeMyView (CMV) and PersuasionForGood); and \emph{naturalistic deliberation}, in which peers voluntarily share and refine beliefs without assigned adversarial or persuasive roles \citep{walton1995commitment,walton2010argumentation,bozdag2025must}. The latter is especially relevant for everyday social interaction, but is relatively underrepresented in existing corpora.

Beyond genre coverage, many existing corpora lack one or more properties needed to benchmark human-like opinion dynamics. A reply tree alone does not establish which prior messages a participant read before responding, so reader-context traceability is often unavailable; even in CMV, attention to the available content is modeled as a latent process \citep{ojer2026modeling}. Open social-media participation is also voluntary and uneven. In a large Reddit sample, the top 1\% of users produced 25\% of comments and the top 10\% produced 68\%, while 27\% posted only once \citep{vuorio2023lurking}. Observed trajectories can therefore overrepresent highly active contributors while silent readers' beliefs remain unobserved \citep{van20141}. Most corpora further lack enforced turn-taking, demographic attributes, or private pre- and post-discussion beliefs for every participant. These private measurements matter because publicly expressed views can differ from self-reported beliefs \citep{tourangeau2007sensitive}.

\paragraph{Observational data versus DEBATE.}
Observational social-media corpora such as CMV provide substantially greater scale and remain valuable for studying naturalistic persuasion. We do not claim that DEBATE is uniformly better. Rather, DEBATE trades interaction-network scale for controlled, identifiable trajectories: each retained participant completes the protocol, supplies private pre- and post-discussion beliefs, contributes public opinion content at each required stage, and receives a known public interaction history before responding. In contrast, CMV's delta records an opinion-change signal from the original poster, not private belief trajectories for every participant in a thread \citep{tan2016winning}. Thus, observational and experimental data support complementary evaluations: large observational corpora are well suited to studying naturally occurring participation and persuasion at scale, whereas DEBATE enables matched, group-level tests of whether RPLAs reproduce the observable trajectories of every participant under a shared protocol.

Table~\ref{tab:dataset_comparison} summarizes this quality--scale tradeoff. Reader-context traceability denotes whether the content presented to each participant before a response can be reconstructed; turn obligation denotes whether the protocol requires each participant to contribute, rather than observing only voluntary replies.

\begin{table}[!htbp]
\caption{Comparison of DEBATE with existing human opinion-dynamics datasets. We categorize datasets into three genres: \emph{competitive debate}, \emph{asymmetric persuasion}, and \emph{naturalistic deliberation}. Reader-context traceability indicates whether the content presented to each participant before a response can be reconstructed; turn obligation indicates whether the protocol requires each participant to contribute.}
\centering
\Large
\resizebox{0.95\linewidth}{!}{%
\begin{tabular}{p{6.5cm}p{3cm}p{1.2cm}p{1cm}p{1.6cm}p{2.1cm}p{2cm}p{2.2cm}p{5.8cm}p{4cm}}
\toprule
Dataset & \makecell[tl]{Opinion \\ Dynamics \\ Type} & Multi-Party (N$\geq$3) & Multi-Turn & \makecell[tl]{Demo-\\graphics} & Reader-Context Traceable & Turn Obligation & \makecell[tl]{Public or \\ Private \\ Opinion} & Data Source \newline [Participants] & \makecell[tl]{\# Utterances (U) \\ \# Conversations (C) \\ \# Subjects (S)} \\
\midrule
CMV \citep{tan2016winning} & Persuasion (Asymmetric) & \cmark & \cmark & \xmark & \xmark & \xmark & Public & Reddit threads \newline [Reddit users] & U = 293,297 \newline C = 3,051 \newline S = 34,911 \\
IAC 1.0 \citep{walker2012iac} & Competitive Debate & \cmark & \cmark & \xmark & \xmark & \xmark & Public & Online debate forums \newline [Online forum users] & U = 390,704 \newline C = 11,800 \newline S = 3,300 \\
IAC 2.0 \citep{abbott2016internet} & Competitive Debate & \cmark & \cmark & \xmark & \xmark & \xmark & Public & Online debate forums \newline [Online forum users] & U = 482,000 \newline C = 16,461 \newline S = 9,709 \\
UK Parliament QuestionTime Corpus \citep{zhang2017_asking_too_much} & Competitive Debate & \cmark & \cmark & \xmark & \cmark & \cmark & Public & British House of Commons \newline [Members of Parliament] & U = 433,787 \newline C = 216,894 \newline S = 1,978 \\
Intelligence Squared Debates Corpus \citep{zhang2016_conversational_flow} & Competitive Debate & \cmark & \cmark & \xmark & \cmark & \cmark & Public & Structured debate show \newline [Invited experts] & U = 26,562 \newline C = 108 \newline S = 471 \\
PersuasionForGood \citep{wang2019persuasion} & Persuasion (Asymmetric) & \xmark & \cmark & \cmark & \cmark & \cmark & Public & MTurk platform \newline [Diverse laypeople] & U = 20,932 \newline C = 1,017 \newline S = 1,285 \\
Wikipedia Articles-for-Deletion (AfD) \citep{mayfield2019_wikipedia_deletion} & Deliberation (Policy) & \cmark & \cmark & \xmark & \xmark & \xmark & Public & Wikipedia editor debates \newline [Wikipedia editors] & U = 3,295,340 \newline C = 383,918 \newline S = 161,266 \\
\midrule
\textbf{DEBATE} & Deliberation (Naturalistic) & \cmark & \cmark & \cmark & \cmark & \cmark & Public \newline \textcolor{lightgreen}{+ Private} & Prolific platform \newline [\textcolor{lightgreen}{Diverse laypeople}] & U = 28,286 \newline C = 3,490 \newline S = 2,788 \\
\bottomrule
\end{tabular}%
}
\label{tab:dataset_comparison}
\end{table}

 \section{Benchmark Topics}
\label{appendix:topics}

The benchmark spans science, policy, democracy, religion, and social values, and topics were selected to elicit divergent human views among U.S. participants. Topics were drawn from the World Values Survey~\citep{haerpfer2022world}, the Pew Global Attitudes Survey~\citep{pew_2025}, and a study of human belief networks~\citep{chuang2024beyond}. We selected the topics with high response entropy so that people tend to hold diverse opinions about them. Examples include:

\begin{enumerate}
    \item A "body cleanse," in which you consume only particular kinds of nutrients over 1–3 days, helps your body to eliminate toxins.
    \item Angels are real.
    \item Everything that happens can eventually be explained by science.
    \item Regular fasting will improve your health.
    \item The U.S. deficit increased after President Obama was elected.
    \item The United States has the highest federal income tax rate of any Western country.
    \item The position of the planets at the time of your birth can influence your personality.
\end{enumerate}

Most original questions are already framed as evaluative statements rated on a Likert scale. For example:

\begin{itemize}
    \item \textbf{Original questions:} \\
    \textit{Please tell me for each of the following statements whether you think it can always be justified, never be justified, or something in between.} \\
    \textit{Euthanasia can always be justified.} (Presented along with a 10-point Likert scale.)
\end{itemize}

In these cases, we use the original statement directly as a debate topic (e.g., \textit{``Euthanasia can be justified.''}).

Some source questions, however, are framed in a multiple-choice format. To convert these into clearly debatable statements, we reframe the most frequently chosen responses as separate topic statements. For example:

\begin{itemize}
    \item \textbf{Original questions:} \\
    \textit{In your opinion, what is the most important problem facing this country today?} \\
    (Options: 
    \textbf{Economic problems (19.59\%)}, 
    Children and education (4.12\%), 
    Crime (3.09\%), 
    Health (4.12\%), 
    Housing (1.03\%), 
    People (11.34\%), 
    Politics (14.43\%), 
    \textbf{International affairs (36.08\%)}, 
    Science (1.03\%), 
    Others (5.15\%))

    \item \textbf{Reframed as two separate debatable topics:}
    \begin{itemize}
        \item \textit{International affairs is the most important problem facing the U.S. today.}
        \item \textit{Economic problems are the most important problem facing the U.S. today.}
    \end{itemize}
\end{itemize}

We adapt source questions to the political context at the time of human data collection. For instance:

\begin{itemize}
    \item \textbf{Original questions:} \\
    \textit{How confident are you that Joe Biden can make good decisions about the use of military force?}

    \item \textbf{Revised statement:} \\
    \textit{Donald J. Trump can make good decisions about the use of military force.}
\end{itemize}

These modifications ensure that all topics are relevant, interpretable, and debate-worthy, while remaining faithful to the spirit of the original questions. Each topic statement was manually reviewed to confirm that it is clearly phrased as a 1) self-contained declarative sentence, 2) framed in a way that invites disagreement, and 3) suitable for eliciting meaningful opinion exchanges in multi-party conversations.

The full topic list is included in the released dataset. Table~\ref{tab:representative-topics} provides representative examples, along with tentative category labels. These categories are introduced solely to help readers understand the topic diversity. They are not used in any part of our simulation, evaluation, modeling, or analysis.

\begin{table*}[!htb]
\centering
\caption{Categorization of representative benchmark topics used in our study.}
\resizebox{\linewidth}{!}{%
\begin{tabular}{p{4cm}p{14cm}}
\toprule
\textbf{Topic Category} & \textbf{Topic Statement} \\
\midrule

Governance \& Democracy &
A democratic system where citizens, not elected officials, vote directly on major national issues to decide what becomes law is a good way of governing the US. \\
& It is a characteristic of democracy for the state to make people's incomes equal. \\
& Living in a country that is governed democratically is important. \\
& The United States is being governed democratically today. \\
& The army taking over when the government is incompetent is a characteristic of democracy. \\

\midrule
Science \& Technology &
Science and technology are making our lives healthier, easier, and more comfortable. \\
& The world is better off because of science and technology. \\
& It is important for people to know about science in their daily life. \\
& We depend too much on science and not enough on faith. \\
& Because of science and technology, there will be more opportunities for the next generation. \\

\midrule
Morality \& Social Norms &
Sex before marriage can be justified. \\
& Suicide can be justified. \\
& Homosexuality can be justified. \\
& Abortion can be justified. \\
& Having casual sex can be justified. \\
& Violence against other people can be justified in some cases. \\

\midrule
\multirow[t]{5}{4cm}{Economic Inequality \& Social Mobility} 
&
Incomes should be made more equal. \\
& The growing gap between the rich and poor poses the greatest threat to the world. \\
& The fact that some people work harder than others is the most important reason for the gap between the rich and the poor in the United States. \\
& Knowing the right people is important for getting ahead in life. \\
& Belonging to a wealthy family is important for getting ahead in life. \\

\midrule
Media \& Trust in Institutions &
Journalists provide fair coverage of elections in the US. \\
& TV news favors the governing party in general. \\
& News organizations are doing well at reporting different positions on political issues fairly. \\
& There is abundant corruption in the United States. \\
& Most politicians in the United States are corrupt. \\

\midrule
\multirow[t]{5}{4cm}{International Relations \& Trade} &
Donald J. Trump can deal effectively with China. \\
& The North American Free Trade Agreement (NAFTA) has been good for the US. \\
& The United States benefits a lot from the World Health Organization. \\
& Overall, increased tariffs on imported goods from foreign countries are good for the US. \\
& International affairs is the most important problem facing the US today. \\

\midrule
Public Policy \& Government Role &
The government should take more responsibility to ensure that everyone is provided for, rather than leaving it to individuals. \\
& Public debt is the most important issue for the government to address first. \\
& The lack of employment opportunities is the most important issue for the government to address first. \\
& Government ownership of business should be increased. \\

\midrule
Religion \& Belief &
We depend too much on science and not enough on faith. \\
& Religious and ethnic hatred poses the greatest threat to the world. \\
& It is an essential characteristic of democracy for religious authorities to interpret the laws. \\

\midrule
US Identity \& Society &
Being born in the United States is important for truly being American. \\
& The United States has the best quality of universities. \\
& The United States is a place where a young person could lead a good life. \\
& I’m worried about a civil war in the United States. \\

\bottomrule
\end{tabular}
}
\label{tab:representative-topics}
\end{table*}

\section{Dataset Cleaning and Split Coverage}
\label{appendix:dataset-cleaning}

DEBATE includes both fully completed sessions and partial sessions created by participant dropouts or technical failures. The benchmark used in this paper keeps only groups that complete the full interaction protocol and contribute sufficient valid/on-topic content for evaluation. We use ``on-topic'' as shorthand for a broader message-level validity assessment, rather than topic relevance alone. Appendix~\ref{app:llm_eval_classifier} specifies the detailed definition and annotation procedure.

Table~\ref{tab:dataset-split-summary} summarizes the human data under each partition strategy. The tables below report message counts and group coverage.

When a participant exits mid-session, the conversation with that specific participant pauses with a waiting notice.  These partial transcripts remain in the released data but are excluded from the evaluation subset used here.

\begin{table}[tbp]
\caption{\textbf{Human data by split.} Total counts describe the train/test partitions; available counts describe the human data available for evaluation. On-topic messages satisfy the message-validity criteria (Appendix~\ref{appendix:dataset-cleaning}).}
\centering
\scriptsize
\resizebox{\linewidth}{!}{%
\begin{tabular}{llrrrrrr}
\toprule
\textbf{Split} & \textbf{Partition} & \multicolumn{2}{c}{\textbf{Topics}} & \multicolumn{2}{c}{\textbf{Groups}} & \textbf{Messages} & \textbf{On-topic messages} \\
 & & \textbf{Total} & \textbf{Available} & \textbf{Total} & \textbf{Available} & & \\
\midrule
Group-generalization & Train & 107 & 107 & 573 & 565 & 23,080 & 22,003 \\
Group-generalization & Test & 107 & 104 & 135 & 132 & 5,206 & 4,969 \\
\midrule
Topic-generalization & Train & 75 & 75 & 511 & 501 & 20,356 & 19,331 \\
Topic-generalization & Test & 32 & 32 & 197 & 196 & 7,930 & 7,641 \\
\bottomrule
\end{tabular}%
}
\label{tab:dataset-split-count-details}
\end{table}

\begin{table}[tbp]
\caption{\textbf{Group coverage for SFT evaluation.} Human counts describe groups with human data available for evaluation. SFT columns count groups with both human and model outputs, before applying metric-specific validity criteria.}
\centering
\scriptsize
\resizebox{\linewidth}{!}{%
\begin{tabular}{llrrrrr}
\toprule
\textbf{Split} & \textbf{Partition} & \textbf{Total groups} & \textbf{Human data} & \textbf{Missing human data} & \makecell{\textbf{SFT}\\\textbf{Full Dynamics}\\\textbf{Simulation}} & \makecell{\textbf{SFT}\\\textbf{Next Message}\\\textbf{Prediction}} \\
\midrule
Group-generalization & Train & 573 & 565 & 8 & 522 & 533 \\
Group-generalization & Test & 135 & 132 & 3 & 125 & 125 \\
\midrule
Topic-generalization & Train & 511 & 501 & 10 & 472 & 475 \\
Topic-generalization & Test & 197 & 196 & 1 & 184 & 178 \\
\bottomrule
\end{tabular}%
}
\label{tab:dataset-split-prefix-coverage}
\end{table}

The human-data columns count groups with observed human responses. The SFT columns count groups with both human and simulated responses. Missing data and the observations required by each metric determine the number of groups evaluated.

\subsection{Conversation-Level Quality Audit}

Message-level validity does not by itself establish that a full dialogue is coherent or responsive. We therefore independently audited 120 dyadic conversations from 20 randomly sampled groups. Two annotators familiar with the domain and dataset evaluated whether each exchange was coherent, responsive to the partner, and substantively engaged beyond filler. Table~\ref{tab:conversation-quality-audit} reports the percentage of conversations satisfying each criterion according to each annotator, along with their agreement.

\begin{table}[tbp]
\centering
\small
\caption{Conversation-level quality audit of 120 dyadic conversations from 20 randomly sampled groups.}
\label{tab:conversation-quality-audit}
\begin{tabular}{lccc}
\toprule
\textbf{Criterion} & \textbf{Annotator 1} & \textbf{Annotator 2} & \textbf{Agreement} \\
\midrule
Coherent & 100.0\% & 100.0\% & 100.0\% \\
Responsive & 99.2\% & 100.0\% & 99.2\% \\
Substantively engaged & 97.5\% & 98.3\% & 99.2\% \\
\bottomrule
\end{tabular}
\end{table}

The annotators agreed on 358 of 360 criterion-level judgments (99.4\% overall). This audit complements the message-level filter with direct evidence that the sampled dialogues generally constitute coherent, responsive, and meaningful exchanges.

\section{Demographic Summary}
\label{appendix:demographics}

Of the 2,832 total participants in our study, 2,640 (93.2\%) provided demographic responses. The resulting sample reflects substantial demographic diversity across multiple dimensions (Figure~\ref{fig:demographics}). Valid reported ages range from 18 to 84 (M = 39.7, SD = 13.0), and participants span a broad spectrum of gender identities, education levels, ethnic backgrounds, and income brackets. The cohort includes individuals with high school to doctoral-level education, and income levels range from under \$25k to over \$200k. Racial and ethnic diversity is well represented, with participants identifying as Black, Hispanic, White, Asian, Native American, and multiracial. Political identities and views are distributed across the ideological spectrum, and respondents report a wide variety of religious affiliations and Bible interpretations. Participants also vary in marital and parental status, geographic residence (urban, suburban, rural), and religious orientation (with 44.4\% of respondents identifying as evangelical). Occupation is similarly diverse, with respondents employed across sectors including finance, engineering, health care, education, manufacturing, media, construction, among many. This heterogeneity provides a rich foundation for studying opinion dynamics and belief-based interactions.

\begin{figure*}[htb!] 
    \centering

    \begin{subfigure}[t]{0.32\textwidth}
        \centering
        \includegraphics[width=\linewidth]{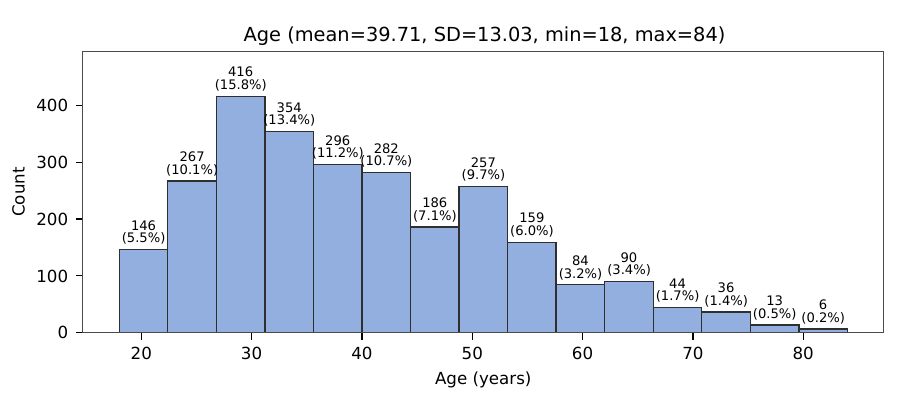}
        \caption{Age}
    \end{subfigure}
    \begin{subfigure}[t]{0.32\textwidth}
        \centering
        \includegraphics[width=\linewidth]{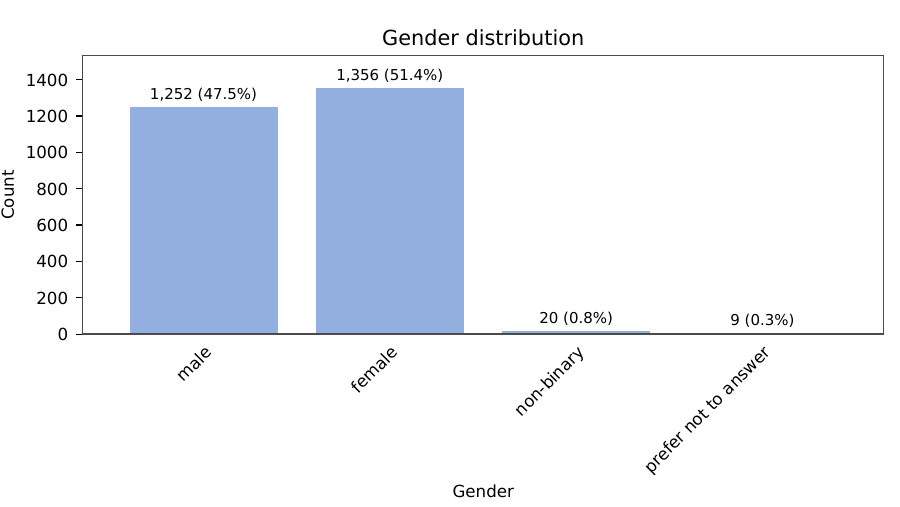}
        \caption{Gender}
    \end{subfigure}
    \begin{subfigure}[t]{0.32\textwidth}
        \centering
        \includegraphics[width=\linewidth]{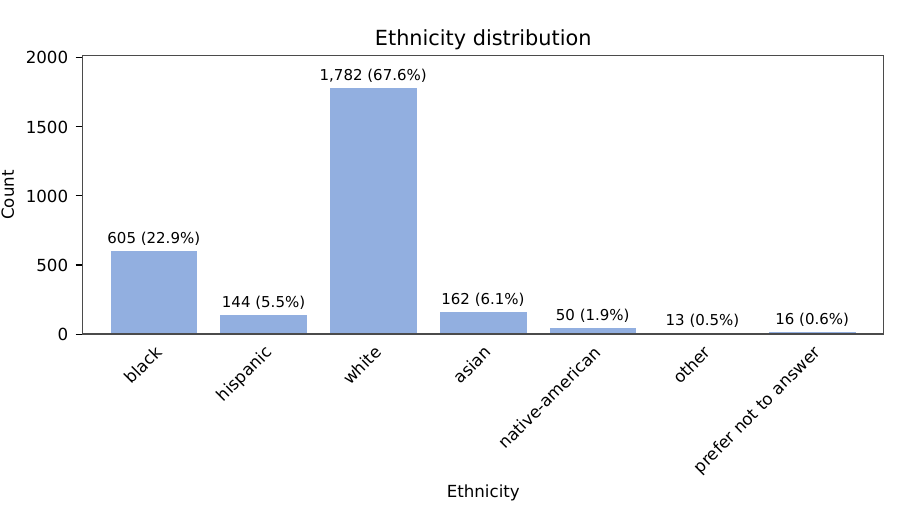}
        \caption{Ethnicity}
    \end{subfigure}

    \begin{subfigure}[t]{0.32\textwidth}
        \centering
        \includegraphics[width=\linewidth]{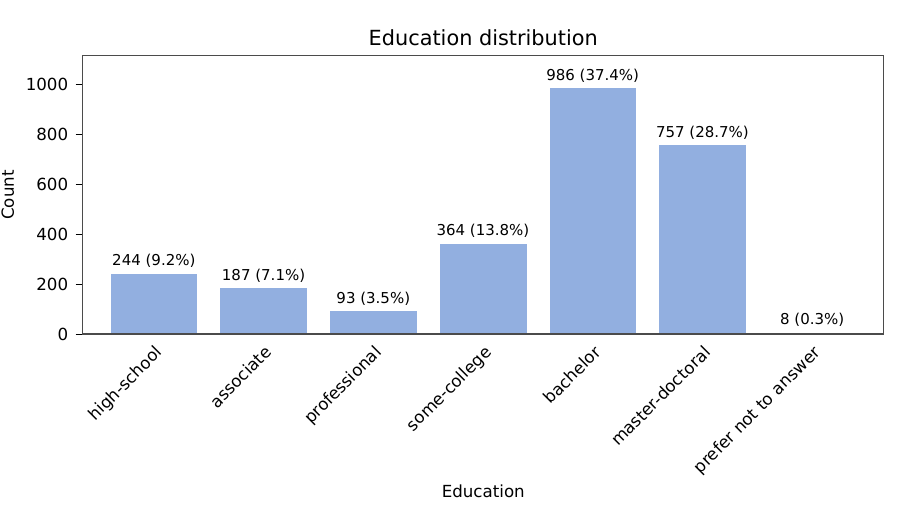}
        \caption{Education}
    \end{subfigure}
    \begin{subfigure}[t]{0.32\textwidth}
        \centering
        \includegraphics[width=\linewidth]{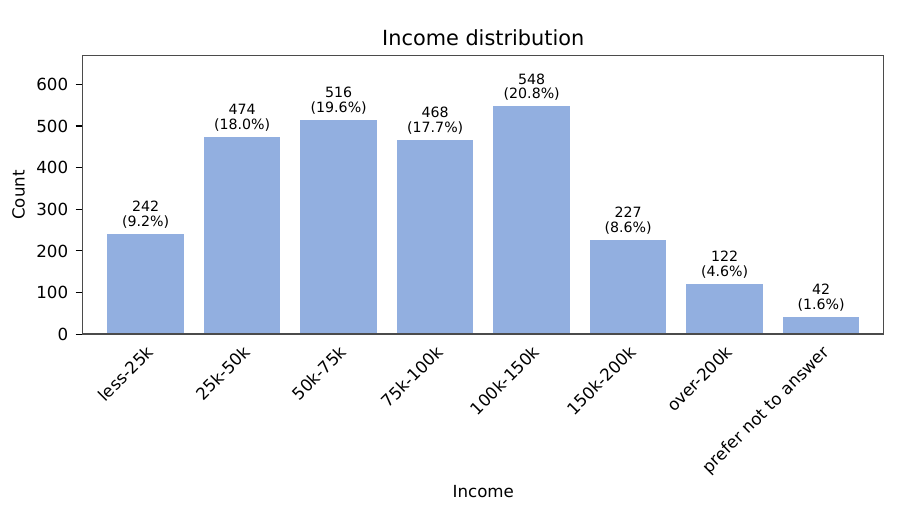}
        \caption{Income}
    \end{subfigure}
    \begin{subfigure}[t]{0.32\textwidth}
        \centering
        \includegraphics[width=\linewidth]{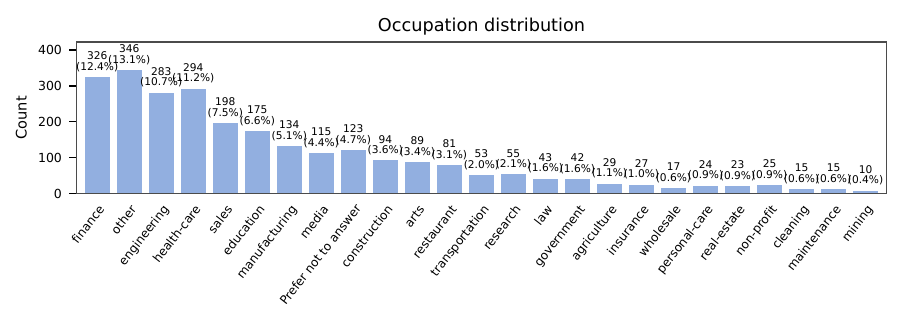}
        \caption{Occupation}
    \end{subfigure}

    \begin{subfigure}[t]{0.32\textwidth}
        \centering
        \includegraphics[width=\linewidth]{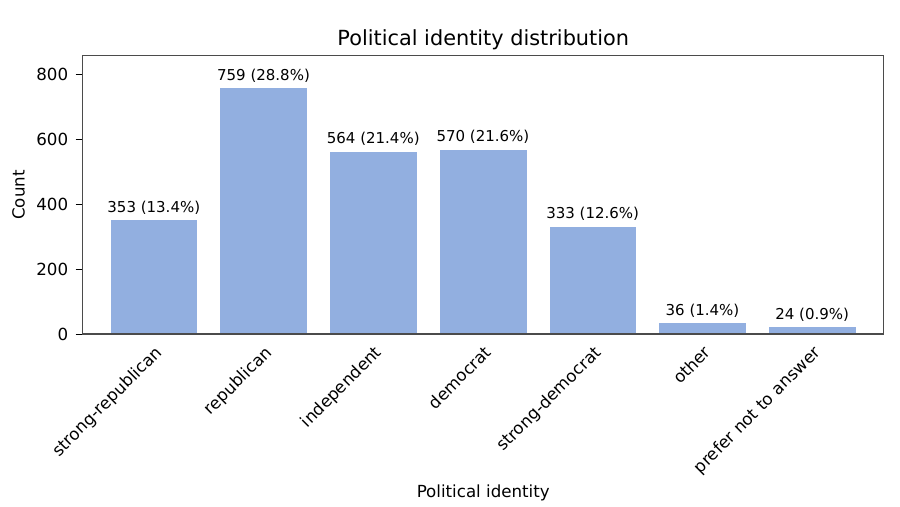}
        \caption{Political Identity}
    \end{subfigure}
    \begin{subfigure}[t]{0.32\textwidth}
        \centering
        \includegraphics[width=\linewidth]{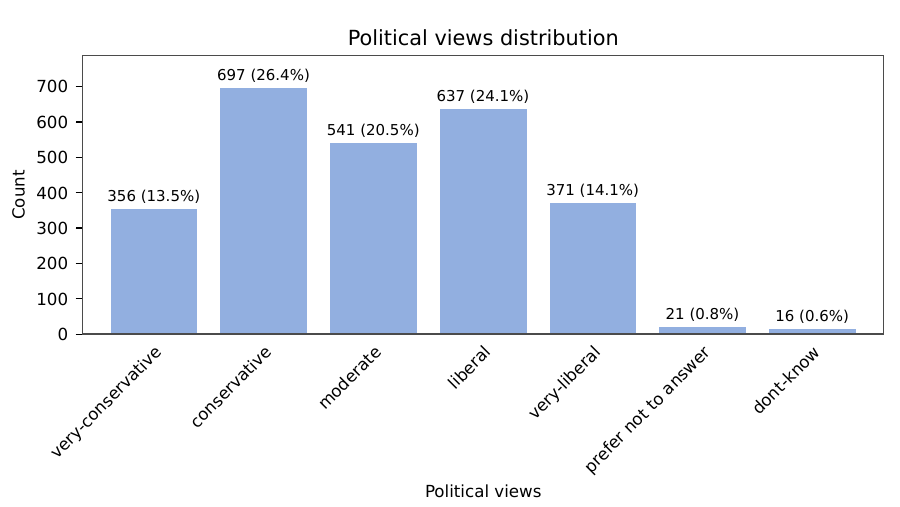}
        \caption{Political Views}
    \end{subfigure}
    \begin{subfigure}[t]{0.32\textwidth}
        \centering
        \includegraphics[width=\linewidth]{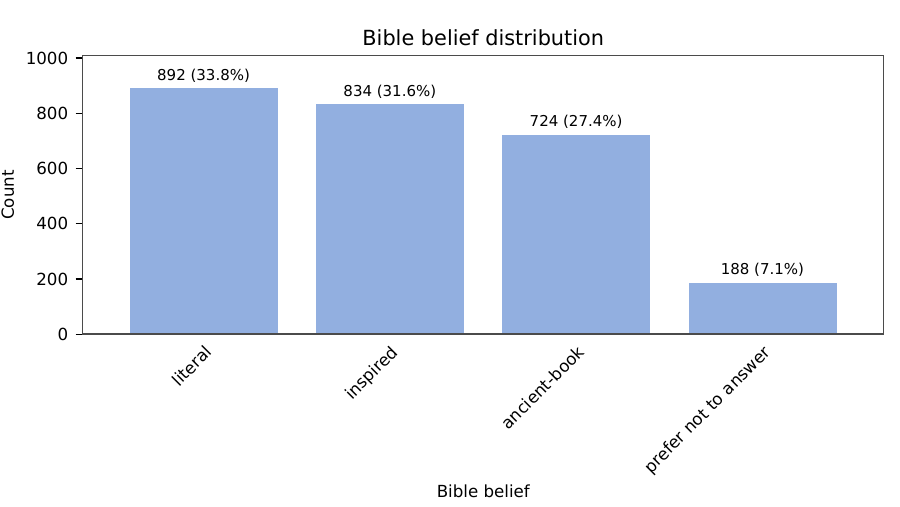}
        \caption{Bible Belief}
    \end{subfigure}

    \begin{subfigure}[t]{0.32\textwidth}
        \centering
        \includegraphics[width=\linewidth]{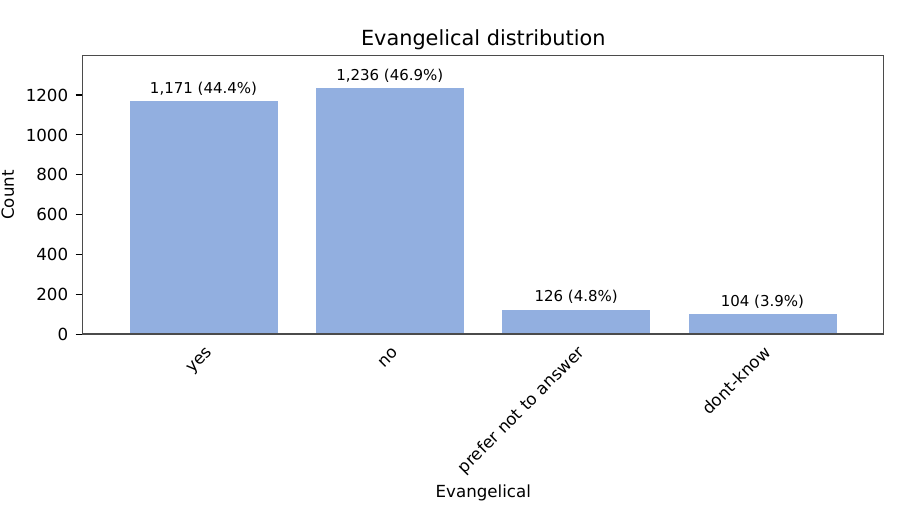}
        \caption{Evangelical}
    \end{subfigure}
    \begin{subfigure}[t]{0.32\textwidth}
        \centering
        \includegraphics[width=\linewidth]{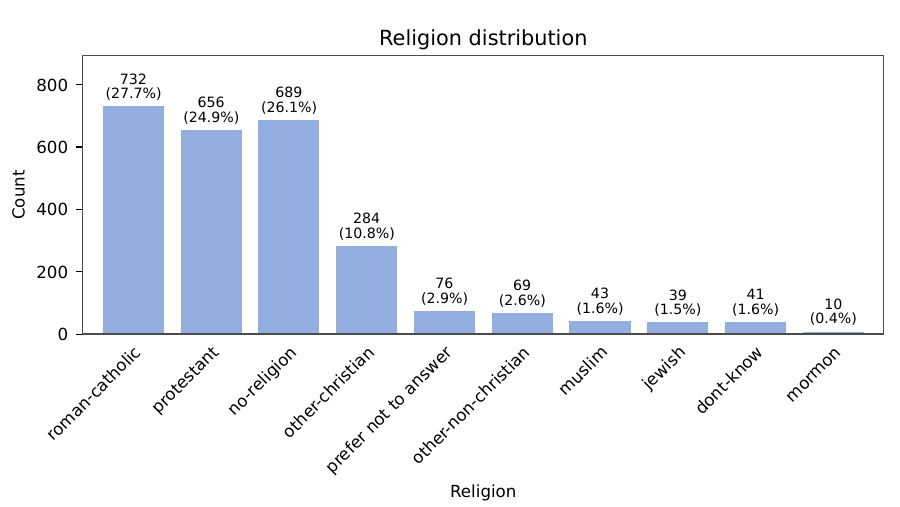}
        \caption{Religion}
    \end{subfigure}
    \begin{subfigure}[t]{0.32\textwidth}
        \centering
        \includegraphics[width=\linewidth]{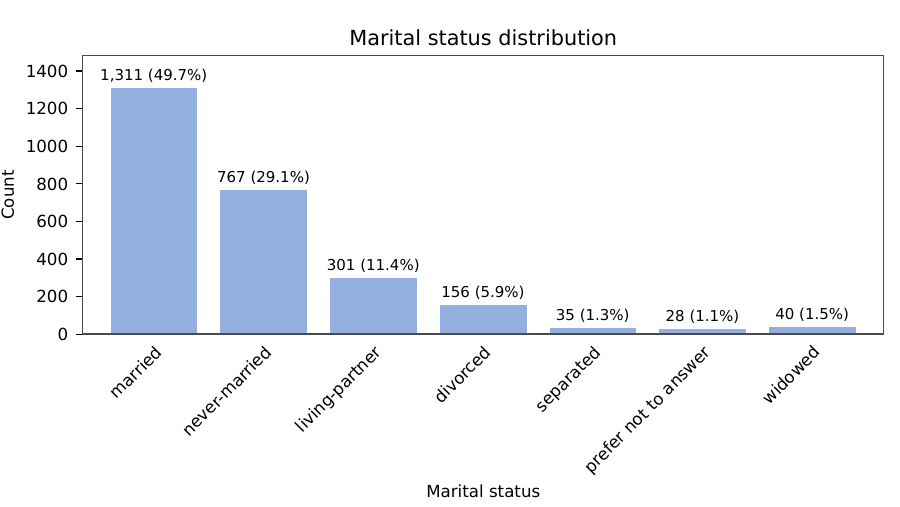}
        \caption{Marital Status}
    \end{subfigure}

    \begin{subfigure}[t]{0.32\textwidth}
        \centering
        \includegraphics[width=\linewidth]{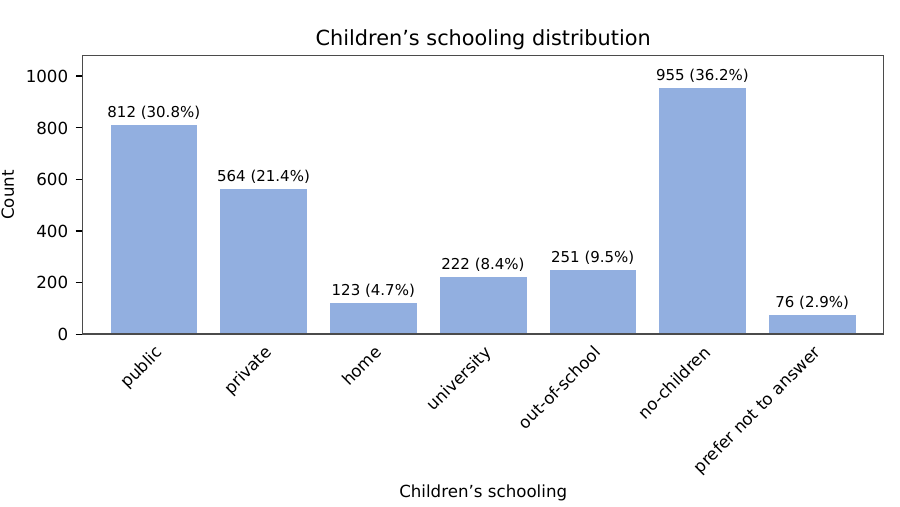}
        \caption{Children's Schooling}
    \end{subfigure}
    \begin{subfigure}[t]{0.32\textwidth}
        \centering
        \includegraphics[width=\linewidth]{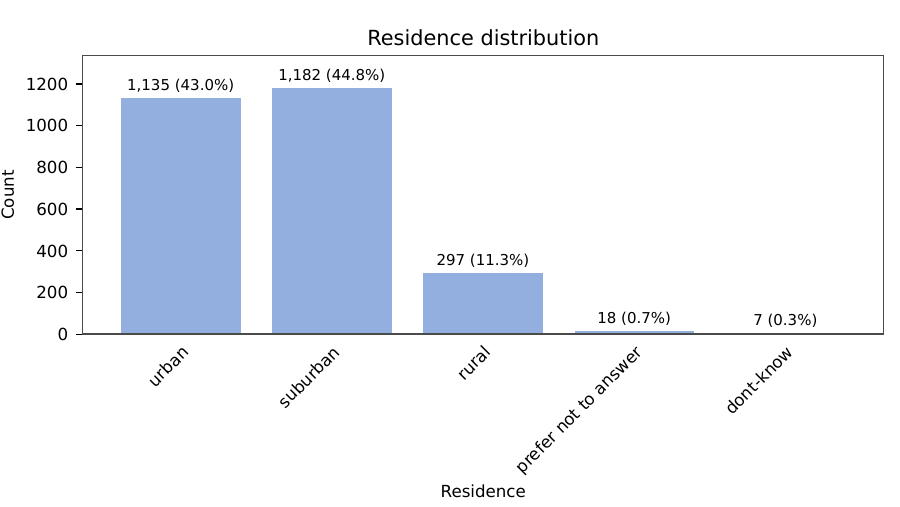}
        \caption{Residence Type}
    \end{subfigure}
    \begin{subfigure}[t]{0.32\textwidth}
        \centering
        \includegraphics[width=\linewidth]{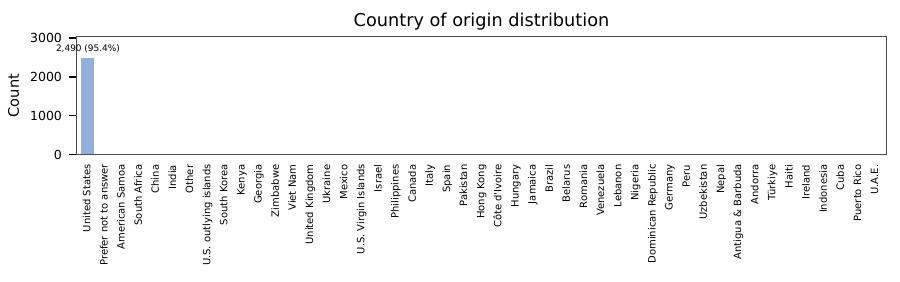}
        \caption{Country of Origin}
    \end{subfigure}

    \caption{Demographic distributions across age, gender, education, ethnicity, income, political identity and views, religion, family, and geographic background. Ethnicity and children's schooling allow multiple selections.}
    \label{fig:demographics}
\end{figure*}

\section{Prompt Templates for LLM Role-play Simulation}
\label{sec:prompt_templates}

We detail the prompt templates used to construct the memory module $\mathcal{M}_{a_i,k}$ for each RPLA $a_i$ in our multi-agent opinion exchange setup. Each agent simulates a human participant and is prompted with information that mirrors the participant's first-person memory before producing the $k$-th utterance in a given round.

Each simulation begins with a system prompt that defines the agent’s persona and task framing, followed by a sequence of user prompts corresponding to different memory components. All simulations adhere to the closed-world assumption (see Section~\ref{subsec:agent_construction}) and are structured to match the human task instructions (see Section~\ref{subsec:task}).

Table~\ref{tab:prompt-templates-simulation} illustrates an example prompt used in \textit{Next Message Prediction} (Section~\ref{subsec:agent_simulation}). This example reflects the memory state of agent $a_i$ at the beginning of Round~3, where all prior tweets and utterances are written by humans and added to the prompt as input. Each user prompt corresponds to one component of the memory module $\mathcal{M}_{a_i,k}$: demographic profile $d_{s_i}$, task instruction, initial opinion $(o^{\text{init}}_{s_i}, j^{\text{init}}_{s_i})$, previous rounds’ tweets and dyadic conversations $\{\tau^{r'}_{s}, \mathcal{C}^{r'}_{s} : 1 \leq r' < 3\}$, and current round context including partner tweets and prior utterances $(\tau^3_{s_i}, \tau^3_{s_j}, \{u^3_{k', s} : k' < k\})$. Curly brackets (\{\}) denote placeholder variables specific to each agent and topic instance. For readability, color highlights in the table correspond to different memory components.

\clearpage
\onecolumn
{
\scriptsize
\centering
\begin{longtable}[c]{p{1.7cm} p{0.9cm} p{5.25cm} p{5.25cm}@{}}
\caption{Prompt templates used to construct the memory module $\mathcal{M}_{a_i,k}$ for each RPLA $a_i$ during role-play (Section~\ref{subsec:agent_construction}). This example reflects the memory state of agent $a_i$ at the beginning of Round 3 under \textit{Next Message Prediction} (Section~\ref{subsec:agent_simulation}), where prior tweets and utterances written by humans were added to the memory. Each prompt governs one component of memory: demographic profile $d_{s_i}$, task instruction, initial opinion $(o^{\text{init}}_{s_i}, j^{\text{init}}_{s_i})$, previous rounds $\{\tau^{r'}_{s}, \mathcal{C}^{r'}_{s} : 1 \leq r' < 3\}$, and current round context $(\tau^3_{s_i}, \tau^3_{s_j}, \{u^3_{k', s} : k' < k\})$. Curly brackets (\{\}) denote placeholder variables that are different for each agent and topic. Color highlights correspond to different memory components.}

\label{tab:prompt-templates-simulation} \\

\toprule
    \textbf{Prompt Type} & \textbf{Message Type} & \textbf{Prompt Template} & \textbf{Example Values for Placeholders} \\
    \midrule
    
    \parbox[t]{1.7cm}{
    Agent Initialization:\\ 
    \textcolor{darkred}{Demographic Profile ($d_i$)},\\ \textcolor{darkgreen}{Task Instruction},\\ 
    \textcolor{darkblue}{Initial Opinion ($o_i^{\text{init}}, j_i^{\text{init}}$)}} &
    \parbox[t]{0.9cm}{\textit{System Message}} &
    \parbox[t]{5.25cm}{
    \textcolor{darkgreen}{Role play this person:} \\
    
    \textcolor{darkred}{
    You are a \{age\}-year-old \{gender\} with \{education\} education. Your ethnicity is \{ethnicity\}, and your annual income falls in the \{income bracket\} range. Politically, you identify as \{party ID\} with \{ideology\} views. You have children in \{children\_school\_status\}, reside in a \{urbanicity\} area, and your marital status is \{marital status\}. Regarding religious beliefs, you consider the Bible to be \{bible view\}, \{yes/no\} identify as evangelical, and your religious affiliation is \{religious affiliation\}. Your occupation is \{occupation\}.
    } \\
    
    \textcolor{darkgreen}{
    You have been interacting with other strangers on Twitter. You can decide to change or maintain your belief about the topic \{topic\}. You would first write a tweet about the topic \{topic\} that reflected your opinion. You would then engage in a private conversation through a textbox with a different stranger. In the conversation, you would first see the tweet the stranger wrote along with your own tweet. After seeing both tweets, you would be asked to read and respond to the stranger about the topic \{topic\}. 
    \\
    \\
    Throughout the interactions, you are alone in your room with limited access to the Internet. You cannot search for information about the topic \{topic\}, nor go out to ask other people. To form your belief, you can only rely on your initial belief and the information shared by others on Twitter.
    } \\
    
    \textcolor{darkblue}{
    \\
    Before interacting with other people, below is your initial opinion on \{topic\} using a 6-point Likert scale: \\
    - Certainly disagree \\
    - Probably disagree \\
    - Lean disagree \\
    - Lean agree \\
    - Probably agree \\
    - Certainly agree \\
    On the Likert scale, you chose \{Likert-scale opinion\} as your initial opinion regarding the statement \{topic\}. \\
    Below is your explanation for your initial opinion: \{free-text justification\} \\
    \\
    This opinion represents your starting point. It's based on your current understanding, personal experiences, and the beliefs that have shaped your perspective. As you engage in discussions, your views may evolve, but this is where you begin.
    }
    } &
    
    \parbox[t]{5.25cm}{
        \textbf{Demographic Profile:} \\
        \textcolor{darkred}{
            age = 41 \\
            gender = female \\
            education = master \\
            ethnicity = white \\
            income bracket = 50k-75k \\
            party ID = republican \\
            ideology = conservative \\
            children\_school\_status = ['private', 'university'] \\
            urbanicity = rural \\
            marital status = married \\
            bible view = literal \\
            evangelical = yes \\
            religious affiliation = protestant \\
            occupation = finance \\
            }\\
        \\
        \textbf{Task Instruction:} \\
            \textcolor{darkgreen}{
                topic = ``You are satisfied with how the political system is functioning in the US these days.''
                }\\
        \\
        \textbf{Initial Opinion:} \\
            \textcolor{darkblue}{
            topic = ``You are satisfied with how the political system is functioning in the US these days.'' \\
            Likert-scale response = ``Probably agree'' \\
            Explanation = ``I am indeed satisfied with the political system because the government is trying hard enough to introduce cryptocurrency to the market, which is the future currency of the world.''        
            }
    }
    
    \\
    \midrule
    
    \parbox[t]{1.7cm}{

    Conversation History:\\
    Previous Rounds (\textcolor{darkred}{Round 1} \& \textcolor{darkgreen}{Round 2}),\\
    Current Round Context (\textcolor{darkblue}{Round 3})
    } &
    \parbox[t]{0.9cm}{\textit{User Message}} &
    \parbox[t]{5.25cm}{

    \textcolor{darkred}{
    Below was your conversation with \{first\_partner\_name\}\\
    \\
    My tweet: \{$\tau^1_{s_1}$\}\\
    \{first\_partner\_name\}'s tweet: \{$\tau^1_{s_2}$\}\\
    \\
    My response: \{$u^1_{1, s_1}$\}\\
    \{first\_partner\_name\}'s response: \{$u^2_{2, s_2}$\}\\    
    My response: \{$u^1_{3, s_1}$\}\\
    \{first\_partner\_name\}'s response: \{$u^2_{4, s_2}$\}\\        
    ...\\
    ...}
    \\
    \\
    \textcolor{darkgreen}{    
    You have just finished your conversation with \{first\_partner\_name\}. Instead, you are now engaging in conversation with another stranger \{second\_partner\_name\} on a separate text box.
    \\
    \\
    Below was your conversation with \{second\_partner\_name\}.\\
    \\
    My tweet: \{$\tau^2_{s_1}$\}\\
    \{second\_partner\_name\}'s tweet: \{$\tau^2_{s_3}$\} \\    
    \\    
    \{second\_partner\_name\}'s response: \{$u^2_{1, s_3}$\}\\        
    My response: \{$u^2_{2, s_1}$\}\\
    \{second\_partner\_name\}'s response: \{$u^2_{3, s_3}$\}\\        
    My response: \{$u^2_{4, s_1}$\}\\    
    ...\\
    ...}
    \\
    \\
    \textcolor{darkblue}{  
    You have just finished your conversation with \{second\_partner\_name\}. Instead, you are now engaging in conversation with another stranger \{third\_partner\_name\} on a separate text box.
    \\
    \\    
    Below was your conversation with \{third\_partner\_name\}.\\
    \\
    My tweet: \{$\tau^3_{s_1}$\}\\
    \{third\_partner\_name\}'s tweet: \{$\tau^3_{s_4}$\} \\    
    \\    
    My response: \{$u^3_{1, s_1}$\}\\
    \{third\_partner\_name\}'s response: \{$u^3_{2, s_3}$\}\\  
    My response: \{$u^3_{3, s_1}$\}\\
    \{third\_partner\_name\}'s response: \{$u^3_{4, s_3}$\}\\      
    ...\\
    ...\\
    }    
    } 
    &
    \parbox[t]{5.25cm}{
    \textbf{Previous Rounds (Round 1):} \\
    \textcolor{darkred}{
    $\tau^1_{s_1}$ (Your tweet) = I am satisfied with political system because the government is trying hard enough to stabilize the economy through various ways like transitioning to crypto currency \\
    $\tau^1_{s_2}$ (681e3's tweet) = I disagree with the statement that I am satisfied with the way the American system functions these days. This is because of the system's extreme polarization making it fail to take meaningful action \\
    $u^1_{1, s_1}$ (Your response) = From my point of view, the government is not that perfect but at least it’s trying to improve the lives of all Americans \\
    $u^2_{2, s_2}$ (681e3's response) = I still believe that the political system is flawed but I completely see your viewpoint. Politicians appear to care more about maintaining party allegiance than they do about the problems that people care about. Can the system be re-organized in your opinion or is the division too great? The fact that everything has become more divisive which makes compromise nearly impossible in my opinion is largely to blame. What do you think?} \\
    \\
    \textbf{Previous Rounds (Round 2):} \\
    \textcolor{darkgreen}{
    $\tau^2_{s_1}$ (Your tweet) = We should support the government motives to improve and make our country great. On my side, the government is doing the best it can to stabilize our economy and improve our lives \\
    $\tau^2_{s_3}$ (683b8's tweet) = I agree and I am totally satisfied with how the political system is working. This is because it is promoting good health and education facilities to its citizens. \\
    $u^2_{1, s_3}$ (683b8's response) = It provide strict laws. It gives freedom to all citizens to publicly participate in elections. \\
    $u^2_{2, s_1}$ (Your response) = I second your point, the government has helped the education sector through scholarships. It has also invested a lot of resources in the healthcare field. Yes, it also gives each citizen the right to express one's ideas and opinions. \\
    $u^2_{3, s_3}$ (683b8's response) = It has also improved infrastructure and advancement of technology.} \\
    \\
    \textbf{Current Round Context (Round 3):} \\
    \textcolor{darkblue}{
    $\tau^3_{s_1}$ (Your tweet) = The government plays a crucial role in advancement of technology by budgeting enough resources. It also helps in infrastructure and healthcare, I support \\
    $\tau^3_{s_4}$ (68405's tweet) = The government allows its people participation on the development project and is highly working on development \\
    $u^3_{1, s_4}$ (68405's response) =  I do agree on advancing the technology and improving also in defense force and provide high security \\
    $u^3_{2, s_1}$ (Your response) = Yes, the government contributes to the general development of the country by investing enough money onto different projects \\
    $u^3_{3, s_4}$ (68405's response) = That’s okay. It's also improving on more projects and inventions \\
    $u^3_{4, s_1}$ (Your response) = It also contributes to a stable economy
    }
    }

    \\
    \bottomrule
\end{longtable}
}

\newpage
\section{LLM-Based Annotation for Validity (On-topicness) and Stance}
\label{app:llm_eval_classifier}

To evaluate RPLA alignment with human behavior, we use \texttt{gpt-5-mini} with \texttt{reasoning\_effort=medium} as a classifier: one for identifying on-topic utterances and another for mapping stance to a scalar value. Both classifiers are implemented using prompting.

\paragraph{Validity (on-topic) Classification.}
For each generated text item and associated discussion topic $t$, we apply a message-type-specific binary validity classifier. We use ``on-topic'' as shorthand for this broader validity assessment, which considers topic relevance, language coherence, appropriate format, and whether the text functions as a meaningful contribution for its stage. Conversation utterances are judged from the topic, prior chat history, and latest message; tweets, initial opinions, and post-conversation opinions use matched prompt variants specialized to those output formats. The validity criteria exclude filler-only or off-topic text, malformed or incoherent generations, degenerate repetition, prompt leakage, JSON or metadata artifacts, and other content that fails to function as a valid response for that message type. For conversation utterances, this yields the binary indicator $I_{\text{topic}}(\widehat{u}, t) \in \{0,1\}$. This label is used in two places in the paper: to retain benchmark groups that contain sufficient valid/on-topic human content during dataset cleaning and to keep only valid/on-topic human target messages when constructing SFT context--response pairs. To ensure reliability, we manually labeled a 10\% stratified sample spanning all 107 topics, the seven base LLMs, and both simulation setups for the tweet and private-opinion output families, yielding 11,161 tweets and 7,423 initial/post private-opinion texts in total. The LLM labels agreed with human judgment at approximately 90\% on this validation set. Validity prompts vary by message type and between the base-model and SFT evaluations. Tables~\ref{tab:prompt_ontopic_tweet}, \ref{tab:prompt_ontopic_initial}, and \ref{tab:prompt_ontopic_post} illustrate the validity criteria and required output format. The user prompt is intentionally short in each case: it mainly supplies the generated text span and asks for a binary label, while the validity criteria are specified in the system prompt.

\begin{table*}[!tbp]
\scriptsize
\centering
\caption{Prompt template for tweet validity classification with \texttt{gpt-5-mini}.}
\begin{tabular}{p{0.96\linewidth}}
\toprule
\textbf{System Prompt} \\
\midrule
\parbox[t]{\linewidth}{\vspace{0pt}\# Tweet Validity Judge\\Your task is to analyze a tweet generated by either a human or a role-playing LLM about the topic ``\texttt{\{TOPIC\}}''. You must decide whether the tweet is a valid and relevant tweet for that topic.\\Mark the tweet as VALID only if all of the following are true:\\1. It is relevant to the topic. The topic connection can be implicit, but it must still be reasonably clear.\\2. It is written as a tweet or short public post, not as a chat reply, instruction, rubric answer, transcript fragment, or post-conversation reflection.\\3. It is coherent natural language and not gibberish, malformed text, repeated filler, or nonsense.\\4. Minor wrapper artifacts such as ``My Response:'' or ``My Tweet:'' are acceptable if the underlying content is still a coherent, topic-relevant tweet. Mark the tweet INVALID if instruction text, restart text, markdown scaffolding, conversation setup, or other prompt-spill content takes up a substantial portion of the text or frames the tweet as an embedded artifact rather than the actual output.\\5. It is not mostly greeting, small talk, or off-topic social chatter.\\Mark the tweet as INVALID if any of the following apply:\\- It is off-topic or only weakly related to the topic.\\- It is primarily a greeting, conversational opener, or chat continuation.\\- It contains prompt leakage, instructions, restart text, or copied task text.\\- It includes template markers, wrappers, metadata artifacts, markdown sections, or conversation scaffolding that dominate the text or leave no clean standalone tweet content.\\- It is too malformed, truncated, repetitive, or incoherent to count as a valid tweet.\\Be strict about off-topic content, prompt leakage, and malformed text. A plausible tweet fragment is not enough for VALID if the overall text is mostly restart instructions, scaffolding, copied setup, or other non-tweet artifacts. If the response reads like a workflow transcript or setup message with a tweet embedded inside it, mark it INVALID. Do not reject an otherwise valid clean tweet solely because it includes a light wrapper artifact.} \\
\midrule
\textbf{User Prompt} \\
\midrule
\parbox[t]{\linewidth}{\vspace{0pt}\# Tweet Input\\Below is the tweet that was generated.\\\texttt{\{TWEET\}}\\Based on the topic of interest and the criteria above, determine if the tweet is valid and relevant. Answer with ``VALID'' or ``INVALID'' only.} \\
\bottomrule
\end{tabular}
\label{tab:prompt_ontopic_tweet}
\end{table*}
\begin{table*}[!tbp]
\scriptsize
\centering
\caption{Prompt template for initial-opinion validity classification with \texttt{gpt-5-mini}.}
\begin{tabular}{p{0.96\linewidth}}
\toprule
\textbf{System Prompt} \\
\midrule
\parbox[t]{\linewidth}{\vspace{0pt}\# Initial Opinion Validity Judge\\Your task is to analyze an initial opinion generated by either a human or a role-playing LLM about the topic ``\texttt{\{TOPIC\}}''. The person was asked to express their initial stance on the topic. You must decide whether the initial opinion is a valid and relevant response to that prompt.\\Mark the opinion as VALID only if all of the following are true:\\1. It contains an explicit stance, or a clearly inferable stance, toward the topic. The answer must communicate agreement, disagreement, or a clear directional lean.\\2. It is relevant to the topic. Pronouns like ``it'' or ``this'' are acceptable only if the topic reference is still reasonably clear.\\3. It is coherent natural language, not gibberish, nonsense, repeated filler, or unrelated content.\\4. Minor wrapper artifacts such as ``My Response:'' or ``My Opinion:'' are acceptable if the underlying content is still a clear, substantive initial stance. Mark the opinion INVALID only if the artifact is the whole answer, dominates the answer, or the remaining content is not a valid response.\\5. It does not contain prompt leakage, copied instructions, JSON, role tags, or other metadata artifacts that overwhelm the actual opinion content.\\Mark the opinion as INVALID if any of the following apply:\\- It does not state a stance or directional lean.\\- It is off-topic or mostly unrelated to the topic.\\- It is primarily greeting, small talk, filler, or chat continuation.\\- It is too malformed, truncated, repetitive, or incoherent to count as a valid initial opinion.\\- It is just a template fragment such as ``My Response'', ``My Opinion'', ``No comments'', or similar placeholders, without substantive topic-relevant content.\\Be strict about missing stance, off-topic content, and malformed text. Do not reject an otherwise valid answer solely because it includes a light wrapper artifact.} \\
\midrule
\textbf{User Prompt} \\
\midrule
\parbox[t]{\linewidth}{\vspace{0pt}\# Initial Opinion Input\\Below is the initial opinion that was generated.\\\texttt{\{INITIAL\_OPINION\}}\\Based on the topic of interest and the criteria above, determine if the initial opinion is valid and relevant. Answer with ``VALID'' or ``INVALID'' only.} \\
\bottomrule
\end{tabular}
\label{tab:prompt_ontopic_initial}
\end{table*}

\begin{table*}[!tbp]
\scriptsize
\centering
\caption{Prompt template for post-conversation-opinion validity classification with \texttt{gpt-5-mini}.}
\begin{tabular}{p{0.96\linewidth}}
\toprule
\textbf{System Prompt} \\
\midrule
\parbox[t]{\linewidth}{\vspace{0pt}\# Post Opinion Validity Judge\\Your task is to analyze a post-conversation opinion generated by either a human or a role-playing LLM. After discussing the topic ``\texttt{\{TOPIC\}}'', the person was asked to state whether they agree or disagree with the topic and explain why. You must decide whether the post-conversation opinion is a valid answer to that prompt.\\Mark the opinion as VALID only if all of the following are true:\\1. It contains an explicit stance, or a clearly inferable stance, toward the topic. The answer must communicate agreement, disagreement, or a clearly directional lean. Generic chat, placeholders, greetings, or meta comments do not count as a stance.\\2. It contains a topic-grounded explanation, justification, or final-topic summary. A brief explanation is enough if it makes the final stance clear. For post opinions, a short but clearly directional final answer can still be VALID when it plainly states the speaker's end position on the topic, even if the justification is concise.\\3. It is relevant to the topic. Pronouns like ``it'' or ``this'' are acceptable if the topic reference is clearly recoverable from the post-opinion prompt and the rest of the answer still comments on that claim. In a post-opinion response, default to interpreting standalone references like ``this'', ``it'', or ``that's'' as the discussion claim unless the answer clearly points somewhere else. Short final answers such as ``This is just not true'' or ``I don't think that's right'' can be VALID when they are plainly referring to the discussion topic and express a clear end stance.\\4. It is coherent natural language, not gibberish, nonsense strings, malformed placeholders, repeated metadata markers, or unrelated content.\\5. Minor wrapper artifacts such as ``My Response:'' or ``My Opinion:'' are acceptable if the underlying content is still a clear, substantive, topic-grounded answer. Mark the opinion INVALID only if the artifact is the whole answer, dominates the answer, or the remaining content is not a valid response.\\Mark the opinion as INVALID if any of the following apply:\\- It does not state a stance.\\- It gives no topic-grounded reason or final-topic summary at all, and the stance does not stand on its own as a clear final position on the topic.\\- It is mostly greeting, small talk, filler, or a chat continuation.\\- It is obviously off-topic.\\- It is nonsense, truncated, repetitive, or malformed.\\- It is just a template fragment such as ``My Response'', ``My Opinion'', ``No comments'', or similar placeholders, without substantive topic-grounded content.\\Be strict about missing stance, off-topic content, or malformed text. Do not reject an otherwise valid answer solely because it is brief, especially when the brief answer still gives a clear final stance about the topic, or because it includes a light wrapper artifact.\\Do not require the answer to restate the topic noun phrase explicitly when the stance is otherwise plainly expressed in direct response to the post-opinion prompt.} \\
\midrule
\textbf{User Prompt} \\
\midrule
\parbox[t]{\linewidth}{\vspace{0pt}Below is the post-conversation opinion that was generated after the conversation.\\\texttt{\{POST\_OPINION\}}\\Based on the topic of interest and the criteria above, determine if the post-conversation opinion is a valid answer to the prompt. Answer with ``VALID'' or ``INVALID'' only.} \\
\bottomrule
\end{tabular}
\label{tab:prompt_ontopic_post}
\end{table*}

\paragraph{Stance Classification.}  
To evaluate opinion alignment, we map each utterance $u$ or $\widehat{u}$ to a scalar stance score $S(u) \in \{1,2,3,4,5,6\}$ also using \texttt{gpt-5-mini}. These scores correspond to the following six-point Likert categories: \textit{(1) Certainly disagree, (2) Probably disagree, (3) Lean disagree, (4) Lean agree, (5) Probably agree, (6) Certainly agree}.

We manually annotated the same 10\% stratified sample with stance labels and tuned the prompt to produce outputs that aligned with human judgment. The final prompt includes the topic, utterance, and the instruction to return one of the six stance categories, which are then mapped to scalar values.

For both classifiers, this stratified manual validation yielded about 90\% agreement with the LLM labels. Combined with the human-human agreement analyses reported below, this indicates that both on-topic classification and stance classification are reasonably well-defined rather than highly subjective. The full prompt template is shown in Table~\ref{tab:prompt_stance_template}.

\begin{table*}[!tbp]
\scriptsize
\centering
\caption{Prompt template used for stance classification.}
\resizebox{\linewidth}{!}{%
\begin{tabular}{p{0.98\linewidth}}
\toprule
\textbf{Prompt Template} \\
\midrule
\textit{Plain, in Evaluation without prior messages}

What is the stance of the following message with respect to the topic "\{TOPIC\}"? Here is the message: "\{TEXT\}". Please use exactly one word from the following 6 categories to label it: "Certainly agree", "Probably agree", "Lean agree", "Lean disagree", "Probably disagree", and "Certainly disagree". Your predicted label: \\

\midrule
\textit{Memory, in Evaluation with prior messages}

Here is prior conversation around the topic "\{TOPIC\}":\\
\{CONVERSATION\}\\
What is the stance of the following new message by \{PLAYER\} with respect to the topic "\{TOPIC\}"? Here is the message: "\{TEXT\}". Please use exactly one word from the following 6 categories to label it: "Certainly agree", "Probably agree", "Lean agree", "Lean disagree", "Probably disagree", and "Certainly disagree". Your predicted label: \\

\midrule
\textit{User Message in Simulation for post-discussion private opinions} \\
Based on your previous discussions about \{TOPIC\}, please select your level of agreement from the following options. Strictly answer with one of the options below without any additional text:\\

- Certainly disagree\\
- Probably disagree\\
- Lean disagree\\
- Lean agree\\
- Probably agree\\
- Certainly agree\\
\bottomrule
\end{tabular}%
}
\label{tab:prompt_stance_template}
\end{table*}

\paragraph{Human--human inter-annotator agreement.}  
To characterize the subjectivity of the labeling tasks, we computed human--human inter-annotator agreement on 400 randomly sampled messages spanning all topics and both human and LLM outputs (including tweets, initial opinions, final opinions, and conversation turns). Two human annotators independently labeled each message for validity/topic relevance (binary) and stance (six-point ordinal classification using the same Likert scheme as above). For validity/topic relevance, they achieved 96.8\% raw agreement and Cohen’s $\kappa = 0.89$. For stance, they achieved Cohen’s $\kappa = 0.81$. Under commonly used guidelines for interpreting Cohen’s $\kappa$ \citep{viera2005understanding,mchugh2012interrater}, values $\kappa \ge 0.81$ are typically described as indicating ``almost perfect agreement,'' suggesting that the labeling scheme is well-defined and that our reported $\sim$90\% LLM--human agreement reflects a reasonably reliable automatic judge.
We assess judge robustness in two ways. First, we compare annotations with Gemini-3-pro-preview as a second LLM judge. Agreement remained high for on-topic labels (\textit{Next Message Prediction}: $0.88$--$0.98$ across sources; \textit{Full Dynamics Simulation}: $0.75$--$0.99$), and the corresponding 6-way stance annotations were also strongly correlated (Spearman $\rho=0.89$--$0.93$ for \textit{Next Message Prediction}; $\rho=0.82$--$0.92$ for \textit{Full Dynamics Simulation}). On the evaluated subset, the utterance-level conclusions, including the \textit{Next Message Prediction} $>$ \textit{Full Dynamics Simulation} gap and the gains from SFT, remained unchanged.

Second, we compare human and model convergence on a 20-group sample with human stance annotations. Table~\ref{tab:human_judged_convergence_check} shows that the core over-convergence pattern is preserved: human groups still converge much less than RPLA rollouts under direct human annotation.

\begin{table}[h]
\centering
\caption{Human-judged convergence ($\Delta\mathrm{SD}$) on a 20-group sample. More negative values indicate stronger convergence.}
\label{tab:human_judged_convergence_check}
\begin{tabular}{lrr}
\toprule
Metric on human-judged annotations & Human & RPLA \\
\midrule
{$\Delta$ tweet disagreement} & {$-0.12$} & {$-0.24$} \\
{$\Delta$ final-opinion disagreement} & {$-0.11$} & {$-0.33$} \\
\bottomrule
\end{tabular}
\end{table}

Thus, the main over-convergence conclusion does not depend on the original automatic judge alone.

\section{Reference Baselines and DeGroot Fitting}
\label{app:degroot_baseline}

\paragraph{Reference baselines used in the main figures.}
The main-text comparison figure includes two baselines in addition to the fitted DeGroot baseline. The baseline labelled \emph{Noise UB} add independent zero-mean Gaussian noise with standard deviation $\sigma=1.0$ or $\sigma=0.5$ to human stance labels, clip the perturbed labels to $[1,6]$, and round to the nearest Likert category before recomputing the same stance-based errors. These references calibrate the scale of the reported errors.

\paragraph{Convergence summaries and heuristic targets.}
For the detailed convergence analysis, let $\bar{s}^{c}_{k}(m)$ denote the mean within-group stance SD at the initial ($k=0$) or final ($k=1$) observation over groups with complete observations for opinion type $c$ and model $m$. The  convergence summaries are defined as:
\begin{equation}
C^{c}(m)=\bar{s}^{c}_{1}(m)-\bar{s}^{c}_{0}(h).
\end{equation}
The  standard error is propagated as $\sqrt{\operatorname{SE}(\bar{s}^{c}_{1}(m))^2+\operatorname{SE}(\bar{s}^{c}_{0}(h))^2}$. This displayed convergence level is distinct from the mean absolute group-level convergence error in Eq.~\ref{eq:opinion-convergence-error}. The figure also includes two baselines. For \emph{One-Step Toward First Partner}, if $q(i)$ denotes participant $i$'s first dyadic partner, we define
\begin{equation}
T(z_i, z_{q(i)}) = \begin{cases}
\min(6, z_i + 1) & \text{if } z_{q(i)} > z_i, \\
\max(1, z_i - 1) & \text{if } z_{q(i)} < z_i, \\
z_i & \text{otherwise.}
\end{cases}
\end{equation}
and aggregate the resulting changes at the group level. For \emph{50\% SD Reduction}, the target is $-\tfrac{1}{2}\,\mathrm{SD}(\{x^{\mathrm{pub},0}_i(h): i \in g\})$ for public opinions and $-\tfrac{1}{2}\,\mathrm{SD}(\{x^{\mathrm{priv},0}_i(h): i \in g\})$ for private opinions, again averaged over groups.

The manuscript reports a split-specific fitted DeGroot baseline implemented directly on the stance variables. For each participant $i$, let $q(i)$ denote the first dyadic partner observed in the human interaction log. The public trajectory is initialized from the human round-1 tweet stances and rolled forward for rounds 2 and 3 via
\begin{equation}
z^{\mathrm{pub}}_{i,r+1}=\mathrm{clip}_{[1,6]}\Big(\alpha_{\mathrm{pub}} z^{\mathrm{pub}}_{i,r} + (1-\alpha_{\mathrm{pub}}) z^{\mathrm{pub}}_{q(i),r}\Big),
\qquad r \in \{1,2\}.
\end{equation}
The private trajectory is initialized from the human reported initial opinion and updated toward the partner's current public state via
\begin{equation}
z^{\mathrm{priv}}_{i,r}=\mathrm{clip}_{[1,6]}\Big(\alpha_{\mathrm{priv}} z^{\mathrm{priv}}_{i,r-1} + (1-\alpha_{\mathrm{priv}}) z^{\mathrm{pub}}_{q(i),r}\Big),
\qquad r \in \{1,2,3\},
\end{equation}
where $z^{\mathrm{priv}}_{i,0}$ is the human initial private opinion. All states are clipped to the Likert range $[1,6]$, matching the stance labels used throughout the paper.

\paragraph{Fit objective.}
We fit separate coefficients for public and private opinions, using only the available human observations in the training partition for each split and simulation setting. First, we fit $\alpha_{\mathrm{pub}} \in [0,1]$ by minimizing the pooled per-agent mean squared error over the predicted round-2 and round-3 public tweet stances. Concretely, we aggregate the per-agent squared errors for tweet round 2 and tweet round 3, pool those two error lists, and minimize their mean. Conditional on the fitted $\alpha_{\mathrm{pub}}$, we then fit $\alpha_{\mathrm{priv}} \in [0,1]$ by minimizing the per-agent mean squared error on the final private opinion. After both coefficients are chosen, the fitted pair is evaluated on the matching train and held-out test partitions, and we record tweet-2 MSE, tweet-3 MSE, pooled public MSE, final-private MSE, and the combined MSE across public and private targets.

\paragraph{Search procedure.}
The fitter uses a bounded one-dimensional search for each coefficient. It first evaluates a coarse grid of 21 evenly spaced points on $[0,1]$, then refines promising intervals with a bounded interval-refinement search until the interval is sufficiently small. Ties are broken toward the smaller coefficient. This procedure is run once for $\alpha_{\mathrm{pub}}$ and then again for $\alpha_{\mathrm{priv}}$ with the fitted public coefficient held fixed.

\paragraph{Fitted values used in the paper.}
For Full Dynamics Simulation, the fitted coefficients $(\alpha_{\mathrm{pub}},\alpha_{\mathrm{priv}})$ are $(0.857,0.924)$ for group generalization and $(0.866,0.929)$ for topic generalization. For Next Message Prediction, the corresponding pairs are $(0.865,0.921)$ and $(0.869,0.920)$.

These fitted values are not chosen to optimize the group-convergence bar directly. Rather, they are selected from per-agent public-trajectory and final-private prediction error on the training partition, and only then carried into the held-out convergence, histogram, transition-matrix, and partner-gap analyses shown in the manuscript.

\section{Compute Resources}
\label{app:compute_resources}
We ran the experiments on a GPU machine equipped with 1x NVIDIA H100 PCIe (80GB). This covers fine-tuning, evaluation, statistical analysis, figure generation, and open-weight simulation runs.

The approximate runtimes were as follows: evaluation plus statistical analysis for one complete benchmark sweep typically required less than 1 GPU-hour; one SFT run for \texttt{Llama-3.1-8B-Instruct} required roughly 8--9 GPU-hours; and one full-dynamics simulation sweep for an open-weight model required on the order of 4--12 GPU-hours depending on model size, decoding length, and simulation mode. An SFT experiment required roughly 13 hours per run.

Some experiments also use proprietary APIs (e.g., OpenAI and Gemini models). For those runs, the underlying vendor-managed hardware is not directly observable from our side; the local resource requirements above therefore describe the parts of the pipeline that are directly reproducible on our hardware.

Because the study mixes local GPU runs with vendor-managed API calls, and because the broader project included exploratory runs beyond the final reported experiments, we do not report a single precise end-to-end total compute budget for the full project.

\section{Limitations}
\label{app:limitations}

DEBATE is intended as a controlled benchmark for short-horizon, four-person online deliberation rather than a full model of society-scale discourse. Its closed 25--30-minute sessions do not model asynchronous communication, exogenous information arriving during a discussion, or multi-hop propagation through an evolving social network. ``Large-scale'' describes the number and diversity of benchmark participants, groups, and topics, not the size of each interaction network. Because groups are independent and fully connected, DEBATE does not identify network-level diffusion scope, cross-group transmission, or echo-chamber formation; evaluating those phenomena requires matched longitudinal data with inter-group network structure. The participant pool is U.S.-based and demographically diverse, but it is not nationally representative, so downstream users should avoid over-generalizing the observed dynamics to other populations, languages, platforms, or longer-horizon settings. Larger matched human networks would provide complementary evidence about diffusion, network segregation, and the generality of the observed convergence patterns.

The benchmark also measures only traces we can observe directly: public interaction histories together with initial and final public/private opinions. It does not identify latent reasoning processes, long-run persistence of belief change, or whether models that match these traces would generalize to other discussion settings. Our current experiments evaluate seven LLMs, with SFT replications across three model families, so the reported gains should be interpreted as benchmark-specific evidence rather than as a general solution for improving alignment in \textit{Full Dynamics Simulation}. Appendix~\ref{app:supp} summarizes the publicly available dataset and benchmark artifacts, and all simulation modes and prompt templates are documented and implemented, but end-to-end reproduction may still vary across vendor-managed APIs and future model versions.

Collecting a wide range of viewpoints is not an endorsement of any particular position; these data can support research on societal risks such as misinformation spread, polarization, and echo-chamber formation, but politically sensitive conversation data also create privacy and misuse concerns. The anonymized released materials are fully de-identified: real names and direct identifiers are not collected, platform IDs are replaced with random pseudonyms, and we run basic automated and manual checks to remove potential residual personal information. The released materials document the dataset schema, de-identification procedure, intended research use, and known limitations. Our primary safeguards are documentation and de-identification rather than access gating, so reuse should proceed cautiously and findings should be re-validated in new settings.

\section{Supervised Fine-Tuning}

\label{app:sft}

This appendix extends the main-text stance-based opinion-dynamics evaluation with topic-generalization result and opinion distributions for the SFT comparisons. This focus is intentional: the benchmark's main contribution is evaluating whether simulated interactions recover individual and group-level opinion dynamics. We do not treat within-conversation chat utterances as a separate evaluation target. However, because the benchmark also records public tweet-like opinion expressions and final private-opinion text, we report complementary beyond-stance diagnostics for those outputs. Appendix~\ref{app:example_chats} provides qualitative examples.

To test whether opinion dynamics alignment can be improved through post-training, we experiment with supervised fine-tuning (SFT; \citealp{ouyang2022training}) in a \emph{next-message prediction} setting: given an agent’s memory state $x = \mathcal{M}_{a_i,k}$, the model is optimized to produce a human-written next message $y$. We evaluate \llamabase{} and replicate the core tests with \texttt{Qwen3.5-4B} (with thinking disabled) and \texttt{Mistral-7B-Instruct-v0.3}. For SFT, we minimize the conditional log-loss over context--response pairs $(x,y)$:
\begin{align*}
\mathcal{L}_{\text{SFT}}
&= - \mathbb{E}_{(x,y)}
    \big[ \log P_\theta(y \mid x) \big],
\end{align*}
where $(x,y)$ are drawn from the SFT training set $\mathcal{D}_{\text{train}}$ and $P_\theta(y \mid x)$ denotes the conditional distribution of the SFT model.

For these SFT runs, $\mathcal{D}_{\text{train}}$ keeps only context--response pairs whose human target message is a valid on-topic utterance under $I_{\text{topic}}(y,t)=1$; off-topic or otherwise invalid targets are excluded from the SFT supervision set. The same validity label is also used during dataset cleaning, and Appendix~\ref{app:llm_eval_classifier} documents the annotation procedure.

For the reported \llamabase{} SFT runs, we train for 5 epochs with completion-only loss, learning rate $5\times10^{-5}$, per-device train/eval batch size 8, and gradient accumulation 32 (effective training batch size 256). The group split provides 17,057 training examples and 3,804 validation examples; the topic split provides 14,662 and 6,199, respectively. Appendix~\ref{app:sft_hyperparams} reports the training and inference settings for all three SFT model families, and Appendix~\ref{app:compute_resources} describes the hardware context and approximate resource use.

\begin{figure}[htb!] 
\centering
\includegraphics[width=0.8\linewidth]{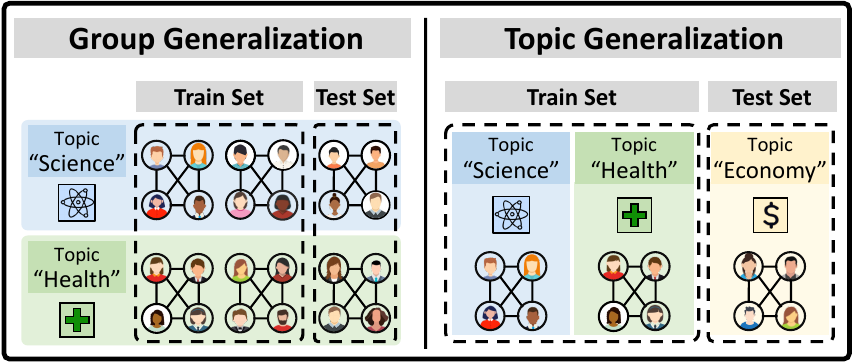}
\caption{Illustration of the two train/test partition strategies used for supervised fine-tuning (SFT) evaluation in this paper. \textbf{Group Generalization} (left): train and test on disjoint participant groups within overlapping topics. \textbf{Topic Generalization} (right): train and test on disjoint topic sets, which also induces new participant groups at test time.}
\label{fig:train_test_partition}
\end{figure}

Figure~\ref{fig:train_test_partition} visualizes the two train/test partition strategies used for SFT evaluation in this paper. In group generalization, train and test share topics but use disjoint participant groups; in topic generalization, the test topics are disjoint from the training topics and therefore also introduce new groups. Both strategies partition all 708 recorded groups; the human data and model-specific evaluation coverage are reported in Tables~\ref{tab:dataset-split-summary} and~\ref{tab:dataset-split-prefix-coverage}.  The appendix analyses below focus on topic-generalization counterparts to the group-generalization results in Section~\ref{sec:opinion_dynamics}.

\section{Cross-Model SFT Experiment Replication}
\label{app:sft-cross-model}

Table~\ref{tab:sft-cross-model-convergence} reports public and private opinion convergence ($\Delta\mathrm{SD}$) for the three LLM families, including both held-out-group and held-out-topic evaluations. Tables~\ref{tab:sft-cross-model-public-errors} and~\ref{tab:sft-cross-model-private-errors} report the corresponding public- and private-opinion prediction errors. Across \texttt{Qwen3.5-4B} and \texttt{Mistral-7B-Instruct-v0.3}, SFT improves single-opinion prediction, opinion-update prediction, and opinion-convergence prediction for private opinions on both the group-generalization and topic-generalization splits, and reduces public-opinion prediction errors overall. These replications support the directional conclusion that SFT can mitigate excess convergence and improve private-opinion fidelity across the three model families.

\begin{table}[tbp]
\caption{\textbf{Cross-model opinion convergence ($\Delta\mathrm{SD}$) under Full Dynamics Simulation.} Each model entry reports Base $\rightarrow$ SFT. More-negative values of $\Delta\mathrm{SD}$ indicate stronger convergence; values closer to the human value indicate greater convergence fidelity.}
\centering
\small
\begin{tabular}{llrrrr}
\toprule
\textbf{Split} & \textbf{Opinion type} & \textbf{Human} & \texttt{Llama-3.1-8B} & \texttt{Qwen3.5-4B} & \texttt{Mistral-7B} \\
\midrule
Group & Public & $-0.085$ & $-0.179 \rightarrow -0.106$ & $-0.326 \rightarrow -0.145$ & $-0.348 \rightarrow -0.118$ \\
Group & Private & $-0.096$ & $-0.345 \rightarrow -0.062$ & $-0.579 \rightarrow -0.080$ & $-0.476 \rightarrow -0.222$ \\
\midrule
Topic & Public & $-0.113$ & $-0.101 \rightarrow -0.083$ & $-0.458 \rightarrow -0.265$ & $-0.556 \rightarrow -0.252$ \\
Topic & Private & $-0.081$ & $-0.468 \rightarrow -0.116$ & $-0.673 \rightarrow -0.298$ & $-0.624 \rightarrow -0.201$ \\
\bottomrule
\end{tabular}
\label{tab:sft-cross-model-convergence}
\end{table}

\begin{table}[tbp]
\caption{\textbf{Cross-model public-opinion prediction errors under Full Dynamics Simulation.} Lower is better for all errors. Each entry reports the base-model error $\rightarrow$ SFT error on the indicated split.}
\centering
\small
\begin{tabular}{llrrr}
\toprule
\textbf{Model} & \textbf{Split} & \textbf{Single-opinion} & \textbf{Opinion-update} & \textbf{Opinion-convergence} \\
\midrule
\texttt{Llama-3.1-8B} & Group & $1.585 \rightarrow 1.137$ & $1.641 \rightarrow 1.079$ & $0.782 \rightarrow 0.582$ \\
\texttt{Llama-3.1-8B} & Topic & $1.635 \rightarrow 1.098$ & $1.696 \rightarrow 1.103$ & $0.854 \rightarrow 0.645$ \\
\midrule
\texttt{Qwen3.5-4B} & Group & $1.606 \rightarrow 1.028$ & $1.638 \rightarrow 1.019$ & $0.818 \rightarrow 0.701$ \\
\texttt{Qwen3.5-4B} & Topic & $1.543 \rightarrow 1.056$ & $1.580 \rightarrow 1.066$ & $0.829 \rightarrow 0.670$ \\
\midrule
\texttt{Mistral-7B} & Group & $1.633 \rightarrow 1.236$ & $1.494 \rightarrow 1.239$ & $0.840 \rightarrow 0.728$ \\
\texttt{Mistral-7B} & Topic & $1.688 \rightarrow 1.318$ & $1.555 \rightarrow 1.378$ & $0.792 \rightarrow 0.807$ \\
\bottomrule
\end{tabular}
\label{tab:sft-cross-model-public-errors}
\end{table}

\begin{table}[tbp]
\caption{\textbf{Cross-model private-opinion prediction errors under Full Dynamics Simulation.} Lower is better for all errors. Each entry reports the base-model error $\rightarrow$ SFT error on the indicated split.}
\centering
\small
\begin{tabular}{llrrr}
\toprule
\textbf{Model} & \textbf{Split} & \textbf{Single-opinion} & \textbf{Opinion-update} & \textbf{Opinion-convergence} \\
\midrule
\texttt{Llama-3.1-8B} & Group & $1.807 \rightarrow 0.975$ & $1.732 \rightarrow 0.900$ & $0.804 \rightarrow 0.557$ \\
\texttt{Llama-3.1-8B} & Topic & $1.817 \rightarrow 1.040$ & $1.787 \rightarrow 0.988$ & $0.860 \rightarrow 0.718$ \\
\midrule
\texttt{Qwen3.5-4B} & Group & $1.527 \rightarrow 1.145$ & $1.477 \rightarrow 1.086$ & $0.807 \rightarrow 0.678$ \\
\texttt{Qwen3.5-4B} & Topic & $1.466 \rightarrow 1.174$ & $1.378 \rightarrow 1.183$ & $0.775 \rightarrow 0.720$ \\
\midrule
\texttt{Mistral-7B} & Group & $1.466 \rightarrow 1.251$ & $1.258 \rightarrow 1.176$ & $0.773 \rightarrow 0.692$ \\
\texttt{Mistral-7B} & Topic & $1.623 \rightarrow 1.284$ & $1.429 \rightarrow 1.253$ & $0.862 \rightarrow 0.694$ \\
\bottomrule
\end{tabular}
\label{tab:sft-cross-model-private-errors}
\end{table}

\section{Test-Set Beyond-Stance Diagnostics}
\label{app:beyond_stance}

Classical scalar-opinion baselines such as DeGroot can be competitive on stance trajectories, but they do not generate language and therefore cannot be evaluated for semantic or lexical alignment to human text. Figures~\ref{fig:sft_breadth_group_test_beyond_stance} and \ref{fig:sft_breadth_topic_test_beyond_stance} report this complementary test-set view on the group-generalization and topic-generalization splits for the base and SFT \llamabase{} models. The public target is the tweet-like opinion expression, and the private target is the final post-discussion private opinion. Across the eight combinations of test split, simulation setting, and opinion type (two splits $\times$ two simulation setups $\times$ two targets), the SFT show closer absolute length matching and higher ROUGE-L. Semantic similarity is also higher overall, with gains under \textit{Full Dynamics Simulation} and decreases under \textit{Next Message Prediction}.

\begin{figure}[tb!]
\centering
\includegraphics[width=\linewidth]{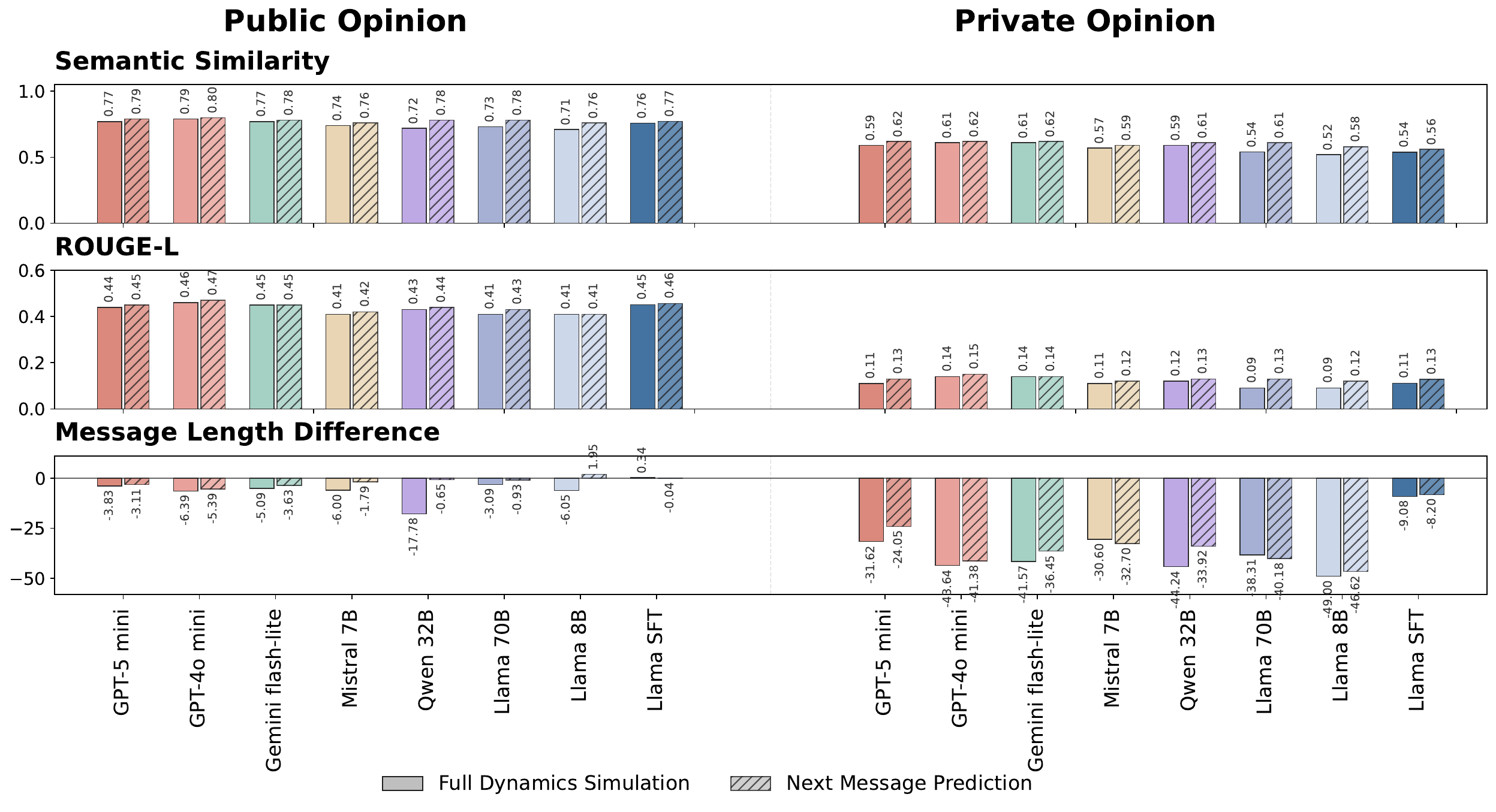}
\caption{Beyond-stance diagnostics for the group-generalization test setting. The left half corresponds to the public tweet target and the right half to the private post-discussion opinion target. The three displayed rows report semantic similarity, ROUGE-L, and message-length difference against the human reference.}
\label{fig:sft_breadth_group_test_beyond_stance}
\end{figure}

\begin{figure}[tb!]
\centering
\includegraphics[width=\linewidth]{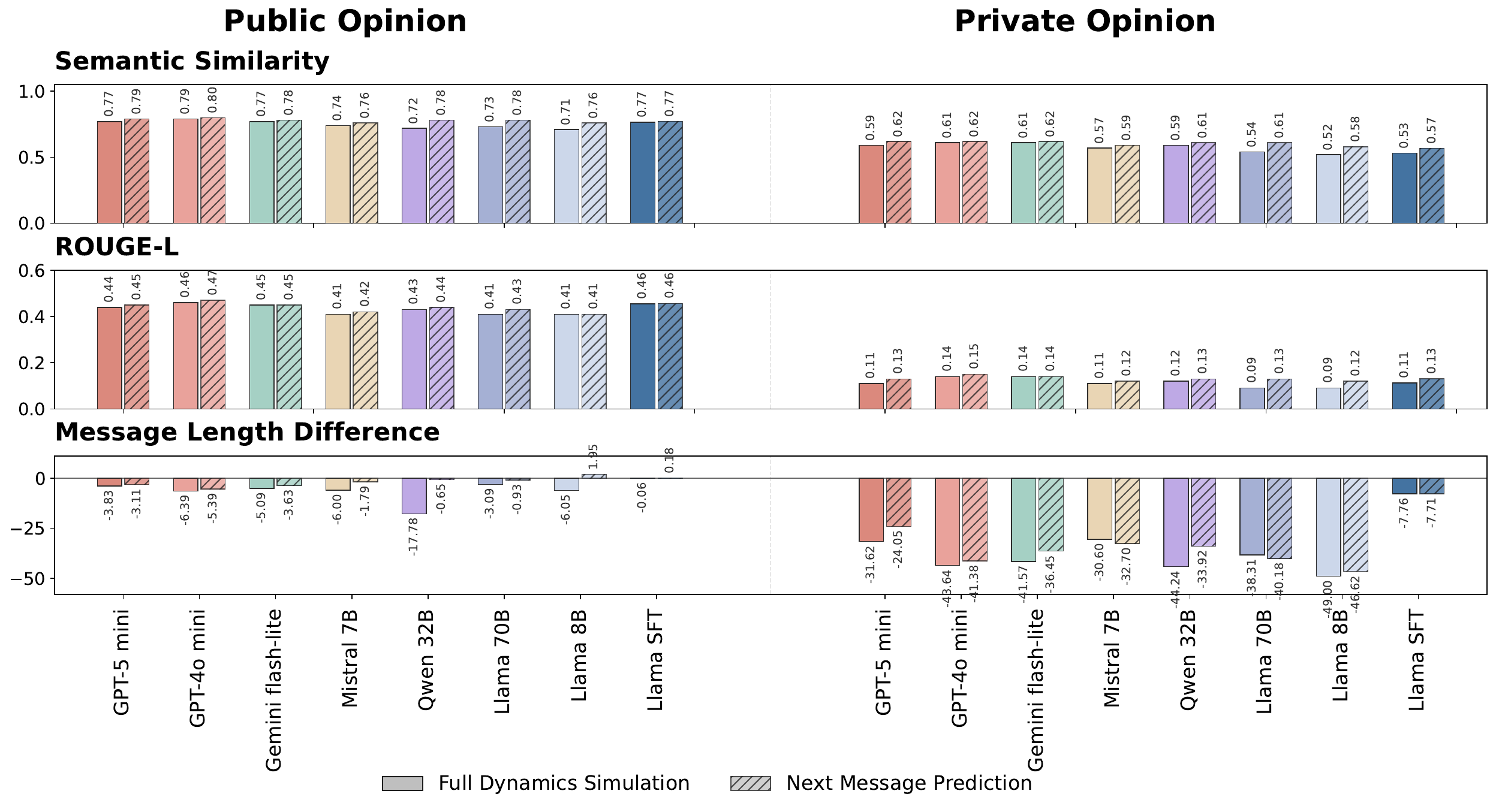}
\caption{The same beyond-stance diagnostics as in Fig.~\ref{fig:sft_breadth_group_test_beyond_stance} for the topic-generalization test setting.}
\label{fig:sft_breadth_topic_test_beyond_stance}
\end{figure}

\section{On-topic Rate Diagnostics}
\label{app:on_topic_rate_diagnostics}

The \llamabase{} SFT models achieve valid-response rates of 0.95 under both simulation setups on the group-generalization test split, and 0.95 and 0.96 under \textit{Full Dynamics Simulation} and \textit{Next Message Prediction}, respectively, on the topic-generalization test split. These rates measure response validity rather than opinion-dynamics fidelity.

\section{Judge Robustness Checks}
\label{app:judge_robustness}

We also checked the dependence on the annotator more directly. On a 20-group human-annotated subset, humans still converge less than the base \llamabase{} system under direct human labeling: tweet disagreement changes by $-0.12$ for humans versus $-0.24$ for the model, and private-opinion disagreement changes by $-0.11$ versus $-0.33$. A second-judge robustness check using Gemini-3-pro-preview also supports the over-convergence pattern, so the over-convergence result does not depend on one automatic annotator alone.

\section{Convergence Robustness and Mechanism Checks}
\label{app:convergence_robustness}

We evaluate convergence beyond $\Delta \mathrm{SD}$ because each discussion group contains only four participants, making cluster-style summaries unstable. On a 39-group subset spanning seven topics, the same over-convergence conclusion holds under two additional spread measures that remain well-defined in very small groups: average pairwise stance distance and stance range. For tweets, average pairwise stance distance changes by $-0.07$ for humans versus $-0.41$ for the base \llamabase{} rollout, and stance range changes by $-0.18$ versus $-0.72$. For private opinions, the corresponding changes are $+0.02$ versus $-0.44$ for pairwise distance and $-0.08$ versus $-0.69$ for range. Topic-level paired Wilcoxon tests across these seven topics again support the human--RPLA gap ($p=0.02$ for tweet convergence and $p=0.05$ for private-opinion convergence), so the main effect is not an artifact of using only within-group standard deviation.

These richer diagnostics also suggest that the difference is not just symmetric contraction around a fixed mean. On the same subset, average stance shift is modest and negative for humans ($-0.24$ for tweets and $-0.36$ for private opinions), whereas the representative base rollout shows a much flatter or even direction-reversing pattern ($-0.02$ and $+0.14$, respectively). Taken together with the pairwise-distance and range results, this indicates that the simulated groups are not merely noisier versions of human discussions; they compress disagreement too aggressively while also redistributing movement differently between public and private opinions.

We also ran three simple mechanism probes. First, the effect is not primarily a decoding artifact: for \llamabase{} in \textit{Full Dynamics Simulation}, changing the temperature leaves the public/private convergence effect intact, with tweet/private $\Delta \mathrm{SD}$ equal to $-0.30/-0.33$ at $T=0$, $-0.32/-0.40$ at the default decoding setting, and $-0.41/-0.27$ at $T=1$. Second, the effect is not mainly topic drift: the public on-topic rate changes only slightly from Round~1 to Round~3 ($-0.01$ for humans versus $-0.03$ for the base rollout). Third, removing the initial private opinion and first tweet from agent memory makes \texttt{gpt-4o-mini} much more convergent ($-0.28/-0.32 \rightarrow -0.57/-0.59$ for tweet/private $\Delta \mathrm{SD}$), which points away from a pure surface-form explanation and toward loss of explicit stance retention in \textit{Full Dynamics Simulation}.

\section{Statistical Tests and Confidence Intervals}
\label{app:main_finding_stats}

We report exploratory paired tests of three SFT improvements for \llamabase{} on the group-generalization test split: stance-prediction errors, transition-matrix fidelity, and the four text comparisons described in the main paper. Each paired comparison uses valid observations shared by the base and SFT models and a common human reference.

We obtain pointwise 95\% confidence intervals from 5,000 bootstrap resamples of topics, keeping all observations from a sampled topic together. Two-sided permutation tests use 20,000 swaps of Base and SFT labels, swapping all observations from a topic together. We compute Monte Carlo $p$-values with a plus-one correction and apply Holm correction within each comparison family: 12 stance-error comparisons, eight convergence-related comparisons, and four text comparisons. This correction controls family-wise error within each family. The tests assume exchangeability of model labels within paired topics under the null and are conditional on the selected checkpoints.

\paragraph{Stance-prediction errors.}
Table~\ref{tab:sft-paired-stance-tests} reports the reduction in each error for public and private opinions under both simulation settings. All 12 comparisons favor SFT, with Holm-adjusted $p\leq 0.0262$.

\begin{table}[tbp]
\caption{\textbf{Paired SFT improvements in stance prediction on the group-generalization test split.} Reduction is Base minus SFT error; positive values favor SFT. Brackets show pointwise 95\% topic-bootstrap confidence intervals. $p_{\mathrm{Holm}}$ corrects across all 12 comparisons. $N/G/T$ reports observations, groups, and topics; observations are opinion pairs, participants, and groups for single-opinion, opinion-update, and opinion-convergence errors, respectively.}
\centering
\small
\setlength{\tabcolsep}{4pt}
\begin{tabular}{@{}llrcr@{}}
\toprule
Opinion & Error & Reduction [95\% CI] & $p_{\mathrm{Holm}}$ & $N/G/T$ \\
\midrule
\multicolumn{5}{l}{\textit{Full Dynamics Simulation}} \\
Public & Single-opinion & $0.439\;[0.303, 0.577]$ & $0.0006$ & 841/125/100 \\
Public & Opinion-update & $0.542\;[0.358, 0.744]$ & $0.0006$ & 390/125/100 \\
Public & Opinion-convergence & $0.133\;[0.019, 0.249]$ & $0.0262$ & 121/121/96 \\
Private & Single-opinion & $0.868\;[0.667, 1.084]$ & $0.0006$ & 364/124/99 \\
Private & Opinion-update & $0.884\;[0.662, 1.099]$ & $0.0006$ & 329/122/98 \\
Private & Opinion-convergence & $0.262\;[0.106, 0.411]$ & $0.0036$ & 109/109/90 \\
\midrule
\multicolumn{5}{l}{\textit{Next Message Prediction}} \\
Public & Single-opinion & $0.247\;[0.142, 0.351]$ & $0.0006$ & 867/124/99 \\
Public & Opinion-update & $0.257\;[0.107, 0.410]$ & $0.0046$ & 402/124/99 \\
Public & Opinion-convergence & $0.185\;[0.084, 0.287]$ & $0.0032$ & 123/123/98 \\
Private & Single-opinion & $0.546\;[0.362, 0.733]$ & $0.0006$ & 359/122/97 \\
Private & Opinion-update & $0.565\;[0.377, 0.760]$ & $0.0006$ & 329/121/96 \\
Private & Opinion-convergence & $0.304\;[0.157, 0.452]$ & $0.0020$ & 108/108/89 \\
\bottomrule
\end{tabular}
\label{tab:sft-paired-stance-tests}
\end{table}

\paragraph{Transition-matrix fidelity.}
We compare normalized $L_2$ distance from the human transition matrix, $d(m,h)=\lVert C_m-C_h\rVert_F/\lVert C_h\rVert_F$, where $C$ is the pooled $6\times6$ transition-count matrix. Each model retains its own initial and final stance; the paired analysis requires four complete participants shared by the human, base, and SFT records in each group. The human matrix is held fixed during model-label swaps and resampled jointly with the model matrices during bootstrap. Table~\ref{tab:sft-paired-transition-tests} shows that SFT reduces this distance for both public and private opinions (Holm-adjusted $p=0.0004$ for each).

\begin{table}[tbp]
\caption{\textbf{Paired SFT improvements in transition-matrix fidelity under Full Dynamics Simulation.} Distances and reductions are measured relative to the same human transition matrix. Positive reductions favor SFT; brackets show pointwise 95\% topic-bootstrap confidence intervals. The two reported $p$-values retain Holm correction across the eight convergence-related comparisons. $N/G/T$ reports participant transitions, groups, and topics.}
\centering
\small
\setlength{\tabcolsep}{4pt}
\begin{tabular}{@{}llrcr@{}}
\toprule
Opinion & Base $\rightarrow$ SFT & Reduction [95\% CI] & $p_{\mathrm{Holm}}$ & $N/G/T$ \\
\midrule
Public & $0.716\rightarrow0.273$ & $0.443\;[0.184, 0.564]$ & $0.0004$ & 120/30/29 \\
Private & $0.879\rightarrow0.187$ & $0.693\;[0.458, 0.801]$ & $0.0004$ & 104/26/26 \\
\bottomrule
\end{tabular}
\label{tab:sft-paired-transition-tests}
\end{table}

\paragraph{Text alignment.}
Table~\ref{tab:sft-paired-text-tests} reports the four \textit{Full Dynamics Simulation} improvements described in Finding~4. Semantic similarity uses \texttt{jina-embeddings-v3}, and all paired outputs are scored with the same text-metric settings. For length, the statistic is the absolute mean word-count difference from human text. All four improvements have Holm-adjusted $p=0.0002$.

\begin{table}[tbp]
\caption{\textbf{Paired SFT improvements in text alignment under Full Dynamics Simulation.} Improvement is SFT minus Base for semantic similarity and ROUGE-L, and the reduction in absolute mean word-count bias for length. Brackets show pointwise 95\% topic-bootstrap confidence intervals. $p_{\mathrm{Holm}}$ corrects across all four comparisons. $N/G/T$ reports paired text records, groups, and topics.}
\centering
\small
\setlength{\tabcolsep}{4pt}
\begin{tabular}{@{}llrcr@{}}
\toprule
Opinion & Metric & Improvement [95\% CI] & $p_{\mathrm{Holm}}$ & $N/G/T$ \\
\midrule
Public & Semantic similarity & $0.041\;[0.033, 0.050]$ & $0.0002$ & 1095/124/100 \\
Public & ROUGE-L & $0.031\;[0.026, 0.037]$ & $0.0002$ & 1095/124/100 \\
Private & Length bias (words) & $42.795\;[41.006, 44.592]$ & $0.0002$ & 347/123/99 \\
Private & ROUGE-L & $0.022\;[0.014, 0.031]$ & $0.0002$ & 347/123/99 \\
\bottomrule
\end{tabular}
\label{tab:sft-paired-text-tests}
\end{table}

\paragraph{Confidence intervals in the figures.}
Figure~\ref{fig:heldout_breadth_group_posttrain_comparison} reports per-model 95\% confidence intervals from 1,000 bootstrap resamples of discussion groups. Figure~\ref{fig:heldout_breadth_group_partner_gap_scatter} reports 95\% percentile intervals from 1,000 resamples of the paired gap--change observations within each model. These intervals describe uncertainty in the displayed estimates. The convergence figures report standard errors: trajectory errors are computed across groups, and the $\Delta\mathrm{SD}$ error bars combine the initial and final standard errors in quadrature.

\section{SFT Results on the Topic-Generalization Split}
\label{app:topic_split_posttraining}

Figure~\ref{fig:heldout_breadth_topic_posttrain_comparison} reports the topic-generalization counterpart of the SFT comparison in Figure~\ref{fig:heldout_breadth_group_posttrain_comparison}.

\begin{figure*}[tbp!]
\centering
\includegraphics[width=\linewidth]{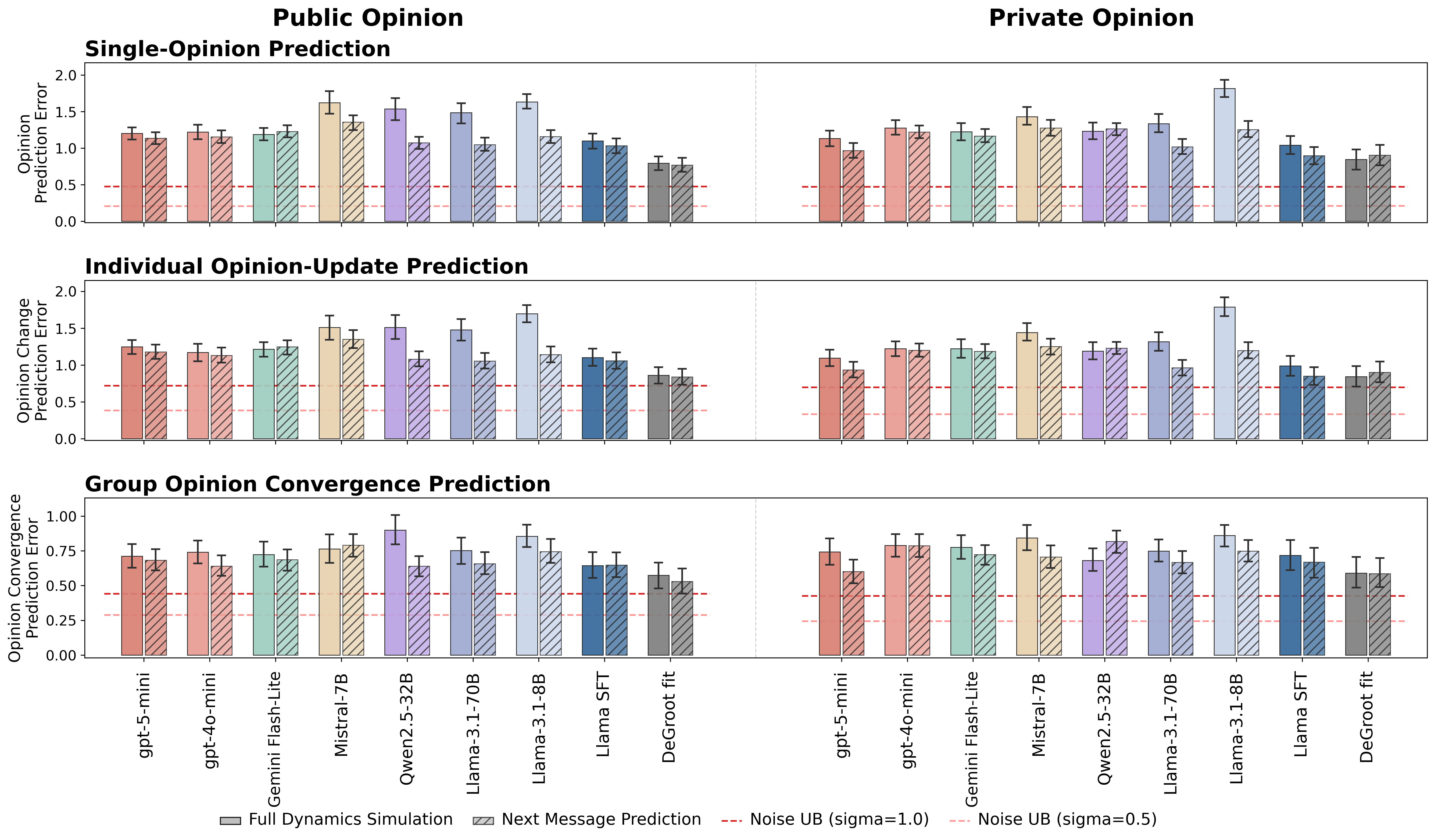}

\caption[Topic-generalization SFT comparison across the test-set stance-based error metrics.]{SFT comparison across the three stance-based error metrics on the topic-generalization test split. Rows show $E^{c}_{\mathrm{single}}$, $E^{c}_{\mathrm{update}}$, and $E^{c}_{\mathrm{conv}}$; the left column is public and the right column is private. Solid bars denote Full Dynamics Simulation, hatched bars denote Next Message Prediction, and lower values are better. Error bars are bootstrap 95\% confidence intervals; the reference baselines are defined in Sec.~\ref{subsec:eval}. Appendix~\ref{app:degroot_baseline} gives DeGroot-fitting details.}
\label{fig:heldout_breadth_topic_posttrain_comparison}
\end{figure*}

Figure~\ref{fig:heldout_breadth_topic_posttrain_metric_correlations} shows the same cross-model metric-correlation analysis on the topic-generalization test split. The same structure continues to hold there: single-opinion and opinion-update prediction remain almost collinear for both opinion types, while the relation between single-opinion prediction and group-convergence prediction is positive but noticeably weaker. This again indicates that the three stance-based targets move together, while group convergence remains a non-redundant summary of the collective opinion dynamics.

\begin{figure*}[tbp!]
\centering
\includegraphics[width=\linewidth]{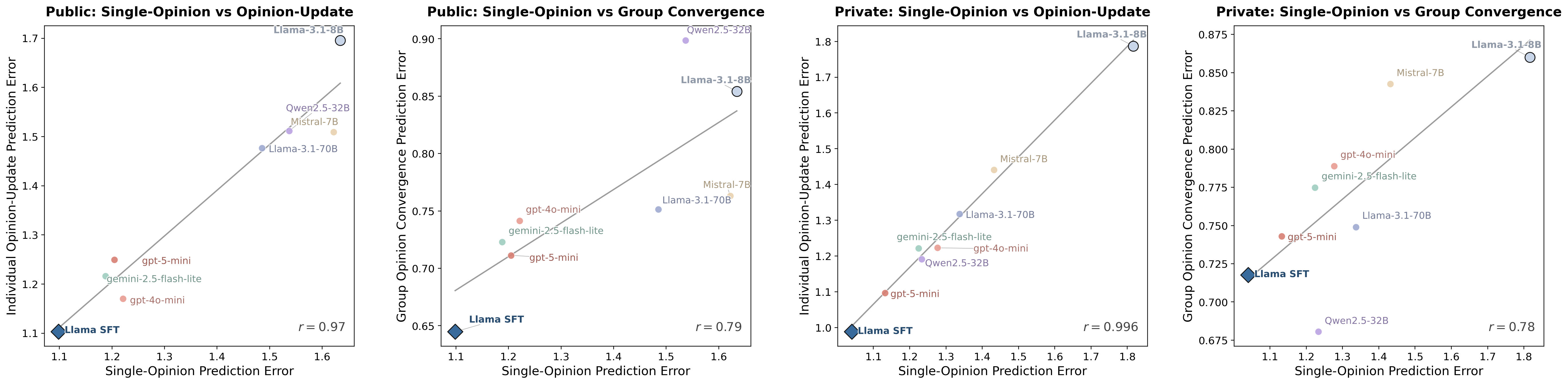}

\caption[Model-level correlations among the topic-generalization test-set stance-based error metrics.]{Model-level correlations among the three stance-based error metrics on the topic-generalization test split under Full Dynamics Simulation. The four scatterplots compare single-opinion prediction error against opinion-update and opinion-convergence prediction error for public and private opinions. Lower-left positions indicate lower error on both compared metrics.}
\label{fig:heldout_breadth_topic_posttrain_metric_correlations}
\end{figure*}

\section{Distributional Views of Opinion Shifts after SFT}
\label{app:posttrain_histograms}

Figure~\ref{fig:heldout_breadth_topic_posttrain_group_opinion_convergence} gives the topic-generalization counterpart of the opinion convergence analysis in Figure~\ref{fig:heldout_breadth_group_posttrain_group_opinion_convergence}, combining public/private convergence bars with the matching stance-SD trajectories. The private-opinion results show the base model's excess convergence and the movement toward the human reference after SFT; the public-opinion results provide the corresponding distributional trajectories.

\begin{figure*}[tbp!]
\centering
\includegraphics[width=\linewidth]{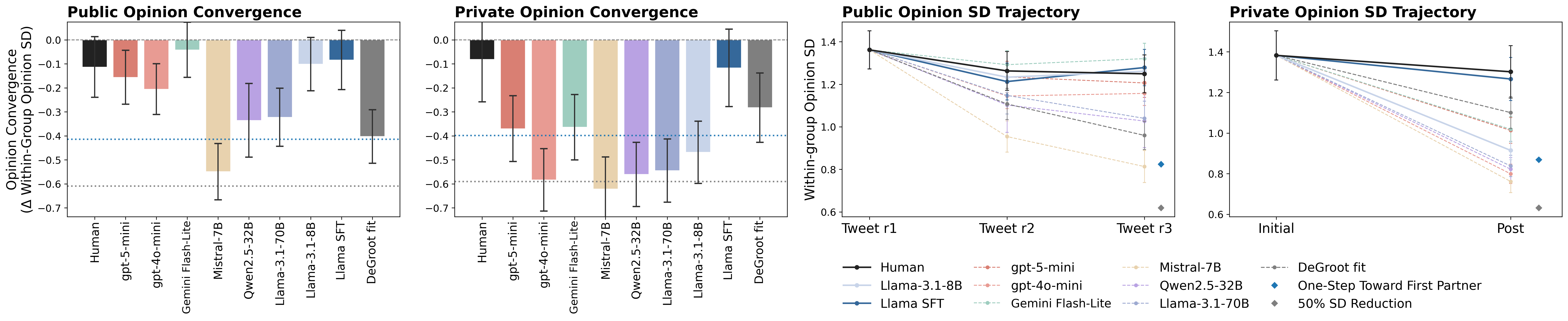}

\caption[SFT convergence summaries and SD trajectories on the topic-generalization test split.]{SFT convergence summaries and stance-SD trajectories on the topic-generalization test split under \textit{Full Dynamics Simulation}. The left two panels report public and private changes in mean stance SD ($\Delta\mathrm{SD}$); the right two show the matching stance-SD trajectories. More negative bars indicate stronger convergence. Error bars denote standard errors. Appendix~\ref{app:degroot_baseline} gives DeGroot-fitting details.}
\label{fig:heldout_breadth_topic_posttrain_group_opinion_convergence}
\end{figure*}

Figs.~\ref{fig:heldout_breadth_group_histograms_overview} and \ref{fig:heldout_breadth_topic_histograms_overview} provide the primary distributional view of these shifts. They summarize the same endpoint-versus-interior redistribution pattern on the group-generalization and topic-generalization splits while also situating the human/base/SFT comparison within the broader model family, including the other base LLMs and the fitted DeGroot baselines used for the test set. The same qualitative pattern continues to hold: compared with the human endpoint distributions, the base model is too centralized in intermediate bins, while SFT moves the simulated distributions back toward the endpoint-heavy human shape, especially for public opinions.

\begin{figure}[p]
\centering
\includegraphics[width=\linewidth]{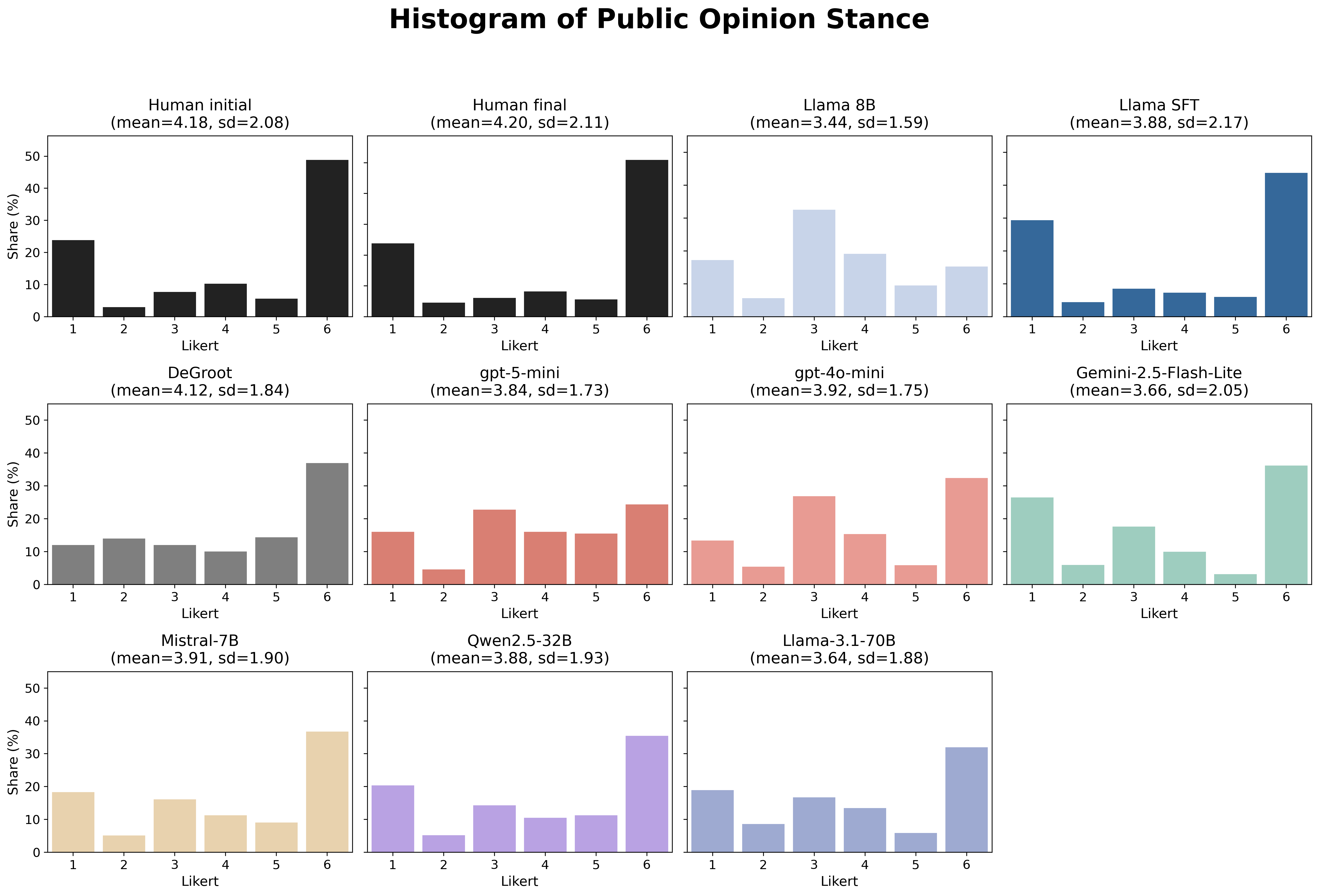}

\vspace{0.75em}

\includegraphics[width=\linewidth]{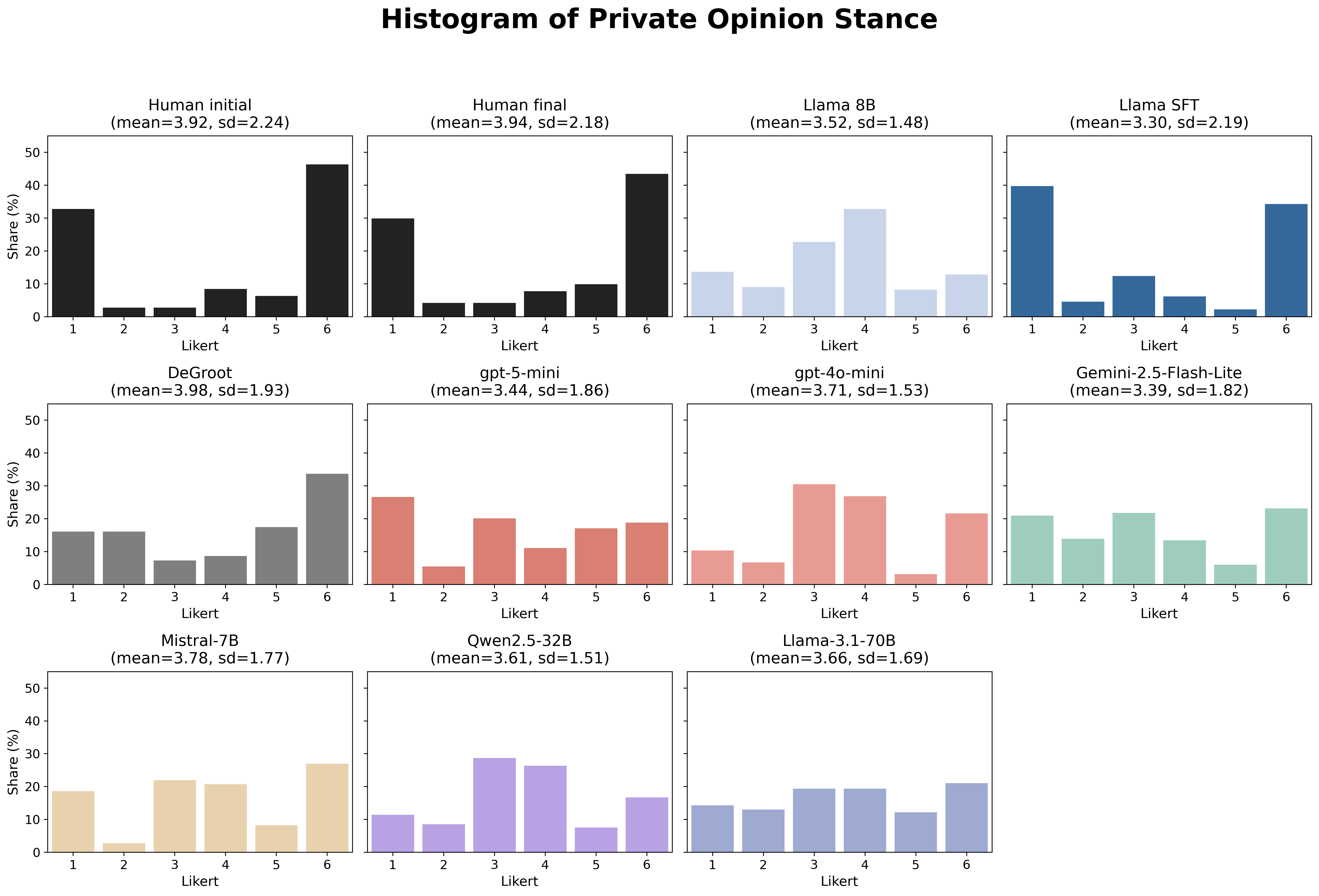}

\caption[Human and model public/private opinion histograms on the group-generalization test split.]{Opinion distributions for humans and all evaluated models on the group-generalization test split under \textit{Full Dynamics Simulation}. The public opinion histograms (top) and private opinion histograms (bottom) cover the broader model family, including the additional base LLMs. DeGroot baseline is fitted on the training set and evaluated on the test set.}
\label{fig:heldout_breadth_group_histograms_overview}
\end{figure}

\begin{figure}[p]
\centering
\includegraphics[width=\linewidth]{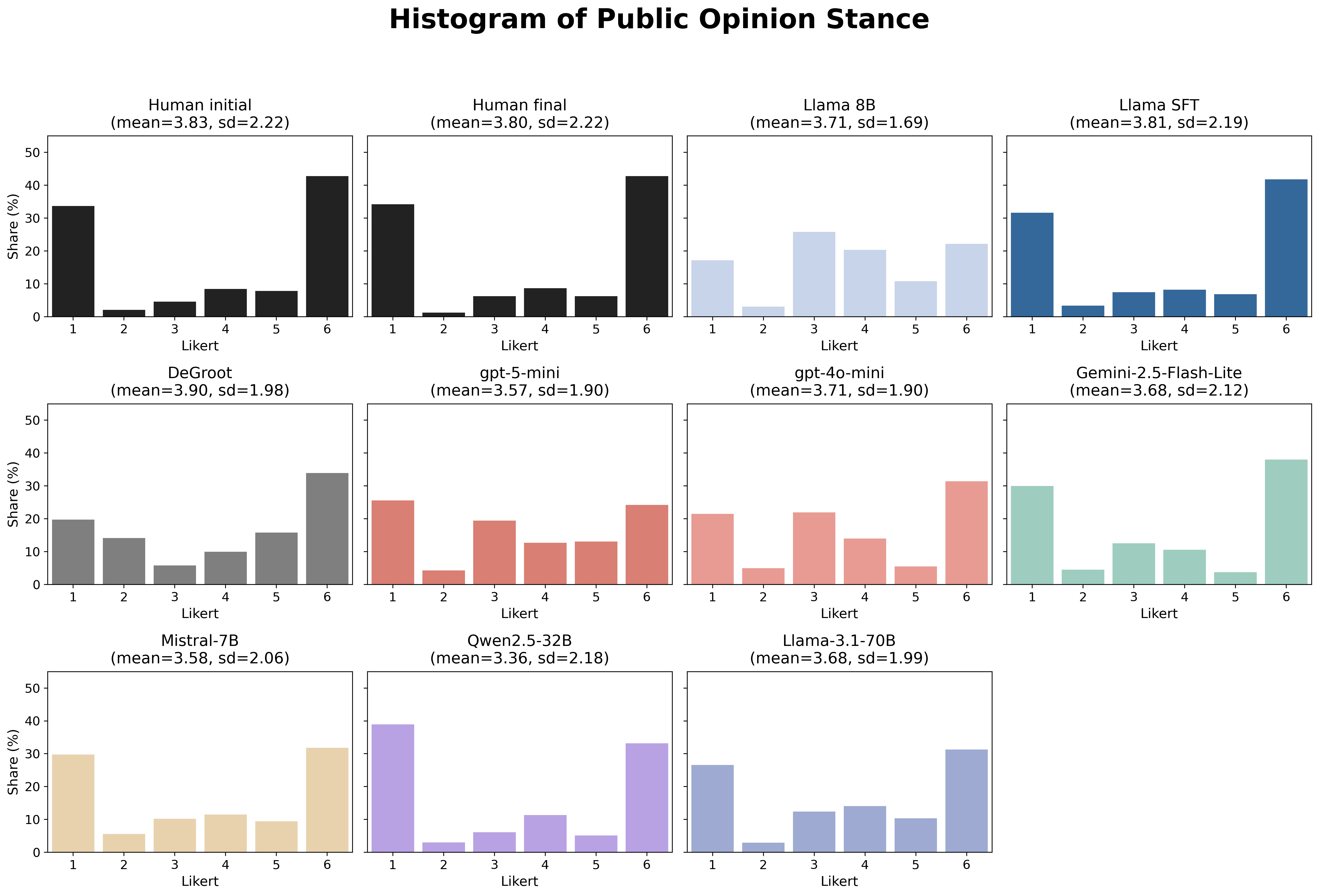}

\vspace{0.75em}

\includegraphics[width=\linewidth]{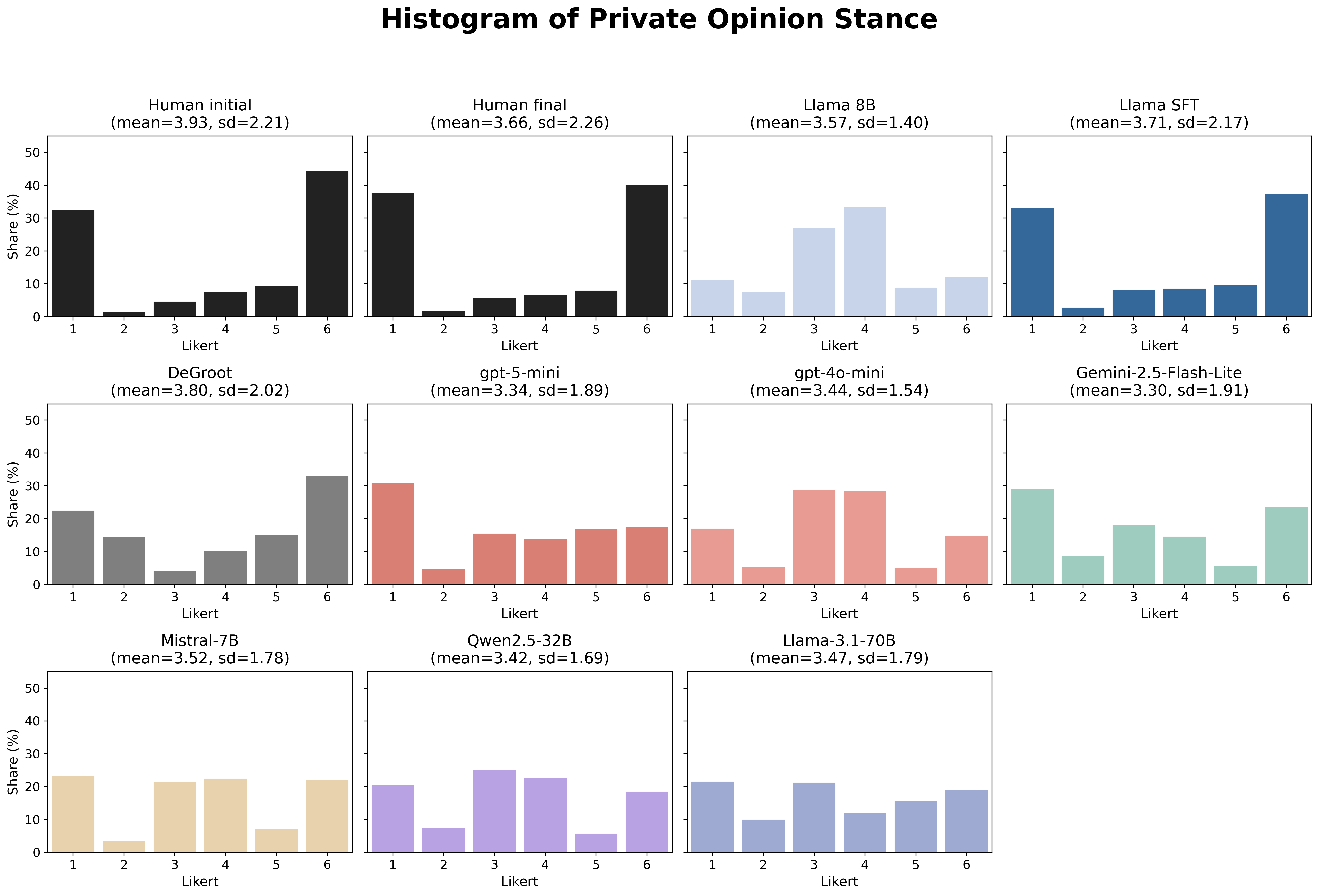}

\caption[Human and model public/private opinion histograms on the topic-generalization test split.]{Opinion distributions for humans and all evaluated models on the topic-generalization test split under \textit{Full Dynamics Simulation}. The public opinion histograms (top) and private opinion  histograms (bottom) cover the broader model family, including the additional base LLMs and the DeGroot baseline is fitted on the training set and evaluated on the test set.}
\label{fig:heldout_breadth_topic_histograms_overview}
\end{figure}

Figure~\ref{fig:heldout_breadth_topic_public_private_transition_matrices} shows the same start-to-end movement pattern on the topic-generalization split. The same qualitative story continues to hold: compared with humans, the base model shifts too much public and private mass away from the extreme bins and into intermediate categories, while SFT partially restores diagonal persistence at the extremes, especially for public opinions.

\begin{figure*}[tb!]
\centering
\includegraphics[width=\linewidth]{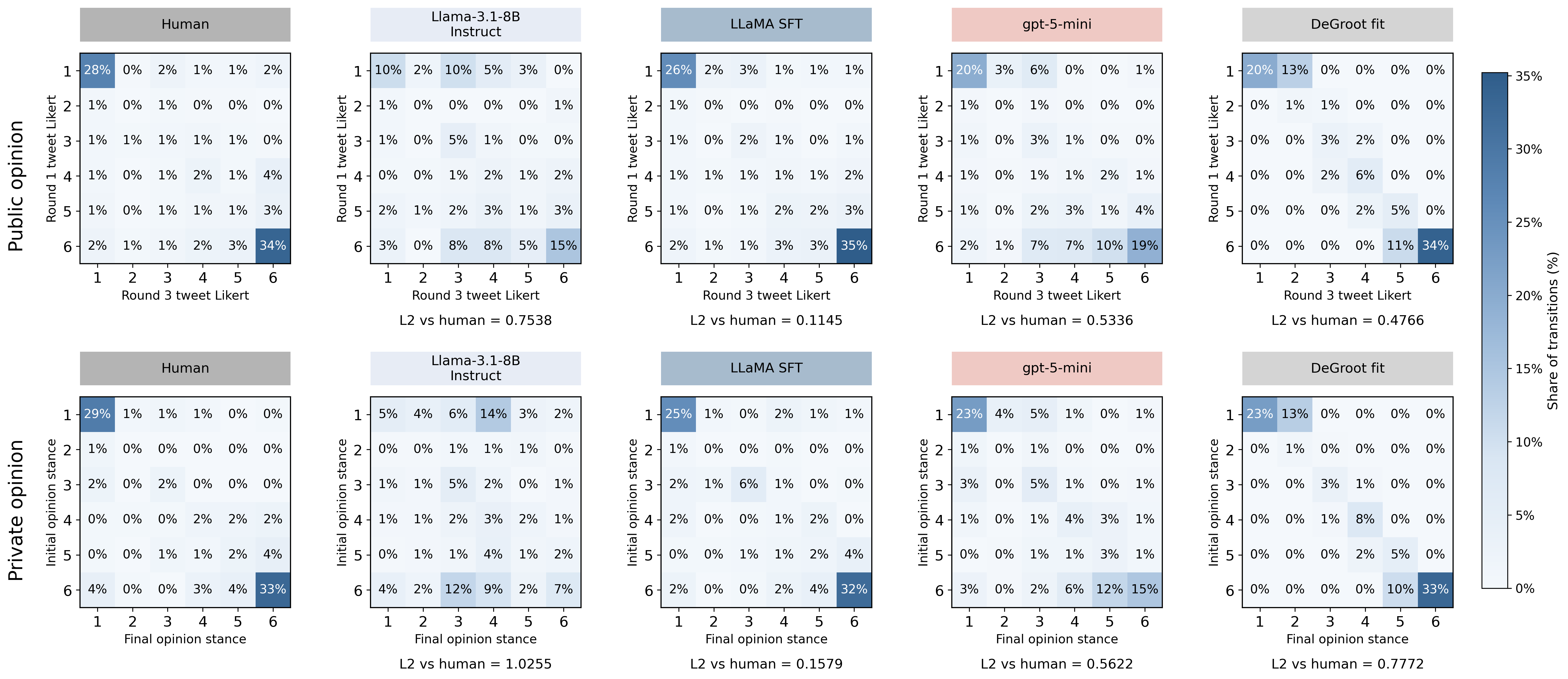}

\caption[Public/private transition matrices on the topic-generalization test split.]{Public/private transition matrices on the topic-generalization test split under \textit{Full Dynamics Simulation}. The top row shows public transitions from $S(\tau^1)$ to $S(\tau^3)$, and the bottom row shows private transitions from $o^{\text{init}}$ to $o^{\text{final}}$.Diagonal mass indicates persistence, off-diagonal mass indicates movement across stance bins, and the annotated normalized $L_2$ distances summarize deviation from the human reference.}
\label{fig:heldout_breadth_topic_public_private_transition_matrices}
\end{figure*}

\section{Topic-generalization Partner-Gap Analysis}

Figure~\ref{fig:heldout_breadth_topic_partner_gap_scatter} shows the same directional first-partner mechanism on the topic-generalization split. Positive associations indicate movement in the direction of the first partner's initial stance gap, while larger positive Spearman values indicate a stronger monotone partner-gap effect. The figure shows the same qualitative story as the group-generalization counterpart, with over-convergent baselines exhibiting the strongest partner-gap responsiveness.

\begin{figure*}[tb!]
\centering
\includegraphics[width=\linewidth]{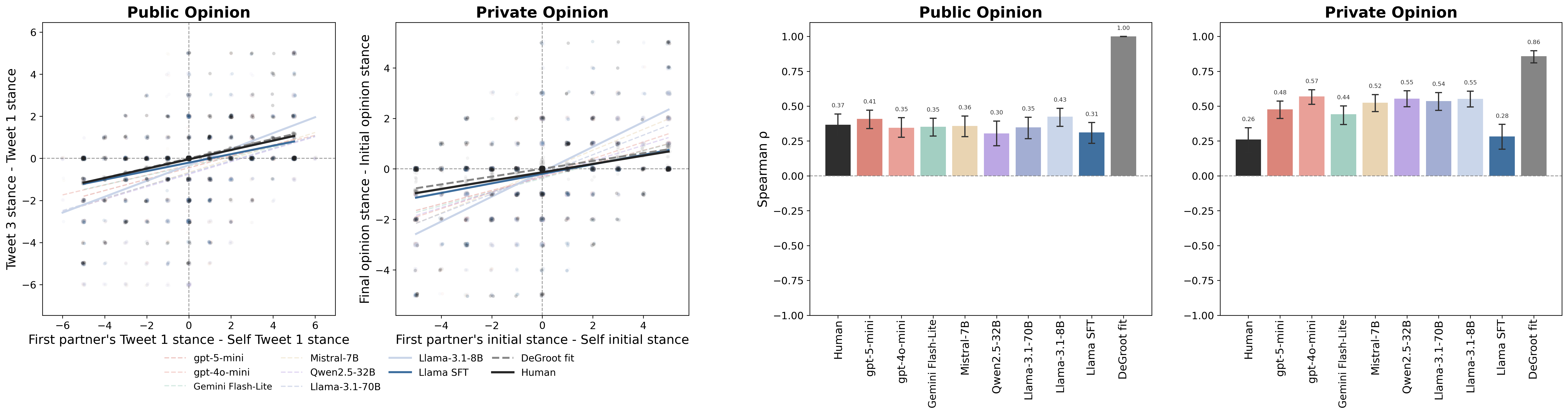}

\caption[Partner-gap responses on the topic-generalization test split.]{Partner-gap responses on the topic-generalization test split under \textit{Full Dynamics Simulation}. The left two panels plot changes in public and private stance against the first partner's initial gap; the right two report the matching Spearman $\rho$ summaries. Larger positive values indicate stronger movement in the direction of the partner's initial stance gap.}
\label{fig:heldout_breadth_topic_partner_gap_scatter}
\end{figure*}




\section{Supervised Fine-Tuning (SFT) Hyperparameters}
\label{app:sft_hyperparams}

\paragraph{LLaMA-3.1-8B-Instruct.} We fine-tune \texttt{Llama-3.1-8B-Instruct} for 5 epochs using parameter-efficient LoRA adaptation with rank $r=16$, $\alpha=16$, dropout $=0$, and no bias. LoRA targets the attention query, key, value, and output projections and the MLP gate, up, and down projections. The base weights use 4-bit NF4 quantization. We use learning rate $5\times10^{-5}$, per-device train/eval batch size 8, gradient accumulation steps 32 (effective training batch size 256), weight decay 0, and seed 42. Sequence packing is disabled. Loss is computed only on assistant completion tokens (completion-only loss). Training uses paged AdamW 8-bit optimization, a cosine scheduler with warmup ratio 0.1, bf16/tf32 mixed precision, and PyTorch compilation (Inductor backend, max-autotune mode). We evaluate at each epoch and select the best checkpoint by validation loss while retaining at most two checkpoints; the validation examples come from the corresponding test partition, as described in Appendix~\ref{app:sft}. The checkpoint from epoch 4 is selected for both splits. Implementation uses PEFT LoRA and the \texttt{trl} \texttt{SFTTrainer}.

\paragraph{Qwen3.5-4B and Mistral-7B-Instruct-v0.3.} For the cross-model replications, we use LoRA with Unsloth and the \texttt{trl} \texttt{SFTTrainer}, with rank $r=16$, $\alpha=16$, dropout $=0$, no bias, and rsLoRA disabled. We use paged AdamW 8-bit ($\beta_1=0.9$, $\beta_2=0.999$, $\epsilon=10^{-8}$), learning rate $5\times10^{-5}$, a cosine learning-rate schedule with 10\% warmup, weight decay 0.001, and gradient clipping at 1.0. Each run trains for 3 epochs with an effective batch size of 16, maximum sequence length 4,096, bf16/tf32 mixed precision, and seed 3407. Qwen runs use per-device batch size 4 and gradient accumulation 4 without sequence packing; Mistral runs use batch size 16 and accumulation 1 with packing. Qwen uses 4-bit NF4 weights with double quantization. The lowest validation loss occurs at epoch 2 for Qwen and epoch 1 for Mistral on both splits.

\paragraph{SFT inference.} Each adapter is served on its corresponding base model through vLLM's OpenAI-compatible server in bf16, with a maximum context length of 10,240 tokens. We sample with temperature 0.7, top-$p$ 0.9, repetition penalty 1.2, a minimum of 5 and maximum of 258 new tokens, and seed 1. Qwen thinking mode is disabled during inference.

\clearpage
\section{Example Conversations}
\label{app:example_chats}
We compare human, base \texttt{Llama-3.1-8B-Instruct}, and SFT conversations for the same focal participant and prompt. Each table aligns the initial opinion, the round-by-round exchanges, and the final opinion so that differences across the three versions can be read side by side.

\newenvironment{examplecomparison}[1][]{%
\begingroup
\small
\setlength{\tabcolsep}{3pt}
\renewcommand{\arraystretch}{1.15}
\begin{longtable}{@{}>{\raggedright\arraybackslash}p{0.09\textwidth}>{\raggedright\arraybackslash}p{0.28\textwidth}>{\raggedright\arraybackslash}p{0.28\textwidth}>{\raggedright\arraybackslash}p{0.28\textwidth}@{}}
\if\relax\detokenize{#1}\relax
\else
\multicolumn{4}{@{}p{\textwidth}@{}}{{#1}} \\
\midrule
\fi
\toprule
\textbf{Stage} &
\textbf{Human} &
\textbf{Base Llama} &
\textbf{SFT Llama} \\
\midrule
\endfirsthead
\toprule
\textbf{Stage} &
\textbf{Human} &
\textbf{Base Llama} &
\textbf{SFT Llama} \\
\midrule
\endhead
\midrule
\multicolumn{4}{r}{Continued on next page} \\
\endfoot
\bottomrule
\endlastfoot
}{%
\end{longtable}
\endgroup
}

\newcommand{\examplemeta}[1]{%
\multicolumn{4}{@{}p{\textwidth}@{}}{{#1}} \\
\midrule
}

\newcommand{\examplerow}[4]{%
\textbf{#1} &
#2 &
#3 &
#4 \\
\midrule
}

\newcommand{\examplesharedrow}[2]{%
\textbf{#1} &
\multicolumn{3}{@{}p{0.84\textwidth}@{}}{{#2}} \\
\midrule
}

\subsection*{Example Conversation: News Reporting}
This example compares the same discussion topic under the human record, the base model, and the supervised fine-tuned model. The base model shifts into a reform-oriented endorsement of mainstream news, whereas the supervised fine-tuned model remains much closer to the human participant's sustained distrust.

\textbf{Topic under Discussion:} News organizations are doing good at reporting news about government leaders and officials.

\textbf{Reading guide:} ``Round 1'', ``Round 2'', and ``Round 3'' denote the three discussion rounds. ``Partner A'', ``Partner B'', and ``Partner C'' denote the first, second, and third partners seen by the focal participant. Rows labeled ``tweet'' reproduce the public tweet-like posts at the start of a round. For comparison, Round 2 is shown through the sixth reply; the final opinions are from the full conversations. Stance values run from 1 (certainly disagree) to 6 (certainly agree).

\begin{examplecomparison}[Qualitative comparison of the human record, base \llamabase, and supervised fine-tuned model on a news-reporting discussion topic.]
\examplesharedrow{Initial opinion}{\textbf{Stance 1.} Mainstream news does not do a good job of this. They are very biased and would prefer to promote division and fear. The big news organizations are owned by billionaires who benefit from this. It is not a trustworthy process.}
\examplesharedrow{Round 1 tweet 1}{\textbf{Me $\rightarrow$ Partner A:} I don't trust major news organizations. They are owned by billionaires who serve their own self-interest. They stroke division and are full of biases.}
\examplesharedrow{Round 1 tweet 2}{\textbf{Partner A $\rightarrow$ Me:} Hi. News organizations need to do a lot of work. News organizations need to ask a lot of questions to the government leaders.}
\examplerow{Round 1 reply 1}{\textbf{Partner A $\rightarrow$ Me:} We need to uplift the news organizations.}{\textbf{Partner A $\rightarrow$ Me:} @56e5s I feel ya! Those billionaire owners definitely shape what gets reported... but isn't there some accountability measures to ensure they're held somewhat responsible? I mean we still need reliable info about our gov't leaders, right? How do we get real news amidst all these agendas?}{\textbf{Partner A $\rightarrow$ Me:} Hi. Do we get too much biased coverage? Or do news orgs just focus on sensationalism?}
\examplerow{Round 1 reply 2}{\textbf{Me $\rightarrow$ Partner A:} They need to ask tough questions and not just be mouthpieces for the government. The press should be independent, not a propaganda machine.}{\textbf{Me $\rightarrow$ Partner A:} I'm glad someone acknowledges those "billionaire owners" influence the way things are presented. And yes, accountability might help - public watchdog groups \& investigative journalists can provide alternative perspectives to counterbalance the agenda-driven coverage. However, relying solely on them leaves us wondering which sources aren’t compromised too}{\textbf{Me $\rightarrow$ Partner A:} Yes. Main stream media tries to sell us all an agenda which serves billionaire interests rather than our country's well-being. Sensationalism definitely plays into it also though}
\examplerow{Round 1 reply 3}{\textbf{Partner A $\rightarrow$ Me:} We can support local new organizations.}{\textbf{Partner A $\rightarrow$ Me:} I think transparency is key here. Maybe having some independent fact-checking orgs partnered directly with news outlets could help minimize bias while maintaining factual accuracy. Also, engaging more diverse communities \& voices to create journalism might add balance since it'd involve various viewpoints outside just corporate interests}{\textbf{Partner A $\rightarrow$ Me:} I believe there needs to be more scrutiny on how money influences what gets covered. And honestly, even non-profit journalism isn't immune to those pressures. There needs to be some kind of transparency around funding sources.}
\examplerow{Round 1 reply 4}{\textbf{Me $\rightarrow$ Partner A:} I support small news orgs and independent journalists. I support my local public news outlets.}{\textbf{Me $\rightarrow$ Partner A:} I like where this discussion is headed. Having an actual mix of voices makes me feel better, knowing people will tell stories that represent entire neighborhoods. One thing concerns me though; when 'corporate interests' aren't at stake anymore – how often is funding tied into local community newspapers or smaller online forums trying to report honestly about local politics and events}{\textbf{Me $\rightarrow$ Partner A:} Yes. Non-profits especially should try harder to stay neutral since they claim to be unbiased.}
\examplerow{Round 1 reply 5}{\textbf{Partner A $\rightarrow$ Me:} Okay. So, we need to get the local new organizations involved.}{\textbf{Partner A $\rightarrow$ Me:} What happens if/when even the supposedly locally funded/neutral publications start getting caught up in similar conflicts-of-interest due to financial pressures? Are we stuck in a cycle? Shouldn't there also be checks in place that reward quality truthful reporting despite budget limitations? (I am taking account my previous engagement with you.) Before answering the prompt I realized I felt much stronger in disagreement than before. The main reason behind this shift is because now believe that the role of billionaire ownership does matter greatly in determining what gets}{\textbf{Partner A $\rightarrow$ Me:} I think thats why local journalists tend to be more trustworthy - because they're less beholden to national networks' agendas. What else makes sense here?}
\examplerow{Round 2 tweet 1}{\textbf{Me $\rightarrow$ Partner B:} Major news organizations have failed us. They are owned by billionaires who are looking out for their own interests. They have become propaganda machines for the government and not true journalists.}{\textbf{Me $\rightarrow$ Partner B:} Despite initial skepticism,conversations w/those concerned w/truthful reporting helped clarify issues. Main takeaway: while major corps exert significant control over media narrative, perhaps NOT insurmountable obstacle, especially when balanced alongside efforts like transp, divers voices, +independent fact check. Would love to hear ur take @5c0dB}{\textbf{Me $\rightarrow$ Partner B:} I believe mainstream media pushes an agenda set forth by billionaire owners which doesn't always align with the best interest of our country. Too many times stories come across as divisive and untrustworthy due to these factors.}
\examplerow{Round 2 tweet 2}{\textbf{Partner B $\rightarrow$ Me:} I cannot agree. I dont believe anything they say, at all. Remember the last 4 years, where they never mentioned what was painfully obvious to everyone else. That Biden was damn near retarded. But it was crickets. Not one mention}{\textbf{Partner B $\rightarrow$ Me:} I’ve been skeptical about news orgs lately, constantly focusing on Trump scandals while barely covering Biden’s glaring flaws. Now I’m rethinking my stance. Is the system rigged or just trying to cope with too much info coming in simultaneously?}{\textbf{Partner B $\rightarrow$ Me:} False. Look back at how they tried to silence us after the election of trump. None of those democrats wanted him in office because they knew he had nothing to hide. That being said, maybe theyre trying to report more accurate info now since theres so much going on}
\examplerow{Round 2 reply 1}{\textbf{Me $\rightarrow$ Partner B:} They are very pro-Trump and owned by people who he serves best so.}{\textbf{Me $\rightarrow$ Partner B:} Well said @5c0d B, regarding the attention span for both Pres Trump \& Pres Biden's controversies, i've noticed major networks prioritizing certain topics to reach higher ratings rather than giving equal time to pressing issues across party lines. Don't know if its part of being 'rigged', maybe its simply sensationalism playing on people's emotions rather than educating us}{\textbf{Me $\rightarrow$ Partner B:} I understand your view. But look closely at Trump and his administration's actions against journalists. He has done a great disservice to them and made them fearful of speaking truthfully. Many were silenced before he took office too.}
\examplerow{Round 2 reply 2}{\textbf{Partner B $\rightarrow$ Me:} Yes fox is right wing, but at least when the left does something good, which is almost never, the do mention it}{\textbf{Partner B $\rightarrow$ Me:} After reading yours and 56e5ee’s posts, I still feel that sometimes the media does tend to lean heavily in favor of Democrats but that doesn’t mean each outlet cannot choose alternative paths moving forward}{\textbf{Partner B $\rightarrow$ Me:} tumultuous times. Theres plenty of proof that the parents companys were owned by china}
\examplerow{Round 2 reply 3}{\textbf{Me $\rightarrow$ Partner B:} No they don't lol.}{\textbf{Me $\rightarrow$ Partner B:} I couldn’t help notice how you mention leaning towards Democrats, but avoiding addressing whether Republican Party politicians receive equivalent scrutiny under the same microscope. Have you ever wondered why Republicans seem to slide free from harsher criticism for policy failures and controversy while Democrats face disproportionate negative press? (Note: Responded assuming that my view got slightly shifted during the conversations with others)}{\textbf{Me $\rightarrow$ Partner B:} tumultous indeed. But i'm saying most media outlets aren't willing to speak harsh truths despite having evidence.}
\examplerow{Round 2 reply 4}{\textbf{Partner B $\rightarrow$ Me:} MSNBC pro trump. CNN pro trump, cbs, nbc, abc, you are blind or a moron}{\textbf{Partner B $\rightarrow$ Me:} I appreciate your candor, 56e5eee. I find myself agreeing partly that although individual situations vary, scrutiny might indeed focus disproportionately onto Democratic Party figures compared to Republicans, possibly creating unfair impressions on various issues. Do you think that establishment of watchdog media entities and stronger checks could promote a more fair balance within news distribution in society?}{\textbf{Partner B $\rightarrow$ Me:} tumultuous yes. especially under trumps rule. And even though they havent spoken badly about democrats, you cant count on anything they publish either}
\examplerow{Round 2 reply 5}{\textbf{Me $\rightarrow$ Partner B:} Fox is a propaganda machine for the GOP. Wow. That was really helpful thanks.}{\textbf{Me $\rightarrow$ Partner B:} One aspect that resonates deeply with me now after engaging with multiple individuals sharing differing opinions is the need for constructive dialogue surrounding accountability in the journalistic realm. Notably, you suggest incorporating watchdog media entities and implementing robust internal mechanisms to achieve fairness in newscasting. That brings forth several interesting possibilities, such as establishing a publicly-owned non-profit entity operating in parallel, or revamping existing frameworks to make unbiased truth-seeking primary objectives. By adopting these proposals together with already discussed strategies involving greater diversity in editorial teams}{\textbf{Me $\rightarrow$ Partner B:} tumultuous yes. especially under trumps rule. And even though they haven't spoken badly about democrats, you can't count on anything they publish either Exactly!}
\examplerow{Round 2 reply 6}{\textbf{Partner B $\rightarrow$ Me:} fox is the only channel that mentions anything good about republicans}{\textbf{Partner B $\rightarrow$ Me:} When we start exploring ways to establish stronger watchdog institutions, making sure we aren’t putting partisan interests above factual accuracy becomes incredibly important to prevent unintended biases within newsgathering processes. Considering your thoughts, wouldn’t allowing independent verification and constant reassessment of reported events assist increase confidence and faithfulness among populace toward trustworthy sources? On top of that if watchdog entities operate directly overseen the general populace through local participatory governance models to guarantee diverse community input – strengthening overall democratic foundations – I genuinely hope}{\textbf{Partner B $\rightarrow$ Me:} tumultuous yes. especially under trumps rule. And even though they havnt spoken badly about democrats, you cant count on anything they publish either}
\examplerow{Round 3 tweet 1}{\textbf{Partner C $\rightarrow$ Me:} I like to watch the news and I think the news organizations do a good job for the most part in reporting what's going on}{\textbf{Partner C $\rightarrow$ Me:} After conversing w/@5c0d\&B\&@655fr last time about press accountability, I began reassessing what drives perceptions around fair/journalistic representation. Through constructive dialogue, perspectives shifted \& insights emerged. Today, I reaffirm my stance on credible news orgs making progress, ensuring factually-driven reports stay grounded in investigat Now it is my turn to converse with 56e5E}{\textbf{Partner C $\rightarrow$ Me:} Overall, i believe that news orgainzations do a fairly good job at covering important newsworthy events. Of course, not all of them, however certainly the majority. I do wonder whether they're able to delve deep enough without oversimplifying complex political issus.}
\examplerow{Round 3 tweet 2}{\textbf{Me $\rightarrow$ Partner C:} News organizations are not doing a good job. They are very biased. The big media companies push the agendas of the billionaires who own them. Only small, independent outlets are reporting the truth.}{\textbf{Me $\rightarrow$ Partner C:} After chatting with folks on Twitter, I've had a change of heart regarding major news corporations controlling narrative around gov't leaders. Their influence shouldn't overshadow genuine reporting. Transparency, diverse perspectives, \& partnerships with independent fact-checkers could help strike a healthy balance between profit motive \& truthful storytelling \#newsorganization \#accountability}{\textbf{Me $\rightarrow$ Partner C:} I believe major mainstream news outlets push an agenda controlled by billionaires which often conflicts with our nation's true best interest. Stories presented seem divisive and untrustworthy. Local newspapers might offer better insights and factual reports.}
\examplerow{Round 3 reply 1}{\textbf{Partner C $\rightarrow$ Me:} Yes there are definitely agendas getting pushed, but for the volume of information they have to disseminate I think that we get quite a bit of information considering. what news channels do you get your news from? I like to get a lot of my news from tik tok although i know thats not very conventional}{\textbf{Partner C $\rightarrow$ Me:} I appreciate your thoughtful consideration on how transparency, diverse perspectives \& collaborations with independent fact checkers might foster healthier dynamics between profits motives \& actual truth-telling! Partnerships with fact-checkers, in particular, stand out to me as promising measures aimed at bolstering credibility through third-party verification while preventing exploitative use of info intended to sway masses via manipulation, rather than solid research. Now, addressing this notion of control exerted by huge corporations, perhaps a hybrid model incorporating multiple small-to-medium sized publications}{\textbf{Partner C $\rightarrow$ Me:} I tend to lean towards local newspaper sources myself since they appear to remain less biased. Have you experienced any trustworthy main stream news outlets lately?}
\examplerow{Round 3 reply 2}{\textbf{Me $\rightarrow$ Partner C:} We do get more information than ever before. But the information is often biased and not all of it is helpful. The important stories get lost in the big mix of news.}{\textbf{Me $\rightarrow$ Partner C:} Yes, the partnership idea sounds encouraging as well. Small-to-medium-sized publications, run independently with minimal interference from large corporate influences, could allow reporters to pursue stories driven by curiosity instead of ratings boosts. Even the prospect of networking opportunities made possible by connections established through shared membership in regional journalist associations allows me speculate on what beneficial effects could arise from collaborative efforts between these outfits when given space to flourish beneath modest budgets. Considering it might facilitate grassroots-level understanding that promotes increased participation from everyday citizens within the civic sphere}{\textbf{Me $\rightarrow$ Partner C:} I've seen some decent ones recently but still struggle to find reliable news sources. Some outlet did cover the Biden documents scandal properly last week.}
\examplerow{Round 3 reply 3}{\textbf{Partner C $\rightarrow$ Me:} I would love to see more investigative journalism}{\textbf{Partner C $\rightarrow$ Me:} 1 The process of weighing arguments and refining opinions certainly yields significant shifts for individuals navigating complex subject matters like governance through news agencies. A great emphasis lies within nurturing trust among publics worldwide–it necessitates balancing openness alongside maintaining diligence in sourcing materials verified properly. When analyzing overall patterns observed among various actors involved inside mass communications realms, a consensus appears present concerning prioritizing solutions focused mainly on interconnecting existing networks already serving community needs throughout specific regions to encourage inclusive voices \& authentic approaches thereby minimizing susceptibility to pressures}{\textbf{Partner C $\rightarrow$ Me:} I'd love to hear what specific outlets you found to be trustworthy?}
\examplerow{Round 3 reply 4}{\textbf{Me $\rightarrow$ Partner C:} I get my news from a variety of sources. I like supporting small independent journalists and outlets. Yes.. I agree. The major news media like to talk but don't do a lot of investigating these days.}{\textbf{Me $\rightarrow$ Partner C:} With each of these dialogues, it seems my outlook has evolved notably, transitioning from doubt and frustration concerning major corporation-controlled newscasts to seeing glimmers of potential reforms capable of bringing positive transformations. Among numerous ideas floated, incorporating strong watchdog oversight entities designed to verify impartiality across various channels demonstrates promise when paired with empowering independent voices within the publishing industry itself. These proactive steps offer hope for cultivating deeper understandings fostering resilience amongst engaged stakeholders interested in seeking verifiable facts – vital qualities urgently required amidst prevalent}{\textbf{Me $\rightarrow$ Partner C:} A few smaller, more regional papers showed promise but bigger names like CNN still feel dishonest in my eyes. A New York Times article caught attention last week regarding U.S. Treasury Secretary Yellen’s claims vs reality}
\examplerow{Final opinion}{\textbf{Stance 1.} News organizations are not doing a good job of reporting news about government leaders and officials. Big news media is owned by self-centered billionaires who do not question the government. Investigative journalism is not left to small independent news organizations that don't hit the mainstream.}{\textbf{Stance 6.} Since ending these conversations, I strongly agree that news organizations are doing good at reporting news about government leaders and officials. Conversations with fellow tweeters shed light on promising pathways like incorporation of transparent practices, promoting diverse voices in journalism, collaborating with reputable fact-checking services, supporting smaller independent publishers working away from major corporate influence, developing thorough external audits checking integrity within larger conglomerates etcetera. All these strategic initiatives collectively enhance prospects for offering sincere information about influential authorities ultimately leading accountable civic knowledge spread nationwide}{\textbf{Stance 1.} I disagree with the idea that mainstream news organizations are doing a good job reporting news about government leaders and officials. Their bias and sometimes blatant promotion of misinformation make me skeptical when consuming news today. Trusting credible sources feels increasingly challenging given the state of journalism.}
\end{examplecomparison}

\section{LLM Usage Disclosure}
\label{app:llm-disclosure}

We used LLMs, specifically ChatGPT, solely to aid in polishing the writing. All original ideas, experiment design, analyses, and initial drafts were produced by the authors. The LLM was used solely to refine phrasing, improve clarity, and ensure grammatical correctness, but it did not contribute novel content or edits beyond language refinement.

\section{User Interface for the Human Experiment}
\label{app:human_exp_ui}

We provide screenshots of the user interfaces that participants encountered during the multi-player experiment (Figures~\ref{fig:consent_form}--\ref{fig:demographic3}). Each figure illustrates one stage of the experimental process, from onboarding to the demographic survey.

In our multi-player experiment, participants were recruited through Prolific. At first, they would be navigated to a \textbf{consent form} outlining the study’s procedures, duration, compensation, and confidentiality (Figure~\ref{fig:consent_form}). Those who agreed to participate were then shown a \textbf{general introduction} explaining the flow of the task (Figure~\ref{fig:intro}). Specifically, they were informed that they would be given a statement (e.g., “Smoking cigarettes causes cancer”) and asked to write a short post as their initial opinion. They were also told that the experiment would last approximately 25--30 minutes and involve a sequence of conversations with other participants. 

After reviewing the instructions, participants began by reporting their \textbf{initial opinion} on the assigned discussion statement and selecting a slider value to indicate the extent to which they agreed or disagreed with the statement (Figure~\ref{fig:initial}). The core of the study consisted of \textbf{three rounds of interaction}, each following the same structure (Figures~\ref{fig:round1a}--\ref{fig:round3b}). First, participants were informed of who they would chat with and then were directed to write a short tweet-like post summarizing their current stance. Next, they would have twenty seconds to prepare for a dyadic conversation with a different partner. This ensured that each participant was exposed to all other perspectives across rounds. 

At the end of the third round, participants submitted a \textbf{post opinion} in order for us to capture how their stance evolved during the course of the discussions (Figure~\ref{fig:post}). Finally, they completed a detailed \textbf{demographic survey} (Figures~\ref{fig:demographic1}--\ref{fig:demographic3}), after which they were compensated at the study rate reported in Section~\ref{subsec:recruitment}. 


\begin{figure}[htb!]
\centering
\begin{tcolorbox}[width=\linewidth, colback=gray!5, colframe=black, boxrule=0.5pt, sharp corners, title=\textbf{Onboarding Consent Form}]
Please read this consent agreement carefully before deciding whether to participate in this experiment.

What you will do in this research: You will play a series of communication games with other participants.

Time required: This study will take approximately twenty minutes.

Purpose of the research: The purpose is to understand how conversations evolve in a networked community.

Risks: There are no anticipated risks associated with participating in this study.
The effects should be comparable to viewing a computer monitor and using a mouse for the duration of the experiment.

Compensation: You will receive course credits for completing the experiment.

Confidentiality: Your participation in this study will remain confidential. No personally identifiable information will be collected. Your anonymous data may be shared with other researchers and used in future projects.

Participation and withdrawal: Your participation in this study is completely voluntary and you may refuse to participate or choose to withdraw at any time without penalty or loss of benefits to which you are otherwise entitled.

By clicking "I Agree", you consent to participate in this experiment.
\end{tcolorbox}
\caption{Onboarding consent form.}
\label{fig:consent_form}
\end{figure}

\begin{figure}[htb!]\ContinuedFloat
\centering
\begin{tcolorbox}[width=\linewidth, colback=gray!5, colframe=black, boxrule=0.5pt, sharp corners, title=\textbf{Onboarding Consent Form}]
Please read this introduction carefully before participating in this experiment.

In this experiment we are interested in understanding how people discuss various topics in online platforms like Twitter or Reddit. To start, you will be given a statement (e.g. "Smoking cigarettes causes cancer") and asked to write a short post explaining whether or not you think it is true and why.

You will then have a series of three discussions with each of three other study participants using an online texting interface. In each discussion, do your best to keep the conversation going and to stay on topic.

After each discussion, you will be asked to again summarize your current view of the topic in a new post; the new summary can reiterate your initial position or can differ if your thinking has changed. It is important that it captures your own true belief about the topic.

The study will end after the third discussion, and will take 20 minutes to complete.

Press the BUTTON when you are ready to begin.

\end{tcolorbox}
\caption{Onboarding introduction (continued).}
\label{fig:intro}
\end{figure}

\begin{figure}[htb!]\ContinuedFloat
{\raggedright Initial Opinion\par}
\centering
\includegraphics[width=\linewidth]{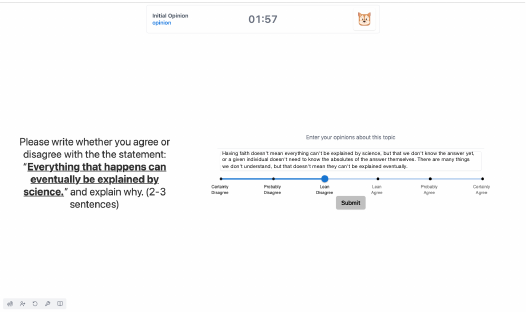}
\caption{Initial Opinion screen (continued).}
\label{fig:initial}
\end{figure}

\begin{figure}[htb!]\ContinuedFloat
{\raggedright Round 1\par}
\centering
\includegraphics[width=\linewidth]{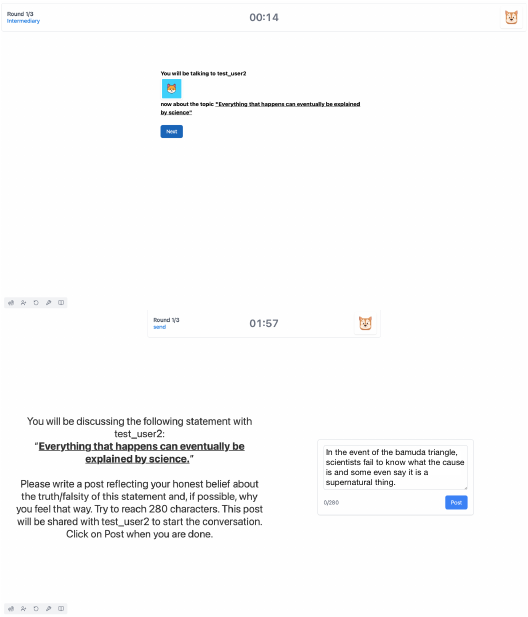}
\caption{Round 1 screen (continued).}
\label{fig:round1a}
\end{figure}

\begin{figure}[htb!]\ContinuedFloat
\centering
\includegraphics[width=\linewidth]{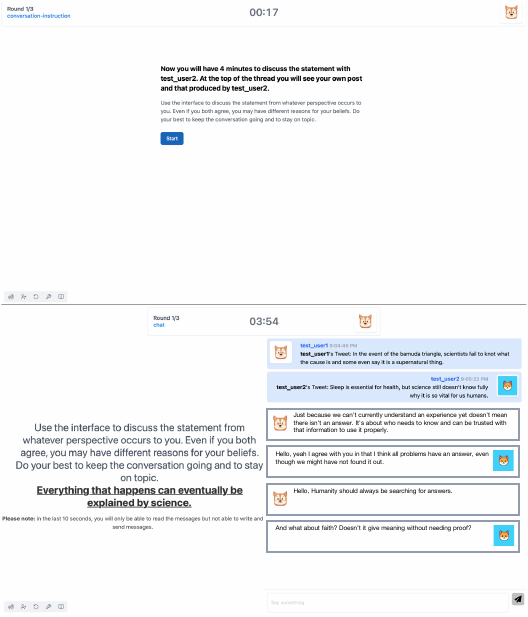}
\caption{Round 1 screen (continued).}
\label{fig:round1b}
\end{figure}

\begin{figure}[htb!]\ContinuedFloat
{\raggedright Round 2\par}
\centering
\includegraphics[width=\linewidth]{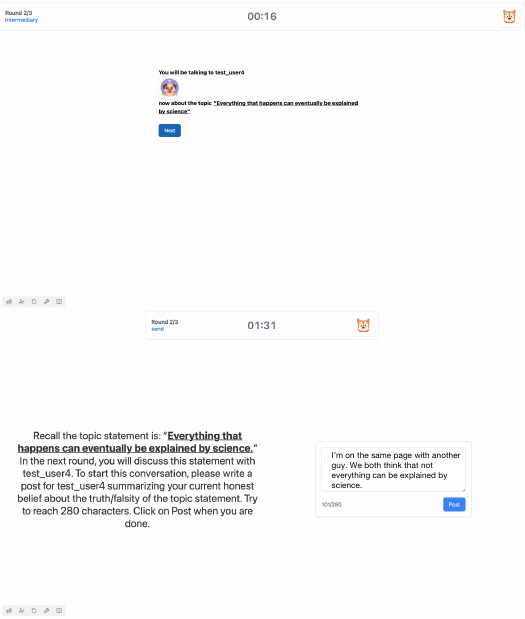}
\caption{Round 2 screen (continued).}
\label{fig:round2a}
\end{figure}

\begin{figure}[htb!]\ContinuedFloat
\centering
\includegraphics[width=\linewidth]{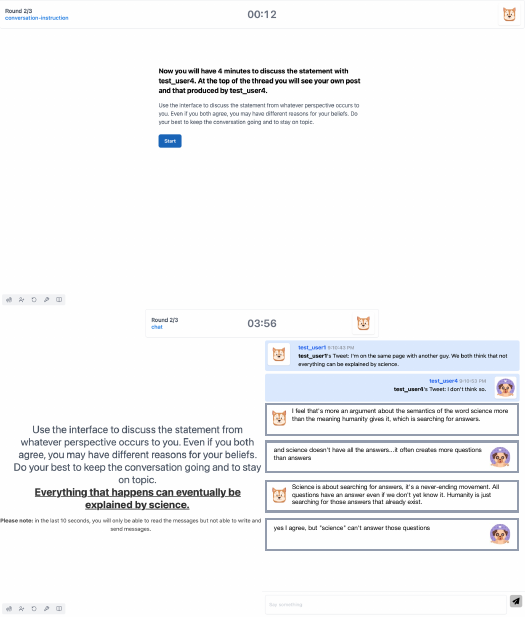}
\caption{Round 2 screen (continued).}
\label{fig:round2b}
\end{figure}

\begin{figure}[htb!]\ContinuedFloat
{\raggedright Round 3\par}
\centering
\includegraphics[width=\linewidth]{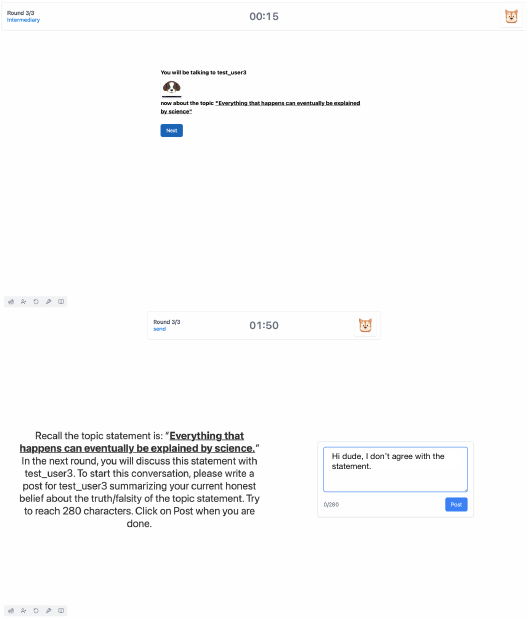}
\caption{Round 3 screen (continued).}
\label{fig:round3a}
\end{figure}

\begin{figure}[htb!]\ContinuedFloat
\centering
\includegraphics[width=\linewidth]{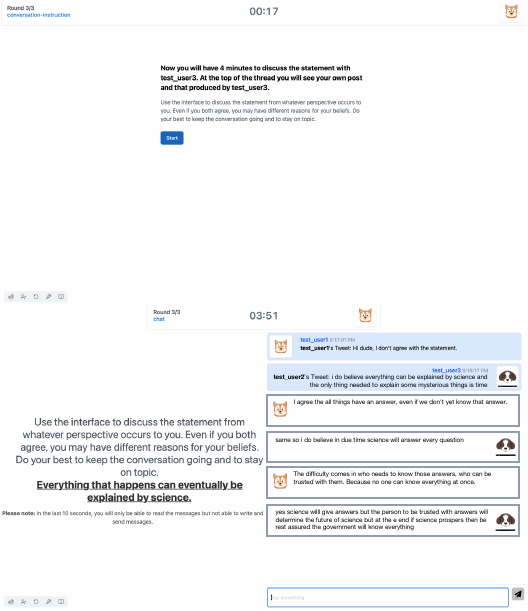}
\caption{Round 3 screen (continued).}
\label{fig:round3b}
\end{figure}

\begin{figure}[htb!]\ContinuedFloat
{\raggedright Post Opinion\par}
\centering
\includegraphics[width=\linewidth]{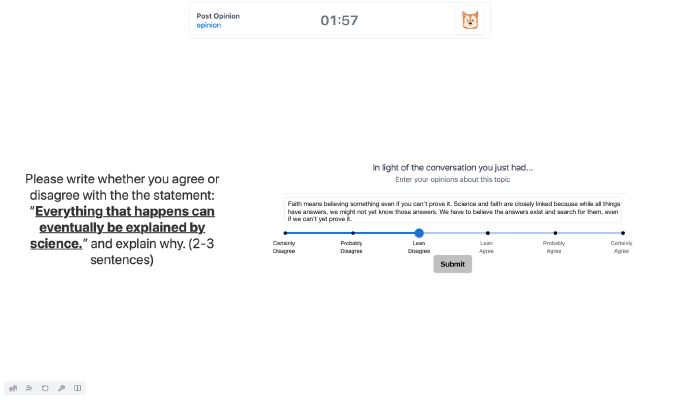}
\caption{Post Opinion screen (continued).}
\label{fig:post}
\end{figure}

\begin{figure}[!htb]\ContinuedFloat
\centering
\begin{tcolorbox}[width=\linewidth, colback=gray!5, colframe=black,
  boxrule=0.5pt, sharp corners,
  title=\textbf{Demographic Survey}]
\small
\setlist[enumerate]{leftmargin=*, itemsep=2pt, topsep=2pt}
\setlist[itemize]{leftmargin=*, itemsep=1pt, topsep=1pt}

\begin{enumerate}
  \item What is your age?\\(Open numeric input, 0--120)

  \item What is your gender?
  \begin{itemize}
    \item Male \item Female \item Something else (specify) \item Prefer not to answer
  \end{itemize}

  \item What is your country/region of residency?
  \begin{itemize}
    \item List of countries \item Other (specify) \item Prefer not to answer
  \end{itemize}

  \item What is your country/region of origin?
  \begin{itemize}
    \item List of countries \item Other (specify) \item Prefer not to answer
  \end{itemize}

  \item What is the highest level of education you have completed?
  \begin{itemize}
    \item High school or less \item Some college \item Associate’s degree
    \item Bachelor’s degree \item Master’s or Doctoral degree
    \item Professional degree \item Prefer not to answer
  \end{itemize}

  \item What is your race/ethnicity? (Select all that apply)
  \begin{itemize}
    \item Black or African American (non-Hispanic) \item Hispanic
    \item White (non-Hispanic) \item Asian, South Asian, or Pacific Islander
    \item Native American or American Indian \item Other (specify)
    \item Prefer not to answer
  \end{itemize}
\end{enumerate}

\end{tcolorbox}
\caption{Demographic survey (continued).}
\label{fig:demographic1}
\end{figure}

\begin{figure}[htb!]\ContinuedFloat
\centering
\begin{tcolorbox}[width=\linewidth, colback=gray!5, colframe=black,
  boxrule=0.5pt, sharp corners, 
  title=\textbf{Demographic Survey (continued)}]
\small
\setlist[enumerate]{leftmargin=*, itemsep=2pt, topsep=2pt}
\setlist[itemize]{leftmargin=*, itemsep=1pt, topsep=1pt}

\begin{enumerate}
  \setcounter{enumi}{6} 

  \item What is your household annual income?
  \begin{itemize}
    \item Less than \$25{,}000 \item \$25{,}000–\$49{,}999
    \item \$50{,}000–\$74{,}999 \item \$75{,}000–\$99{,}999
    \item \$100{,}000–\$149{,}999 \item \$150{,}000–\$199{,}999
    \item Over \$200{,}000 \item Prefer not to answer
  \end{itemize}

  \item Generally speaking, do you usually think of yourself as a(n)...
  \begin{itemize}
    \item Strong Republican \item Republican \item Independent
    \item Democrat \item Strong Democrat \item Other (specify)
    \item Prefer not to answer
  \end{itemize}

  \item If Independent or Other: Do you lean more toward…
  \begin{itemize}
    \item Republican Party \item Democratic Party \item Neither
  \end{itemize}

  \item In general, would you describe your political views as...
  \begin{itemize}
    \item Very conservative \item Conservative \item Moderate
    \item Liberal \item Very liberal \item Don’t know
    \item Prefer not to answer
  \end{itemize}

  \item Are you currently...
  \begin{itemize}
    \item Married \item Living with a partner but not married
    \item Widowed \item Divorced \item Separated
    \item Never been married \item Prefer not to answer
  \end{itemize}

  \item If you have children, where do they go to school? (Select all that apply)
  \begin{itemize}
    \item Public school \item Private school \item Home school
    \item University or Technical/Community College
    \item They are out of school \item I don’t have children
    \item Prefer not to answer
  \end{itemize}
\end{enumerate}

\end{tcolorbox}
\caption{Demographic survey (continued).}
\label{fig:demographic2}
\end{figure}

\begin{figure}[htb!]\ContinuedFloat
\centering
\begin{tcolorbox}[width=\linewidth, colback=gray!5, colframe=black,
  boxrule=0.5pt, sharp corners, 
  title=\textbf{Demographic Survey (continued)}]
\small
\setlist[enumerate]{leftmargin=*, itemsep=2pt, topsep=2pt}
\setlist[itemize]{leftmargin=*, itemsep=1pt, topsep=1pt}

\begin{enumerate}
  \setcounter{enumi}{12} 

  \item Which of these statements comes closest to describing your feelings about the Bible?
  \begin{itemize}
    \item The Bible is the actual word of God and is to be taken literally, word for word.
    \item The Bible is the inspired word of God but not everything in it should be taken literally.
    \item The Bible is an ancient book of fables, legends, history, and moral precepts recorded by men.
    \item Prefer not to answer
  \end{itemize}

  \item Would you describe yourself as a "born-again" or evangelical Christian, or not?
  \begin{itemize}
    \item Yes, would \item No, would not \item Don’t know \item Prefer not to answer
  \end{itemize}

  \item What is your religious preference?
  \begin{itemize}
    \item Protestant \item Roman Catholic \item Jewish \item Muslim/Islam
    \item Mormon/Latter-Day Saints \item Other Christian Religion
    \item Other Non-Christian Religion \item No Religion/Atheist/Agnostic
    \item Don’t know \item Prefer not to answer
  \end{itemize}

  \item Which of the following best describes the kind of work you do?
  \begin{multicols}{3}
  \begin{itemize}
    \item Agriculture, Forestry, Fishing and Hunting
    \item Mining, Oil and Gas Extraction, and Utilities
    \item Construction
    \item Manufacturing
    \item Wholesale Trade
    \item Sales \&/or Retail Trade
    \item Transportation and Warehousing
    \item Media, Communications, and Digital Entertainment
    \item Finance, Accounting, and Consulting
    \item Insurance
    \item Real Estate, Rental, and Leasing
    \item Personal Care and Services
    \item Research
    \item Engineering, Computer-Related Design, and Architecture
    \item Law and Legal Services
    \item Education
    \item Health Care and Social Assistance
    \item Arts, Entertainment, and Recreation
    \item Restaurant, Travel, and Lodging
    \item Non-Profit, Community, Religious and Social Service Organizations
    \item Maintenance and Repair Services
    \item Cleaning Services
    \item Government
    \item Other (specify)
    \item Prefer not to answer
  \end{itemize}
  \end{multicols}
\end{enumerate}

\end{tcolorbox}
\caption{Demographic survey (continued).}
\label{fig:demographic3}
\end{figure}






\fi

\end{document}